\documentclass{article} 
\usepackage{iclr2021_conference,times}

\usepackage{amsmath,amsfonts,bm}

\def\eqref#1{equation~\ref{#1}}

\def\floor#1{\lfloor #1 \rfloor}
\def\1{\bm{1}}

\DeclareMathAlphabet{\mathsfit}{\encodingdefault}{\sfdefault}{m}{sl}
\SetMathAlphabet{\mathsfit}{bold}{\encodingdefault}{\sfdefault}{bx}{n}

\usepackage{hyperref}
\usepackage{url}

\usepackage{booktabs}       
\usepackage{amsfonts}       
\usepackage{nicefrac}       
\usepackage{microtype}      
\usepackage{xcolor} 
\usepackage{graphicx}
\usepackage{amsmath}
\usepackage{subcaption}
\usepackage[capitalise]{cleveref} 
\usepackage{amssymb}
\usepackage{units}
\usepackage{multirow}
\usepackage{xr}
\usepackage{xfrac}
\usepackage{mathtools}
\usepackage{hyperref}
\usepackage{algorithm}
\usepackage{algpseudocode}

\newcommand*\samethanks[1][\value{footnote}]{\footnotemark[#1]}

\title{Neural gradients are near-lognormal:     \\ improved quantized and sparse training}

\author{
Brian Chmiel\,${^\dagger}{^\circ}$\thanks{Equal contribution.}\quad
Liad Ben-Uri\,${^\circ}$\samethanks[1]\quad
Moran Shkolnik\,${^\dagger}{^\circ}$\quad
\\[0.15cm]\textbf{ Elad Hoffer\,${^\dagger}$\quad
Ron Banner\,$^\dagger$\quad
Daniel Soudry\,$^\circ$}
\\[0.2cm]
$^\dagger$Habana Labs  --  An Intel company, Caesarea, Israel,\\
$^\circ$Department of Electrical Engineering - Technion, Haifa, Israel
\\[0.2cm]
\small{\texttt{\{\href{mailto:bchmiel@habana.ai}{bchmiel},
\href{mailto:mshkolnik@habana.ai}{mshkolnik},
\href{mailto:ehoffer@habana.ai}{ehoffer},
\href{mailto:rbanner@habana.ai}{rbanner}\}@habana.ai}}\\
\small{\texttt{\{\href{mailto:liadgo2@gmail.com}{liadgo2},   \href{mailto:daniel.soudry@gmail.com}{daniel.soudry}\}@gmail.com}}\\
}

\iclrfinalcopy 
\begin{document}

\maketitle

\begin{abstract}
While training can mostly be accelerated by reducing the time needed to propagate neural gradients back throughout the model, most previous works focus on the quantization/pruning of weights and activations. These methods are often not applicable to neural gradients, which have very different statistical properties. Distinguished from weights and activations, we find that the distribution of neural gradients is approximately lognormal. Considering this, we suggest two closed-form analytical methods to reduce the computational and memory burdens of neural gradients. The first method optimizes the floating-point format and scale of the gradients. The second method accurately sets sparsity thresholds for gradient pruning.  Each method achieves state-of-the-art results on ImageNet. To the best of our knowledge, this paper is the first to (1) quantize the gradients to 6-bit floating-point formats, or (2) achieve up to 85\% gradient sparsity --- in each case without accuracy degradation. \href{https://github.com/brianchmiel/Neural-gradients-are-lognormally-distributed-understanding-sparse-and-quantized-training} {Reference implementation} accompanies the paper.


\end{abstract}

\section{Introduction}
Neural gradients (gradients with respect to intermediate neural layer outputs) are used in the training process of deep networks to backpropagate the error-gradient throughout the model, thus allowing to compute the required weight updates. As these neural gradients are needed for a substantial ratio of the underlying computations (about $\frac{2}{3}$), compressing them can alleviate data-throughput requirements and accelerate the training process.

Many previous works \citep{banner2018post,Fang2020PostTrainingPL} compress tensors such as weights and activations by approximating their distributions using an analytically
tractable density. These works often assume a bell-shaped distribution such as Gaussian or Laplace distributions, which have been reported to fail for neural gradients \citep{acceleratedSpars2019}. One key observation in this paper is that neural gradient distributions are heavy-tailed,  fundamentally different from the light-tailed distributions of weights and activations. Further statistical and distributional tests reveal gradient magnitudes follow a lognormal distribution.


Adopting this lognormal observation, our paper suggests two main applications --- quantization and pruning, used to reduce the computational and memory burden of neural gradients. To tackle these challenges, we first formalize the problems and find closed-form expressions that enable us to predict the optimal quantization and pruning policies. These measures are easy to use and depend only on the estimated lognormal parameters.

\begin{figure}[h]
\centering
  \includegraphics[width=0.65\linewidth]{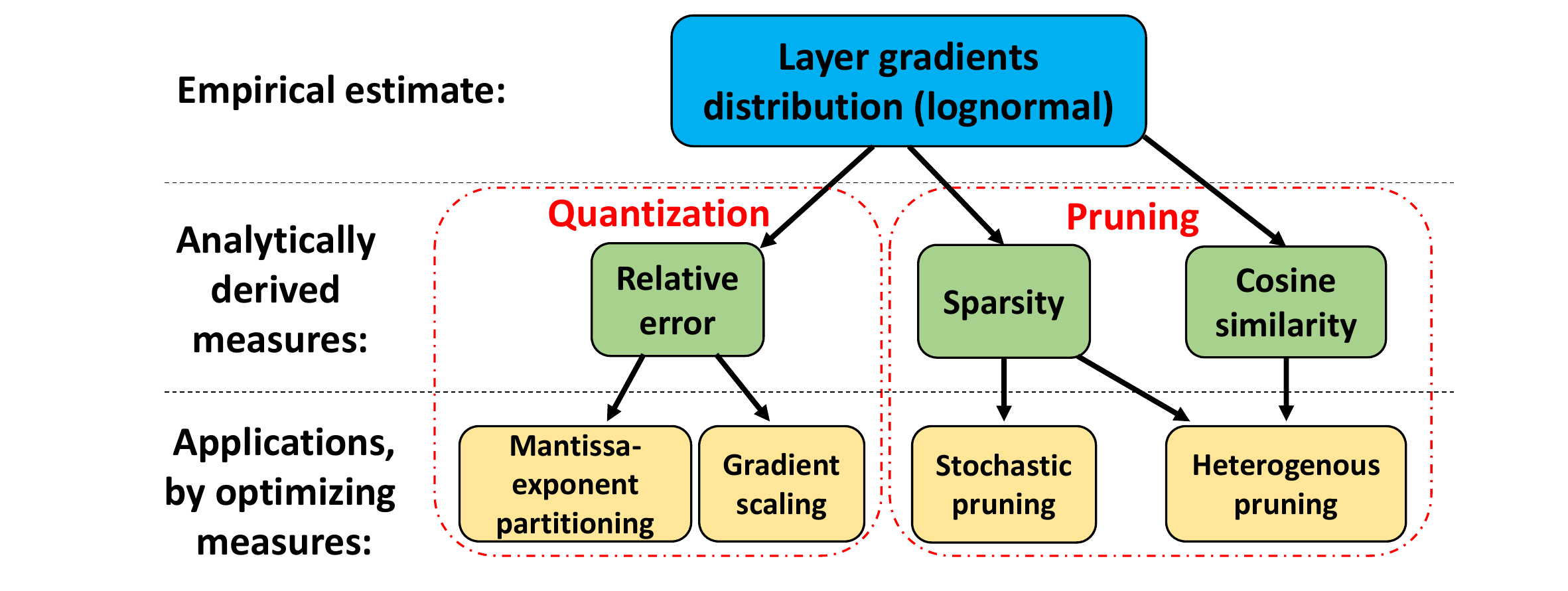}
  \caption{High-level overview of the different methods developed in this work. Adopting the lognormal observation of the neural gradients, we suggest three different measures we can \textit{analytically} optimize: two for pruning and one for quantization. These analytical measures are used in four different schemes.}
  \label{fig:Contributions}
  \vspace{-5mm}
\end{figure}

In Figure \ref{fig:Contributions} we summarize these applications and their derivation. The first application uses the lognormal prior to enabling low-precision floating-point (FP) quantization of the gradients. Here we optimize two tasks. The first task is to find a partition between mantissa and exponent bit-widths that minimizes quantization noise for a given $n$ bit FP gradient representation. The second task is to scale these gradients so that they would be properly represented within a limited dynamic range. We provide useful insights that make empirically-based heuristics such as \emph{loss scaling} \citep{Micikevicius2017MixedPT} a more grounded approach with a theoretical basis. Optimizing both tasks we obtain state-of-the-art results for FP quantization of the neural gradients. The second application performs accurate and predictable stochastic pruning of gradients on the fly, which results in two state-of-the-art pruning schemes. The first translates the desired sparsity level into an accurate threshold, and the other enables combined use of different sparsity levels at different layers (heterogeneous sparsity).

\section{Related Work}

Quantization and pruning of neural networks have seen a tremendous amount of works (e.g., \cite{Nahshan2019LossAP,Choi2018PACTPC,baskin2018nice,Louizos2017LearningSN,Frankle2018TheLT,Cambier2020ShiftedAS}) aiming to reduce both bandwidth and memory footprint, as well as computation time. Some of these methods  \citep{banner2018post,acceleratedSpars2019,Fang2020PostTrainingPL} use a systemic and rigorous statistical approach to optimize various distortion measures. For example, \cite{banner2018post} used the normal distributional assumption (of weights and activations) to analytically minimize the mean-squared quantization error. Our work follows a similar line to rigorously optimize similar performance measures for quantization and pruning of gradient distributions, which are different from that of the weights and activations.


\textbf{Gradient Quantization.}  While a lot of research focused on the quantization of weights and activations for inference \citep{krishnamoorthi2018quantizing,choi2018quantization,jain2019quantization}, there were also major advances in quantization during training, many of them faced difficulty trying to represent the high dynamic range of the gradients \citep{Banner2018ScalableMF,wu2018quantization}. \cite{Cambier2020ShiftedAS} suggests keeping for each tensor shift and scale in full precision numbers to make them fit in FP8 format dynamic range. Mantissa vs exponent allocation of the available bits has proven crucial in deep learning workloads, where for example, BF16 has shown greater success compared to traditional FP16 format due to wider dynamic range \citep{henry2019leveraging,kalamkar2019study}. Research over required format and trade-offs in exponent versus mantissa is on-going with growing interest over lower precision representations such as FP8. Some works have explored using different FP8 formats: \cite{wang2018quantization} has used (1-5-2: sign-exponent-mantissa), while \citep{Sun2019Hybrid8F} suggests using one type of FP8 format for the forward (1-4-3), and a different FP8 format (1-5-2) for the backward pass, after empirically assessing the different possible formats.
Additionally, with the growing usage of FP16 mixed-precision training, researchers and practitioners faced the need to use loss scaling \citep{Micikevicius2017MixedPT}, adjusting the tensor distribution to avoid over/under-flow. This procedure required, in some cases, an intensive parameter search and was not guaranteed to succeed in all cases. 

\textbf{Gradient pruning.}
Focusing on the computationally-intensive back-propagation, "meProp" \citep{meProp2017} prunes the K smallest absolute-valued entries of the neural gradients on the fly, using the top-k algorithm. Works following it, replaced pruning with quantization to induce sparsity \citep{wiedemann2020ditheredBackprop}, or used top-k pruning as well on the copies of weights and activations used in back-prop \citep{SWAT2020}. \cite{acceleratedSpars2019}, inspired by conventional unbiased estimators like stochastic rounding, suggested "stochastic pruning", reaching higher sparsity levels. Yet, the authors assumed the gradients are normally distributed, leading to incorrect estimation of the threshold and a large difference between the required sparsity and the one obtained. As we shall see later, using the correct statistical distribution model is essential to determine the proper threshold.

\section{Neural gradients distribution}


Many prior works \citep{banner2018post,Bernstein2018signSGDCO} take the assumption that tensor data e.g., weights ($W$) and activations ($\mathcal{A}$) is sampled from a Gaussian distribution. Recently, \citet{acceleratedSpars2019,wiedemann2020ditheredBackprop} used the same assumption for the distribution of neural gradients $\nabla \mathcal{A}$. In the section, we discuss this assumption. We show that neural gradients are better approximated by lognormal distributions, i.e., the gradient logarithm values are normally distributed, as opposed to the gradients themselves.



\begin{figure}[h]
\begin{subfigure}{.24\textwidth}
  \centering
  \includegraphics[width=\linewidth]{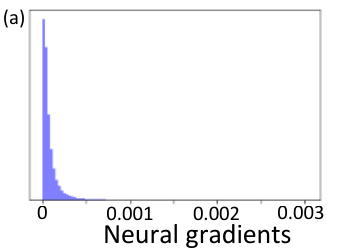}  
  \label{fig:NormalFit}
 \end{subfigure}
\begin{subfigure}{.24\textwidth}
  \centering
  \includegraphics[width=\linewidth]{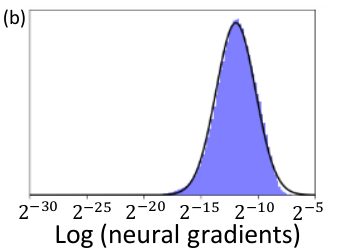}  
  \label{fig:LognormalFit}
\end{subfigure}
\begin{subfigure}{.24\textwidth}
  \centering
  \includegraphics[width=\linewidth]{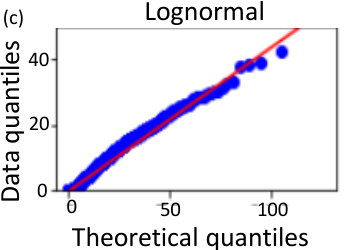}  
  \label{fig:lognormalQuantiles}
\end{subfigure}
\begin{subfigure}{.24\textwidth}
  \centering
  \includegraphics[width=\linewidth]{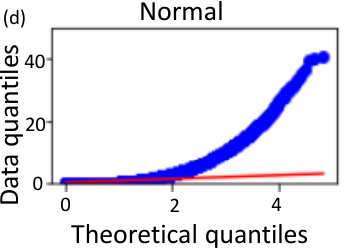}  
  \label{fig:normalQuantiles}
\end{subfigure}
\caption{Identifying the distribution of neural gradients (normal vs. lognormal): \textbf{(a)} probability density function of gradient magnitudes; \textbf {(b)} the same histogram on a logarithmic scale --- notice that the shape looks very much like the symmetrical shape of the regular normal distribution; \textbf {(c)} quantiles of gradient magnitudes against the quantiles of a lognormal distribution; and  \textbf {(d)} quantiles of gradient magnitudes against the quantiles of a normal distribution. Additional layers' distribution in \cref{fig:normalVsLognormalApp}. } 
\label{fig:normalVsLognormal}
\vspace{-5mm}
\end{figure}

In \cref{fig:normalVsLognormal}, we plot the histogram of neural gradient magnitudes at linear and log scales in one layer of ResNet18 - ImageNet dataset.  At a linear scale, the distribution has a few gradients of huge magnitude and a great many gradients of small magnitudes (Fig 2a). Plotting the histogram on a logarithmic scale reveals a distribution close to a symmetric normal (Gaussian) distribution (Fig 2b). This is the hallmark of the lognormal distribution.  Finally, when plotting the theoretical quantiles of the lognormal distribution against the quantiles of the gradient distribution (Q-Q plot), we see that almost all points lie on a straight line (Fig 2c). Note that the points follow a strongly nonlinear pattern in Fig 2d, suggesting that the data is not distributed as a standard normal distribution.



We further estimate the goodness of fit of the neural gradients to normal and lognormal distributions. To that end, we measure the static distance (largest vertical line) between the cumulative distribution function (CDF) of the empirically observed distribution and the CDF of the reference distribution (also known as Kolmogorov-Smirnov test \citep{KS_Test}). For each model and dataset in \cref{tab:KS-test}, we calculate the average (across all layers) of the static distance to normal and lognormal distributions. The analysis is performed on the absolute value of the gradients, excluding the zero-valued entries. Note that lognormal distribution gets a better fit. Additional statistical distributions in \cref{secAp: dist discussion} .   

\begin{table}[]
\centering
\caption{Mean ($\pm$ std) over all layers over time of KS test on different models and datasets for normal and lognormal distribution. Notice the lognormal distribution gets the higher fit across all models. We compare to additional distributions such as Laplace, uniform, Cauchy and loglaplace in \cref{secAp: dist discussion}
}
\label{tab:KS-test}
\resizebox{\columnwidth}{!}{%
\begin{tabular}{c||c|c|c|c|c|c}
\toprule
\multirow{3}{*}{Distribution} & \multicolumn{6}{c}{\multirow{2}{*}{\begin{tabular}[c]{@{}c@{}}Model\\ (Dataset)\end{tabular}}} \\
 & \multicolumn{6}{c}{} \\ \cline{2-7} 
 & \begin{tabular}[c]{@{}c@{}}BERT\\ (CoLa)\end{tabular} & \begin{tabular}[c]{@{}c@{}}BERT\\ (MRPC)\end{tabular} & \begin{tabular}[c]{@{}c@{}}ResNet18\\ (ImageNet)\end{tabular} & \begin{tabular}[c]{@{}c@{}}MobileNetV2\\ (ImageNet)\end{tabular} & \begin{tabular}[c]{@{}c@{}}VGG16\\ (ImageNet)\end{tabular} & \begin{tabular}[c]{@{}c@{}}DenseNet121\\ (ImageNet)\end{tabular} \\ \midrule
\multicolumn{1}{l||}{Normal} & 0.46$\pm$0.02 & 0.39$\pm$0.04 & 0.38$\pm$0.1 & 0.22$\pm$0.09 & 0.35$\pm$0.08 & 0.33$\pm$0.1 \\ \hline
\multicolumn{1}{l||}{Lognormal} & \textbf{0.05$\pm$0.002} & \textbf{0.04$\pm$0.002} & \textbf{0.02$\pm$0.002} & \textbf{0.07$\pm$0.003} & \textbf{0.06$\pm$0.002} & \textbf{0.05$\pm$0.001} \\ 
\bottomrule
\end{tabular}}

\end{table}

\section{Application I - Optimal Floating-Point Quantization} 
Floating-point (FP) representations can cover a very wide dynamic range with a relatively small number of digits. This dynamic range is especially important for the heavy-tailed distributions that characterize neural gradients. In this section, we study the optimal characteristics of the FP quantizer.  

\subsection{Problem formulation}
We can decompose any positive real value $x\in \mathbb{R}^{+}$ as follows:
\begin{equation}
     x = 2^{\ln x}= \overbrace{2^{\ln x-\lfloor\ln x\rfloor }}^{ M\in [1,2)} \cdot 2^{\overbrace{\lfloor\ln x\rfloor}^{ E\in\mathbb{Z}}},
\end{equation}
where $M\in [1,2)$ is the mantissa and $E\in \mathbb{Z}$ the exponent. Given $N$ bits, we allocate 1 bit for the sign and seek the optimal allocation of  $n_1$ bits to $M$ and $n_2$ bits to $E$, such that $n_1 + n_2 = N-1$. Accordingly, we define the quantized $x_q$ as: 
\begin{equation}
      x_q = \Bigg\{\begin{array}{lr}
        2^{E_{\max}} &  E \geq E_{\max}\\
        M_q\cdot 2^{E} &  -E_{\max}\leq E\leq E_{\max}\\
        0 &  E\leq -E_{\max}
        \end{array}
\end{equation}
where $E_{\max} = 2^{n_2-1}$ and $M_q$ is the quantized mantissa with the range $[1,2)$ divided into $2^{n_1}$ quantization levels with a spacing of $\Delta = \frac{1}{2^{n_1}}$.

Finally, we measure the relative error between the FP number $x_q$ and the real number $x$, which is simply the difference between the two numbers divided by the real number \citep{widrow2008quantization}:

\begin{equation}
   \eta(n_1,n_2) = \left| \frac{x_q-x}{x}\right|
   \label{eq:relativeError}
\end{equation}
\subsection{Analytical derivation of the relative error}

We assume that $x \sim \mathrm{Lognormal}(\mu,\sigma^2)$. Note that $E=\lfloor\ln x\rfloor\approx \ln x \sim \mathcal{N}(\mu,\sigma^2)$. In \cref{secApp:fp8RelativeFull} we split the range into three parts according to $E$: (i) $-E_{\max}\leq E\leq E_{\max}$; (ii) $E\geq E_{\max}$; (iii) $E\leq -E_{\max}$, and calculate the expected contribution for each term. A closed-form formula for the expected relative error could be obtained as follows:  
 \begin{equation}
\begin{split}
 E\left[\eta(n_1,n_2) \right] =& \frac{2\Phi\left(\frac{E_{\max}}{\sigma}\right) -1}{8\cdot \ln\left(2\right)\cdot (2^{n_1}-1)} + 
 2^{E_{\max}-1}\mathrm{e}^\frac{\sigma^2\ln^2\left(2\right)}{2}\left(\operatorname{erf}\left(\frac{\sigma \ln 2 }{\sqrt 2}+\frac{E_{\max}}{\sqrt{2}\sigma}\right)-1\right)
   \\ &-\frac{1}{2}\operatorname{erf}\left(\dfrac{E_{\max}}{\sqrt 2\sigma}\right)+\frac{3}{2}- \Phi\left(\frac{E_{\max}}{\sigma}\right)
 \end{split}
 \label{eq:relativeFinal}
\end{equation}
where $\Phi(x)$ is the CDF of $\mathcal{N}(0,1)$. In \cref{analysisVsExpiremnt} we show that analytical results stated by \cref{eq:relativeFinal} are in good agreement with simulations for FP8 with various number of exponent bits. Simulations were obtained by quantizing 10,000 values, generated from a lognormal distribution with $\sigma=1,3,5$.

\subsection{The optimal mantissa-exponent representation}

\begin{figure}[h]
\begin{subfigure}{.31\textwidth}
  \centering
  \includegraphics[width=\linewidth]{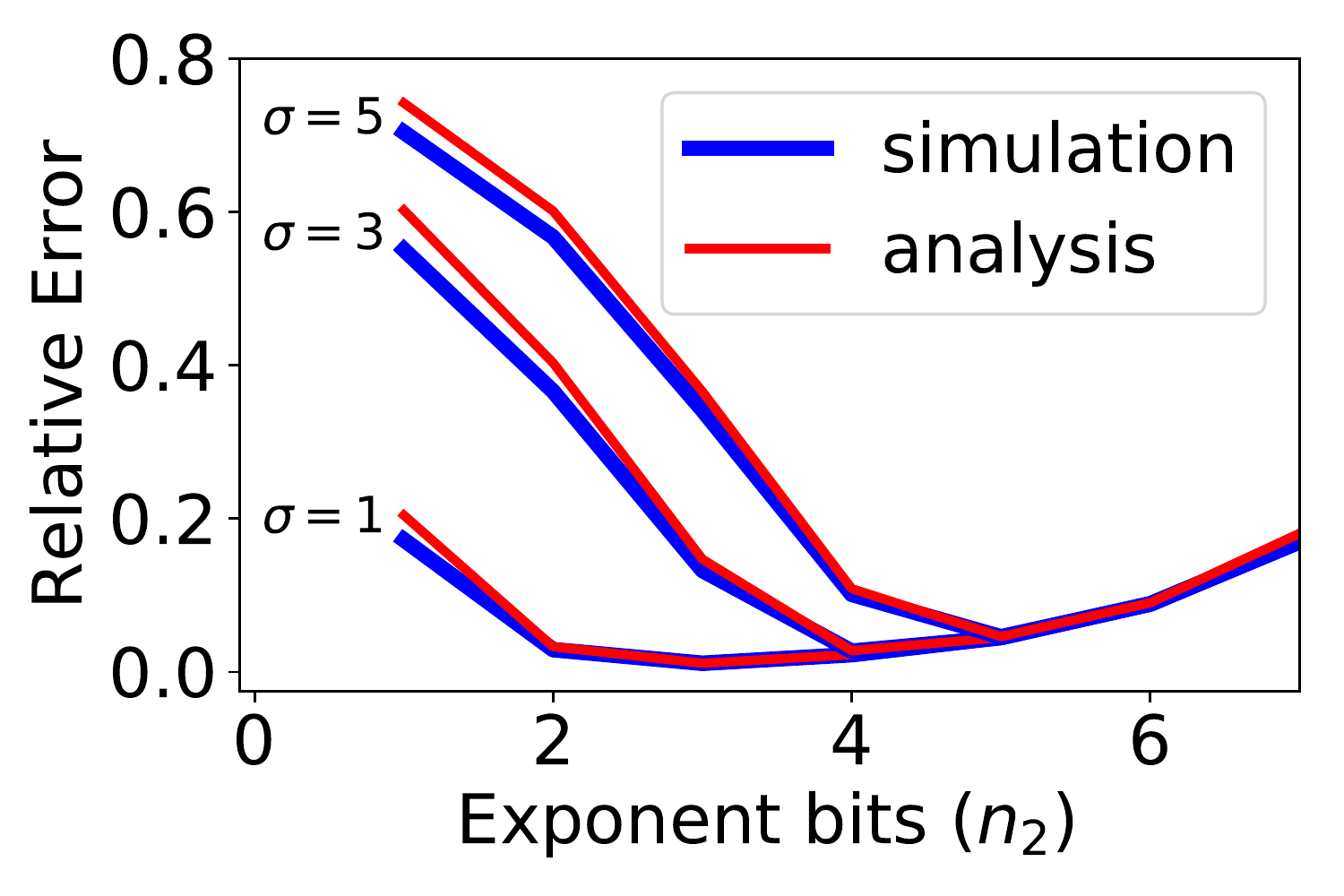}  
  \label{analysisVsExpiremnt}
 \end{subfigure}
\begin{subfigure}{.34\textwidth}
  \centering
  \includegraphics[width=\linewidth]{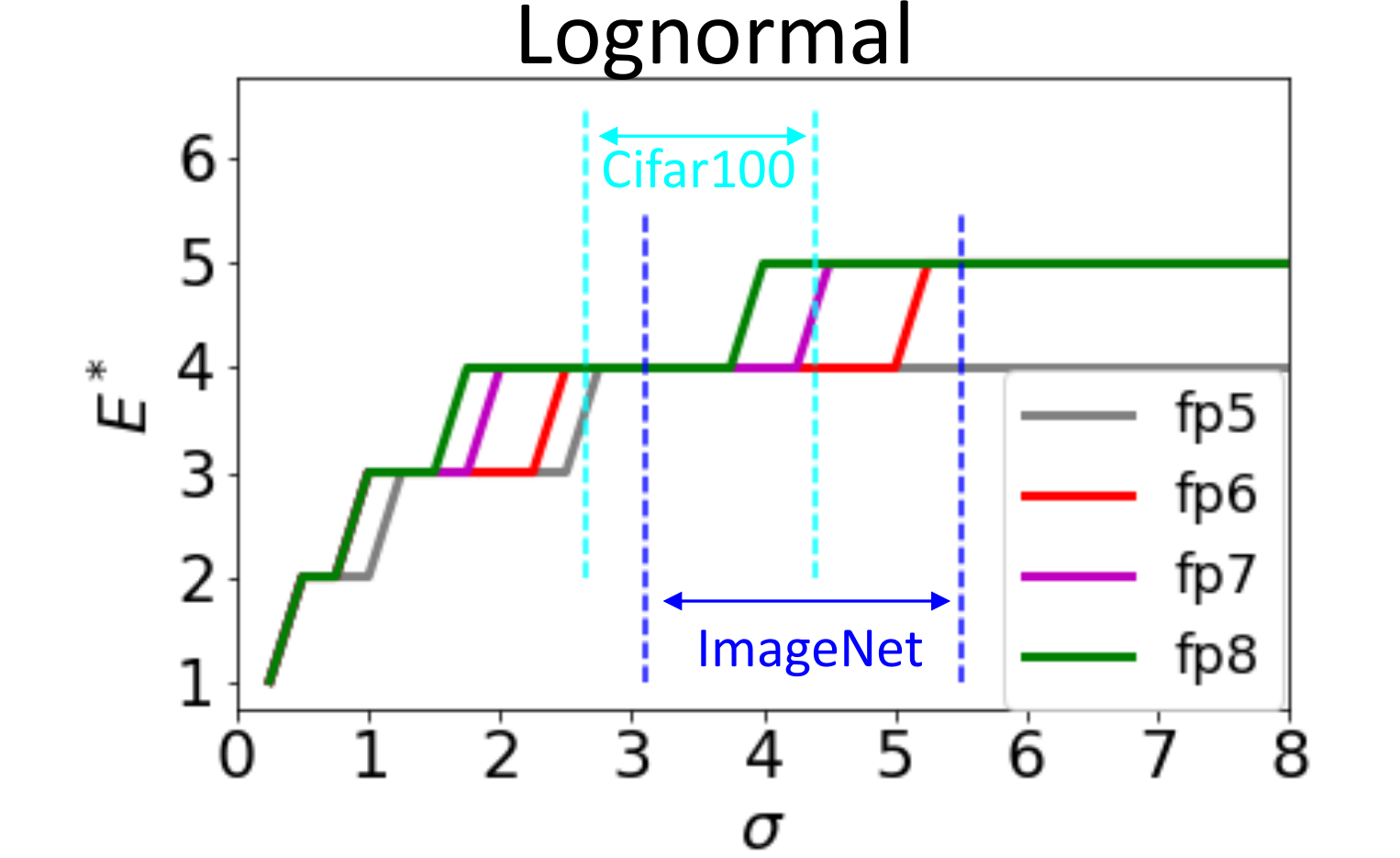}  
  \label{fig:expBits}
\end{subfigure}
\begin{subfigure}{.34\textwidth}
  \centering
  \includegraphics[width=\linewidth]{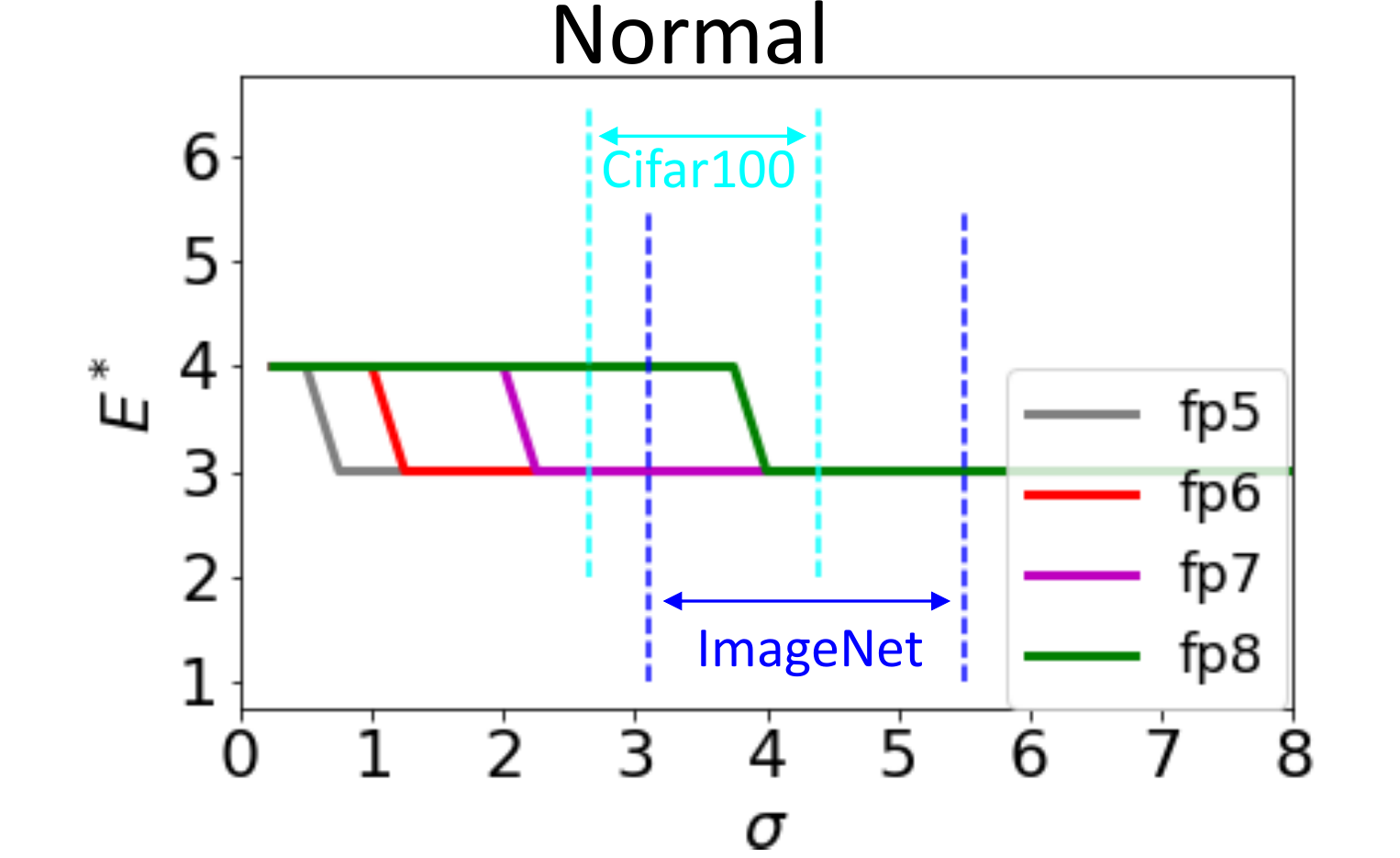}  
  \label{fig:expBitsNormal}
\end{subfigure}
\caption{\textbf{(Left)} Expected relative error as a function of $n_2$ for FP8 with a lognormal distribution. The simulation is in good agreement with the analytical results stated by \cref{eq:relativeFinal}. \textbf{(Middle)} Ideal bit allocation for the exponent ($E^*$) as a function of $\sigma$, using \cref{eq:relativeFinal}. Notice that $\sigma$ varies between datasets, imposing different optimal allocations. \textbf{(Right)}
Ideal bit allocation assuming a normal distribution, which leads to wrong FP formats, e.g., the number of bits assigned to the exponent decreases when $\sigma$ (dynamic range) increases. } 
\label{fig:fp8format}
\vspace{-2mm}
\end{figure}

The relative error in Eq. \ref{eq:relativeFinal} depends on the scale parameter $\sigma$, the number of bits of the mantissa  $n_1$ and exponent $n_2$, respectively (the latter through $E_{\max}=2^{n_2-1}$). Given any $N$-bit FP format, we wish to find a mantissa-exponent partition that minimizes the expected relative error such that $n_1+n_2=N-1$. Minimizing Eq. \ref{eq:relativeFinal} yields this optimal partition. To do so we set $n_1= N-n_2-1$, equate the derivative to zero and solve. The computational cost of such a solution is negligible (details are in \cref{SecAppen:finalRelated}). This allocation depends on $N$ and $\sigma$. In \cref{fig:expBits} we show the optimal allocations for $N=5,6,7,8$ bits and $\sigma \in [1,8]$. In \cref{fig:expBitsNormal} we show the FP format obtained by solving \cref{eq:relativeError} with normal distribution assumption for neural gradients, which leads to a sub-optimal format as shown in \cref{tab:fpresults}. The full solution can be found in \cref{secApp:fp8RelativeFull-Normal}.

Table \ref{tab:fpFormats} summarizes the statistical analysis applied for various floating point formats. Empirical observations prove that gradient distributions have a range of [3,5.5] and  $[2.5,4.5]$ in ImageNet and Cifar100 datasets respectively, which we use to determine the optimal mantissa-exponent partition for FP4--FP8. In Section \ref{sec:experiments}, we use these partitions to train ImageNet and Cifar models at reduced precision (bit-width lower than 8), and show empirically that these partitions provide the best results.

\begin{table}[h]
\centering
\caption{Ideal sign-exponent-mantissa bit allocations based on the optimization of Equation \ref{eq:relativeFinal} and the gradient statistics of ResNet18, ResNet101 and SqueezeNet in Cifar100 and ImageNet datasets. The analysis matches the previously suggested allocation by \cite{Sun2019Hybrid8F} for FP8, showing a clear benefit of 1-5-2 for gradient quantization.}
\label{tab:fpFormats}
  \vskip 0.1in
  \centering
\begin{tabular}{c|c|c|c|c|c|c|c}
    \toprule

\multicolumn{1}{c|}{Models} & \multicolumn{1}{c|}{Dataset}  &  \multicolumn{1}{c|}{$\sigma$ Range } & 
\multicolumn{1}{c|}{FP4} & \multicolumn{1}{c|}{FP5} & \multicolumn{1}{c|}{FP6} & \multicolumn{1}{c|}{FP7} & \multicolumn{1}{c}{FP8} \\ 

\midrule
\multicolumn{1}{c|}{\multirow{1}{*}{ResNet18, ResNet101}} &
\multicolumn{1}{c|}{\multirow{1}{*}{Cifar100}} & \multicolumn{1}{c|}{\multirow{1}{*}{2.5-4.5}} & 
\multicolumn{1}{c|}{1-3-0}  & \multicolumn{1}{c|}{1-4-0} & \multicolumn{1}{c|}{1-4-1} & \multicolumn{1}{c|}{1-4-2} & \multicolumn{1}{c}{1-5-2}
\\ \midrule
\multicolumn{1}{c|}{\multirow{1}{*}{ResNet18, SqueezeNet}} &
\multicolumn{1}{c|}{\multirow{1}{*}{ImageNet}} & \multicolumn{1}{c|}{\multirow{1}{*}{3-5.5}} &  
\multicolumn{1}{c|}{1-3-0}  & \multicolumn{1}{c|}{1-4-0} & \multicolumn{1}{c|}{1-5-0} & \multicolumn{1}{c|}{1-5-1} & \multicolumn{1}{c}{1-5-2}

\\  \bottomrule

\end{tabular}
\end{table}

\subsection{Per-Layer Gradient scaling}

The use of loss scaling \citep{Micikevicius2017MixedPT}, is key to the quantization of gradients using low precision FP formats. The idea is to shift the neural gradients' dynamic range to fit the floating-point range thus avoiding possible underflows. Loss scaling is usually performed by multiplying the loss value by a large constant and then dividing weight gradients by the same value after back-propagation and before any update has taken place.  
As gradient distribution changes across training, dynamic loss scaling is sometimes needed.
In this setting, gradients are monitored for any overflow or underflow that may occur. Upon this occurrence, gradients are discarded for that step and loss scale value is changed to combat the phenomena (either increased or decreased heuristically). 

Choosing either a fixed or dynamic \emph{global} loss scale that can fit across all the layers is challenging and may prove impossible for low precision FP formats that further reduce the dynamic range. In \cref{fig:transformerStd}, we show the standard deviation $\sigma$ of the gradient at the log scale for different transformer layers. The high variability in $\sigma$ between the layers makes the choice of one global loss scale unlikely. The variation in gradient statistics across layers may also explain the need for previous hybrid approaches that kept some layers at higher precision.  For example, \citet{Sun2019Hybrid8F} reported that when training ResNet18 models using FP8, they had to keep the first and last layers at FP16 precision. \cref{fig:resnet_first_last} suggests a reason --- the first and last layers exhibit std that is very different from the other layers (i.e., they require a different gradient scaling ).  \cite{Cambier2020ShiftedAS} showed the low results achieved using a global loss scale in FP8 format in transformer and suggested a full precision (expensive) computational operation of rescaling and shifting the neural gradients to FP8 dynamic range. 

We therefore suggest to use a per-layer \textit{gradient scale}, instead of a global loss scale. As detailed in \cref{sec:apGradientScaling}, our gradient scaling method keeps the largest gradients representable but sacrifices the smallest gradients. These tiny gradients can be pruned without significantly distorting the original tensor because: (1) the lognormal distribution suggests that gradients of tiny magnitude are relatively scarce; (2) such tiny gradients are typically less significant in training compared to the larger gradients. Pseudo-code appears in \cref{alg:scale}




\section{Application II - Stochastic pruning}

Inspired by conventional unbiased estimators for quantization such as "stochastic rounding", researchers have recently proposed "stochastic pruning" \citep{acceleratedSpars2019}, an unbiased pruning method that introduces zero bias on expectation. 

\begin{figure}[h]
 \centering
 \vspace{-4pt}
    \includegraphics[width=0.65\linewidth]{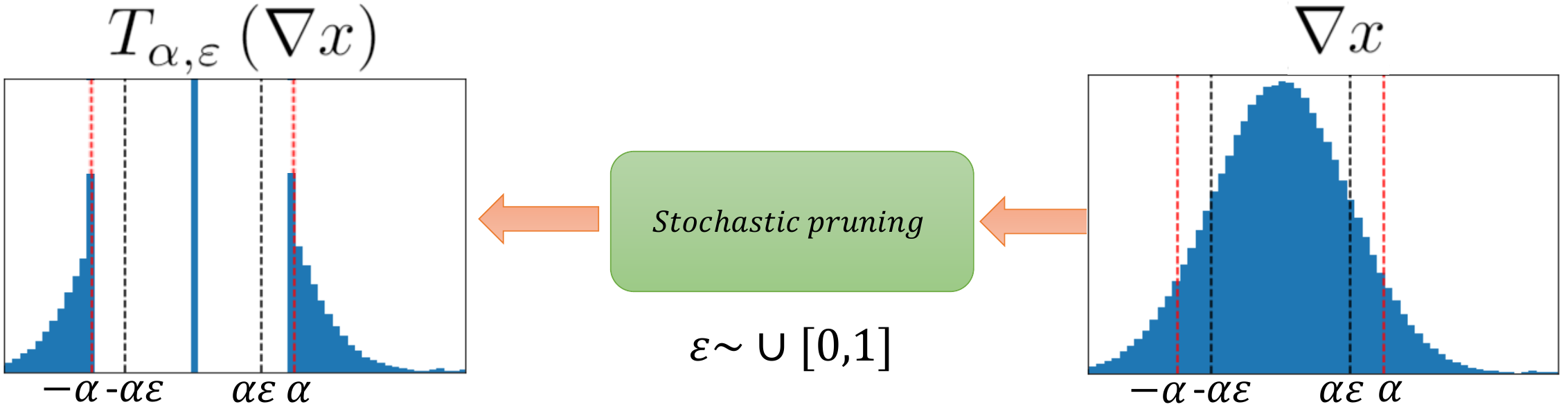}
    \caption{Effect of stochastic pruning on a lognormally distributed tensor. The threshold can be found from \cref{cdf} according to the gradients' distribution. $S$ is the ratio of values mapped to 0, notice that a large fraction of values, all values $\alpha\cdot\varepsilon \leq x \leq \alpha$ are mapped to $\pm \alpha$ --- two values that can have a special encoding, thus reducing the tensors' memory footprint --- details in \cref{sec:encodingAppendix}. }

        \label{fig:linearVsLog}
\end{figure}

Given a threshold $\alpha$, we sample a uniform variable $\varepsilon \sim U[0,1]$ and prune $x$ as follows: 
\begin{equation}
 T_{\alpha,\varepsilon}\left (x\right)=\begin{cases}
x &\quad |x|>\alpha\\
\text{sign}(x) \cdot \alpha & \quad \alpha\cdot\varepsilon \leq |x| \leq \alpha \\
0 & \quad |x| < \alpha\cdot\varepsilon
\end{cases}
\label{StochasticPruning}
\end{equation} 
The method is graphically illustrated in Figure \ref{fig:linearVsLog}. Note that all values in the range $[-\alpha, \alpha]$ need to be mapped to only one of the three possible values (0, $\pm \alpha$). Their increased frequency of occurrence can be used to design a custom encoding method to compress their representation. In \cref{sec:encodingAppendix}, we propose an encoding method with a compression ratio equivalent to quantizing to 4 bits at 80\% sparsity or to only 2 bits at 90\% sparsity. 

\subsection{Problem formulation}
Our goal would be to find an analytical expression for a proper threshold $\alpha$ that induces the desired sparsity $S$ using stochastic pruning. Specifically, let $x$ be a random variable with a known distribution and $S$ a given sparsity level ($0<S<1$). Using stochastic pruning, with $\varepsilon \sim U[0,1]$, we obtain:
\begin{equation}\label{cdf}
    S = \mathbb{E}_\varepsilon \int_{0}^{\alpha\cdot\varepsilon}f(x) \,dx
\end{equation}
\subsection{Analytical derivation of Sparsity}
Understanding the lognormal distribution of the gradients, \cref{cdf} is simply the CDF of a lognormal distribution at $\alpha \cdot \epsilon$, that is: 
\begin{equation}
\label{integral}
S = \mathbb{E}_\varepsilon \left[ \frac{1}{2} + \frac{1}{2}\text{erf}\left(\frac{\ln{(\alpha\cdot\varepsilon) - \mu}}{\sqrt{2}\sigma}\right)\right] \underset{\varepsilon' = \frac{\varepsilon}{e^\mu}}{=} \int_{0}^{\frac{\alpha}{e^\mu}} \left[ \frac{1}{2} + \frac{1}{2}\text{erf}\left(\frac{\ln{(\tau)}}{\sqrt{2}\sigma}\right)\right] \frac{e^\mu}{\alpha} \,d\tau
\end{equation}
The complete solution for this integral can be found in \cref{sec:spToThFull}, resulting in:
\begin{equation}
    \label{spToTh}
    S = \frac{1}{2} + \frac{e^\mu}{2\alpha}\left[ e^{\frac{\sigma^2}{2}}\text{erf}\left(\frac{\sigma}{\sqrt{2}} - \frac{\ln{(\frac{\alpha}{e^\mu})}}{\sqrt{2}\sigma}\right) + \frac{\alpha}{e^\mu} \cdot \text{erf}\left(\frac{\ln{(\frac{\alpha}{e^\mu})}}{\sqrt{2}\sigma}\right) - e^\frac{\sigma^2}{2} \right]
\end{equation}

\cref{spToTh} can easily be solved numerically to find $\alpha$. As shown in \cref{fig:actualMeanStd} the lognormal parameters $\mu$ and $\sigma$ of the gradients' distribution at each layer are pretty stable throughout the training, which allows sampling $\sigma, \mu$ and calculate the threshold $\alpha$ not very frequently. In practice, we perform this procedure only once per epoch while achieving stable sparsity levels throughout the training. Moreover, the computational complexity of solving \cref{spToTh} is negligible (empirically it converges in a few iterations); further details are found in \cref{sec:complexity}.

\subsection{Heterogeneous sparsity allocation}
We show in  \cref{fig:AccVsCosine} that the angle between the tensors before and after the stochastic pruning (measured by the cosine similarity) can serve as an important proxy to the overall validation accuracy achieved. Interestingly, using the cosine similarity, we observed that stochastic pruning takes a different toll from the different layers, i.e. pruning all layers to the same sparsity level damages some of them more than others. This phenomenon can be explained and assessed by analyzing the cosine similarity of a lognormal distribution, where the difference between the layers is the parameters of the distribution. We derive the cosine similarity as another analytical measure and propose an algorithm that better preserves the cosine similarity of some of the layers, by decreasing their sparsity level, while increasing the sparsity level of other layers --- maintaining the overall sparsity budget (mean sparsity of all the layers). Further details can be found in \cref{sec:allocationAppendix}. This allows us to increase the sparsity level while preserving accuracy, results can be seen in \cref{fig:results}.

\section{Experiments}
\label{sec:experiments}

In this section, we evaluate the methods and predictions above, all stemming from the lognormal distribution of the neural gradients, for the two suggested applications: floating point format quantization and stochastic pruning. Experiments, details, and additional results appear in \cref{sec:ap_exp}.

\paragraph{Floating point format.}

In \cref{tab:fpresults} we show the results of different allocations between exponent and mantissa for different FP formats in Cifar100 and ImageNet dataset. We quantize all convolutional layers' gradients, unlike previous methods \citep{Sun2019Hybrid8F} that keep part of them in FP32. All results were achieved using the suggested gradient scaling, where the mean is sampled once every epoch. For all FP formats, the results fit the analysis in \cref{tab:fpFormats}. In contrast, we get sub-optimal FP format if instead we solve \cref{eq:relativeError} with a normal distribution assumption.

\begin{table}[h]
\centering
\caption{Floating-point formats for neural gradient representations. The table reports the validation accuracy of ResNet18  on ImageNet \& Cifar100,  ResNet101 on Cifar100 and SqueezeNet on ImageNet. $E^*$ denotes the optimal exponent bit-width, which minimizes the expected relative error of \cref{eq:relativeFinal}. N/A refers to cases of invalid format (e.g., 1-5-0  is the optimal FP6 format for ImageNet, but FP6 cannot have  $E^* + 1=6$ exponent bits). Note that in all cases, the formats that are predicted by  \cref{eq:relativeFinal} to minimize the relative error also provide the highest accuracies \textbf{(bolded)}. We marked (by $^\dagger$) the results for the optimal format assuming normal distribution for neural gradients in \cref{eq:relativeError}. }
\label{tab:fpresults}
  \vskip 0.1in
  \centering
\begin{tabular}{c|c|c|c|c|c|c|c|c}
    \toprule
\multicolumn{1}{c|}{Dataset}  & \multicolumn{1}{c|}{Model} & \multicolumn{1}{c|}{$\sigma$ Range } &  \multicolumn{1}{c|}{Baseline}  & \multicolumn{1}{c|}{FP} & \multicolumn{1}{c|}{$E^*$} & \multicolumn{1}{c|}{$E^*$+1} & \multicolumn{1}{c|}{$E^*$-1} & \multicolumn{1}{c}{$E^*$-2} \\ 

\midrule

\multicolumn{1}{c|}{\multirow{4}{*}{Cifar100}} & \multicolumn{1}{c|}{\multirow{2}{*}{ResNet18}} & \multicolumn{1}{c|}{\multirow{2}{*}{2.5 - 4.5}} &  \multicolumn{1}{c|}{\multirow{2}{*}{64.9\%}}  & \multicolumn{1}{c|}{FP5} & \multicolumn{1}{c|}{\textbf{64.0\%}} & \multicolumn{1}{c|}{N/A} & \multicolumn{1}{c|}{58.9\%$^\dagger$} & \multicolumn{1}{c}{26.6\%} \\ \cline{5-9} 
\multicolumn{1}{c|}{} & \multicolumn{1}{c|}{} & \multicolumn{1}{c|}{} & \multicolumn{1}{c|}{} & \multicolumn{1}{c|}{FP6} & \multicolumn{1}{c|}{\textbf{64.9\%}} & \multicolumn{1}{c|}{64.6\%} & \multicolumn{1}{c|}{59.7\%$^\dagger$} & \multicolumn{1}{c}{28.6\%} \\ \cline{2-9} 
\multicolumn{1}{c|}{} & \multicolumn{1}{c|}{\multirow{2}{*}{ResNet101}} & \multicolumn{1}{c|}{\multirow{2}{*}{2.5-4.5}} & \multicolumn{1}{c|}{\multirow{2}{*}{71.3\%}} & \multicolumn{1}{c|}{FP5}  & \multicolumn{1}{c|}{\textbf{70.4\%}}  &\multicolumn{1}{c|}{N/A} & \multicolumn{1}{c|}{66.5\%$^\dagger$} & \multicolumn{1}{c}{35\%} \\ \cline{5-9} 
\multicolumn{1}{c|}{} & \multicolumn{1}{c|}{} & \multicolumn{1}{c|}{} & \multicolumn{1}{c|}{} & \multicolumn{1}{c|}{FP6} &
\multicolumn{1}{c|}{\textbf{70.97\%}} &\multicolumn{1}{c|}{70.82\%} & \multicolumn{1}{c|}{67.5\%$^\dagger$} & \multicolumn{1}{c}{42.7\%}  \\
\hline
\multicolumn{1}{c|}{\multirow{4}{*}{ImageNet}} & \multicolumn{1}{c|}{\multirow{2}{*}{ResNet18}} & \multicolumn{1}{c|}{\multirow{2}{*}{3 - 5.5}} & \multicolumn{1}{c|}{\multirow{2}{*}{70.4\%}}  & 
\multicolumn{1}{c|}{FP6} & \multicolumn{1}{c|}{\textbf{70.0\%}} & \multicolumn{1}{c|}{N/A} & \multicolumn{1}{c|}{67.1\%$^\dagger$} & \multicolumn{1}{c}{30.8\%} \\ \cline{5-9} 
\multicolumn{1}{c|}{} & \multicolumn{1}{c|}{} & \multicolumn{1}{c|}{} & \multicolumn{1}{c|}{}  & \multicolumn{1}{c|}{FP7}  & \multicolumn{1}{c|}{\textbf{70.4\%}} & \multicolumn{1}{c|}{70.1\%} & \multicolumn{1}{c|}{66.7\%} & \multicolumn{1}{c}{47.5\%$^\dagger$} \\ \cline{2-9} 
\multicolumn{1}{c|}{} &  \multicolumn{1}{c|}{\multirow{2}{*}{SqueezeNet}}  &  \multicolumn{1}{c|}{\multirow{2}{*}{3 - 5.5}}  &  \multicolumn{1}{c|}{\multirow{2}{*}{58.19 \%}}  &  \multicolumn{1}{c|}{\multirow{1}{*}{FP5}}  &  \multicolumn{1}{c|}{\multirow{1}{*}{\textbf{55.2\%}}}  &  \multicolumn{1}{c|}{\multirow{1}{*}{N/A}} &  \multicolumn{1}{c|}{\multirow{1}{*}{47.3\%$^\dagger$ }} &  \multicolumn{1}{c}{\multirow{1}{*}{33.2\%}}\\ \cline{5-9} \multicolumn{1}{c|}{} & \multicolumn{1}{c|}{} & \multicolumn{1}{c|}{} & \multicolumn{1}{c|}{} &  \multicolumn{1}{c|}{\multirow{1}{*}{FP6}} &  \multicolumn{1}{c|}{\multirow{1}{*}{\textbf{57.8\%}}}  &  \multicolumn{1}{c|}{\multirow{1}{*}{N/A}} &  \multicolumn{1}{c|}{\multirow{1}{*}{56.1\%$^\dagger$ }} &  \multicolumn{1}{c}{\multirow{1}{*}{54.3\%}} \\  \bottomrule

\end{tabular}
\end{table}

\paragraph{Per-layer gradient scaling}
In \cref{tab:lossScale} we compare the suggested gradient scaling with static and dynamic \citep{Micikevicius2017MixedPT} global loss scaling. We clipped the values to maximum/minimum FP representation for our method and the 'static' method to avoid overflow/underflow, respectively. On the other hand, the dynamic method \citep{Micikevicius2017MixedPT} clips the values to adjust the scale in response to overflow/underflow in the weights updates. Our suggested layer-wise gradient scaling achieves better results,   highlighting that one global loss scale for the entire network might be too restrictive in practice. In \cref{fig:lossScaleAp}, we show the variability of the suggested gradient scale across different layers. Lastly, to verify that gradient quantization is the main bottleneck, we also quantized the weights and activations to INT4 and found a minor degradation in accuracy ($0.3\%$).

\begin{table}[]
   \renewcommand*{\arraystretch}{1.3}

\centering
\caption{Comparison of the suggested gradient scaling against different global fixed loss scaling and global dynamic loss scaling \citep{Micikevicius2017MixedPT} in ResNet18, on the ImageNet dataset for the optimal format in FP4 (1-3-0).}
\label{tab:lossScale}
\resizebox{\columnwidth}{!}{%
\begin{tabular}{c|c|c|c|c|c|c|c|c|c}
\toprule
\multirow{3}{*}{Baseline} & \multirow{3}{*}{\begin{tabular}[c]{@{}c@{}}Gradient\\ scaling\\ (Ours)\end{tabular}} & \multirow{3}{*}{\begin{tabular}[c]{@{}c@{}}Dynamic\\ Global\end{tabular}} & \multicolumn{7}{c}{\multirow{2}{*}{Static (Global)}} \\
 &  &  & \multicolumn{7}{c}{} \\ \cline{4-10} 
 &  &  &  \multicolumn{1}{c|}{$2^{13}$ } & \multicolumn{1}{c|}{$2^{14}$ } &
\multicolumn{1}{c|}{$2^{15}$ } &
\multicolumn{1}{c|}{$2^{16}$ } &
\multicolumn{1}{c|}{$2^{17}$ } &
\multicolumn{1}{c|}{$2^{18}$ } &
\multicolumn{1}{c}{$2^{19}$ }  \\ \midrule

\multicolumn{1}{c|}{\multirow{1}{*}{70.4\%}} &  \multicolumn{1}{c|}{\multirow{1}{*}{\textbf{64.8\%}}}  &\multicolumn{1}{c|}{3\%} &\multicolumn{1}{c|}{26.7\%} & \multicolumn{1}{c|}{26.4\%} & \multicolumn{1}{c|}{38.7\%} & \multicolumn{1}{c|}{41.5\%} & \multicolumn{1}{c|}{42.2\%}   & \multicolumn{1}{c|}{54.9\%} 
 & \multicolumn{1}{c}{53.9\%}  \\ \bottomrule
\end{tabular}}
\end{table}

\paragraph{Stochastic pruning.}
\begin{figure}[h]

\begin{subfigure}{.5\textwidth}
  \centering
  \includegraphics[width=\linewidth]{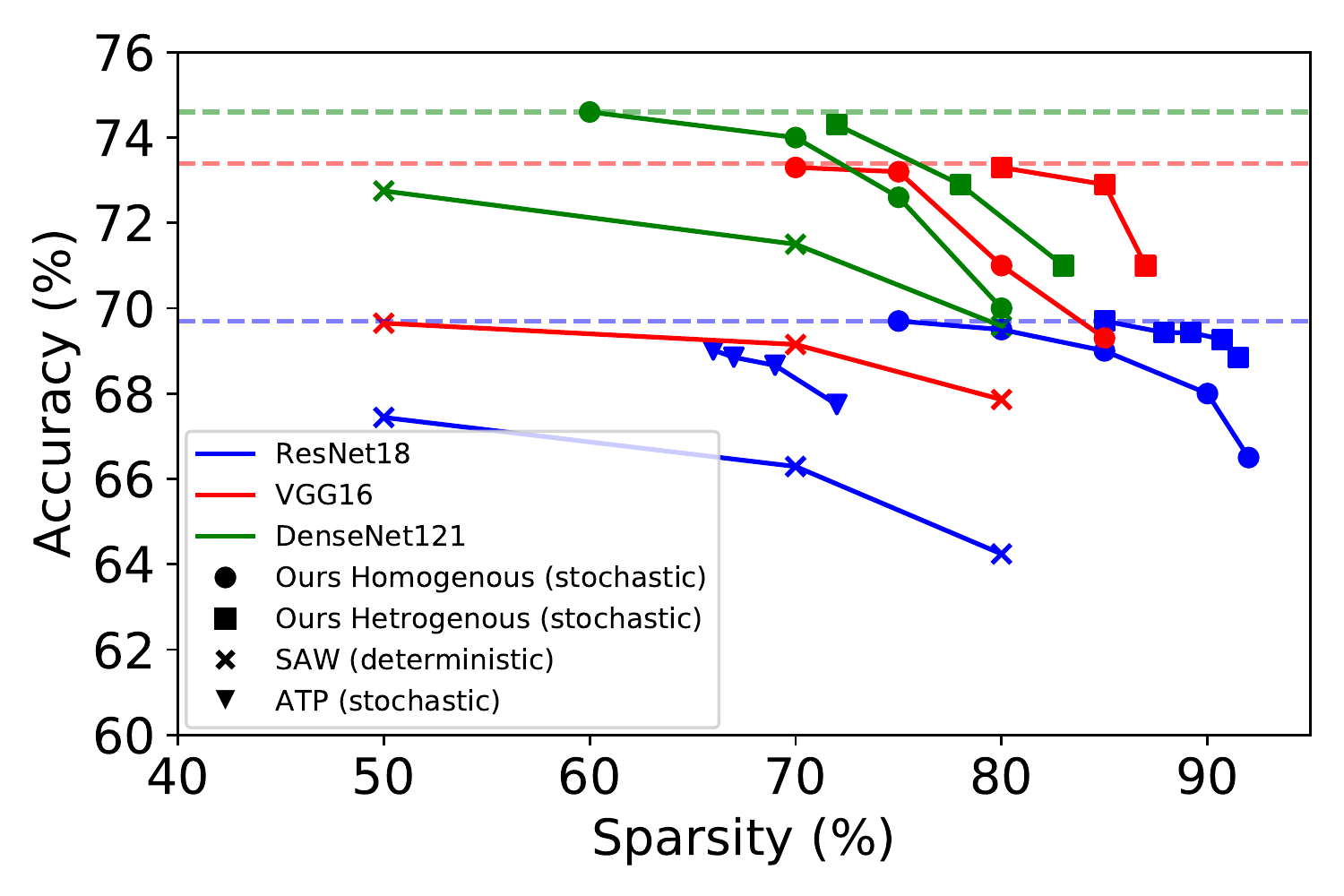}  

  \label{fig:results}
\end{subfigure}
\begin{subfigure}{.5\textwidth}
  \centering
  \includegraphics[width=0.95\linewidth]{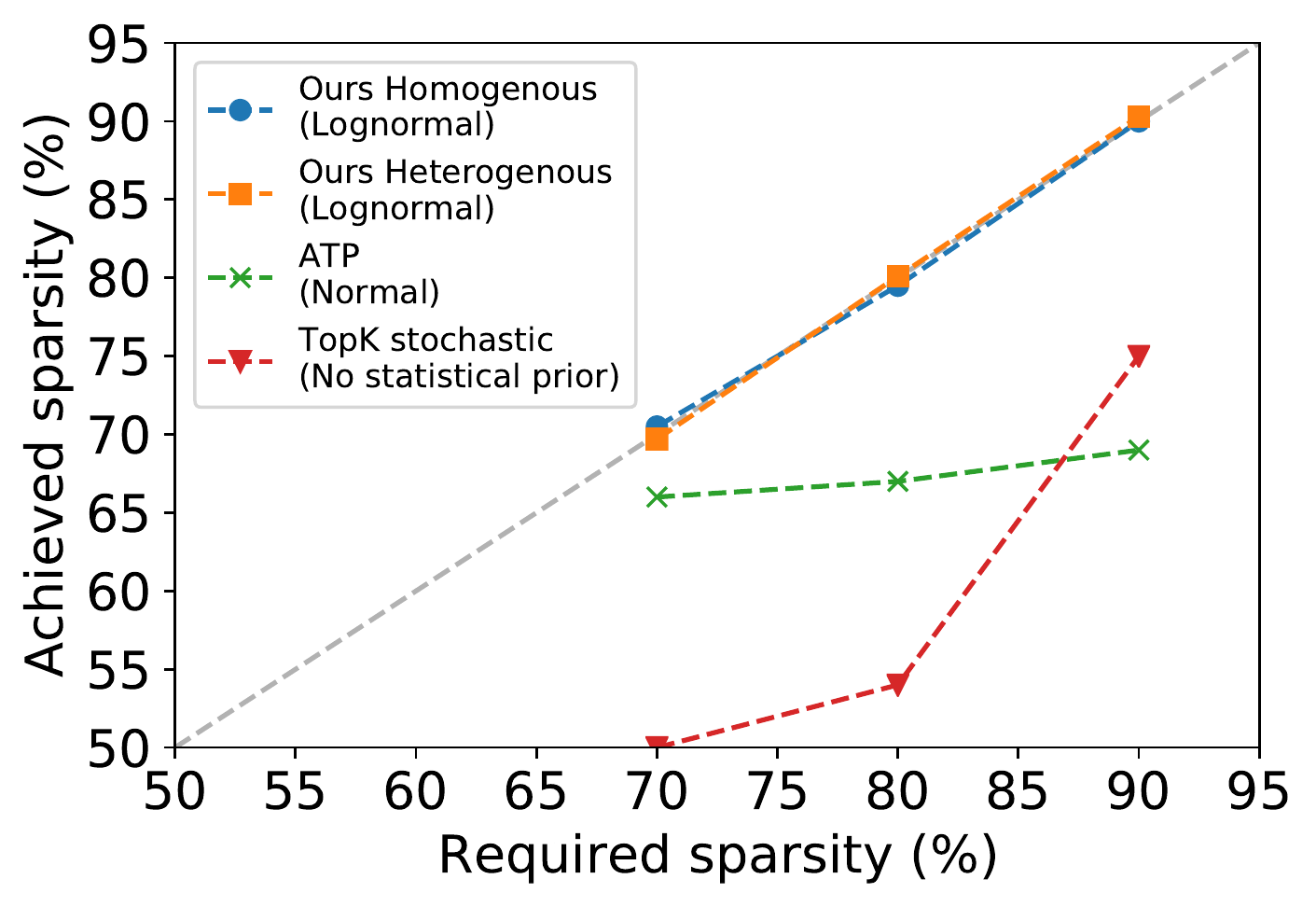}  
 
  \label{fig:requiredVsAchieved}
\end{subfigure}
\caption{
\textbf{(Left)} Comparison of the homogeneous and heterogeneous stochastic pruning methods against pruning with SAW \citep{SWAT2020}, which uses the "top-k" method and ATP \citep{acceleratedSpars2019}, which also uses stochastic pruning, but assumes the neural gradients are normally distributed. These are compared on 3 different architectures (ResNet18, VGG16 and DenseNet121). Our proposed methods achieve higher sparsity while maintaining baseline validation accuracy. \textbf{(Right)}  Comparison of the required and achieved sparsity of our method, ATP \citep{acceleratedSpars2019} and applying "top-k" followed by stochastic pruning. The validation accuracy for our method is never less than the accuracy of the other methods for the same achieved sparsity. Notice the large deviation in the other methods between the achieved and required sparsity. This emphasizes the importance of using the correct distribution in order to both enjoy the improved performance of stochastic pruning over regular "top-k" and maintain the ability to fully control the achieved sparsity. Additional details and results are in \cref{sec:ap_exp}}
\label{fig:Comparison}
\vspace{-1mm}
\end{figure}
In \cref{fig:results} we compare the proposed methods for homogeneous and heterogeneous stochastic pruning against SAW \citep{SWAT2020} that uses "top-k" pruning and ATP  \citep{acceleratedSpars2019} that also uses stochastic pruning but assumes the gradients are normally distributed. Notice that stochastic pruning outperforms the non-stochastic top-k, and that the heterogeneous version surpasses the homogeneous one. The validation accuracy during training for different sparsity levels and different datasets can be found in \cref{fig:ImagenetTrain}. In \cref{fig:requiredVsAchieved} we demonstrate our methods' ability to produce the required sparsity level, for both the homogeneous and heterogeneous versions. In contrast, the sparsity is not accurate for the baseline methods: (1) finding the threshold using top-k and then applying stochastic pruning, and (2) using ATP \citep{acceleratedSpars2019}, which assumes a normal distribution. In \cref{fig:deviation} we see how the sparsity inaccuracy occurs at all layers, and in \cref{fig:requiredVsachievedDistr} we see how other distributions (not lognormal) cause an inaccuracy. This strengthens the importance of using the correct distribution of the neural gradients in \cref{cdf}.

\section{Summary}
We evaluated the distribution of neural gradients and showed they can be well-approximated as a lognormal distribution. We use this distribution to analytically derive accurate measures (e.g., sparsity and local distortion metrics), useful for the following applications.

\textbf{Quantization.} We found the optimal bit allocation to the mantissa and exponent for a floating-point gradient representation, explaining prior results for FP8 and paving the way towards lower accuracy representations. We suggest using a per-layer gradient scale and find its optimal value, preventing under/over-flow in scenarios that challenged prior methods or required an intensive parameter search. Combining both methods, we trained using low precision neural gradients on ImageNet and achieved, for the first time, no noticeable validation accuracy degradation with FP7 and FP6.

\textbf{Pruning.} We can use stochastic pruning to prune the neural gradients during training precisely to a predetermined sparsity level, with minimal overhead computation. We show that this form of stochastic pruning is superior to deterministic pruning. Specifically, we have achieved up to 80\% gradient sparsity without hurting validation accuracy (ResNet18 on ImageNet) using a homogeneous sparsity level for all layers. We also show that the uniform sparsity method is sub-optimal with respect to an analytical error measure we derive (the cosine similarity), and suggest allocating different sparsity levels to the different layers to preserve it better. We suggest and test an algorithm for this allocation - allowing for more aggressive overall pruning, achieving 85\% gradient sparsity while preserving baseline accuracy, and reaching nearly 90\% with less than 0.3\% accuracy degradation.

\bibliography{iclr2021_conference}

\begin{thebibliography}{32}
\providecommand{\natexlab}[1]{#1}
\providecommand{\url}[1]{\texttt{#1}}
\expandafter\ifx\csname urlstyle\endcsname\relax
  \providecommand{\doi}[1]{doi: #1}\else
  \providecommand{\doi}{doi: \begingroup \urlstyle{rm}\Url}\fi

\bibitem[{Aamir Raihan} \& {Aamodt}(2020){Aamir Raihan} and {Aamodt}]{SWAT2020}
Md~{Aamir Raihan} and Tor~M. {Aamodt}.
\newblock Sparse weight activation training.
\newblock \emph{arXiv:2001.01969}, 2020.

\bibitem[Banner et~al.(2018{\natexlab{a}})Banner, Hubara, Hoffer, and
  Soudry]{Banner2018ScalableMF}
Ron Banner, Itay Hubara, Elad Hoffer, and Daniel Soudry.
\newblock Scalable methods for 8-bit training of neural networks.
\newblock In \emph{NeurIPS}, 2018{\natexlab{a}}.

\bibitem[Banner et~al.(2018{\natexlab{b}})Banner, Nahshan, Hoffer, and
  Soudry]{banner2018post}
Ron Banner, Yury Nahshan, Elad Hoffer, and Daniel Soudry.
\newblock Post-training 4-bit quantization of convolution networks for
  rapid-deployment.
\newblock \emph{arXiv preprint arXiv:1810.05723}, 2018{\natexlab{b}}.
\newblock URL \url{http://arxiv.org/abs/1810.05723}.

\bibitem[Baskin et~al.(2018)Baskin, Liss, Chai, Zheltonozhskii, Schwartz,
  Girayes, Mendelson, and Bronstein]{baskin2018nice}
Chaim Baskin, Natan Liss, Yoav Chai, Evgenii Zheltonozhskii, Eli Schwartz, Raja
  Girayes, Avi Mendelson, and Alexander~M Bronstein.
\newblock Nice: Noise injection and clamping estimation for neural network
  quantization.
\newblock \emph{arXiv preprint arXiv:1810.00162}, 2018.
\newblock URL \url{http://arxiv.org/abs/1810.00162}.

\bibitem[Bernstein et~al.(2018)Bernstein, Wang, Azizzadenesheli, and
  Anandkumar]{Bernstein2018signSGDCO}
Jeremy Bernstein, Yu-Xiang Wang, Kamyar Azizzadenesheli, and Anima Anandkumar.
\newblock signsgd: compressed optimisation for non-convex problems.
\newblock In \emph{ICML}, 2018.

\bibitem[Biau \& Mason(2015)Biau and Mason]{biau2015high}
Gerard Biau and David~M Mason.
\newblock High-dimensional norms.
\newblock In \emph{Mathematical Statistics and Limit Theorems}, pp.\  21--40.
  Springer, 2015.

\bibitem[Cambier et~al.(2020)Cambier, Bhiwandiwalla, Gong, Nekuii, Elibol, and
  Tang]{Cambier2020ShiftedAS}
L{\'e}opold Cambier, Anahita Bhiwandiwalla, Ting Gong, Mehran Nekuii, Oguz~H.
  Elibol, and Hanlin Tang.
\newblock Shifted and squeezed 8-bit floating point format for low-precision
  training of deep neural networks.
\newblock \emph{ArXiv}, abs/2001.05674, 2020.

\bibitem[Choi et~al.(2018{\natexlab{a}})Choi, Chuang, Wang, Venkataramani,
  Srinivasan, and Gopalakrishnan]{choi2018quantization}
Jungwook Choi, Pierce I-Jen Chuang, Zhuo Wang, Swagath Venkataramani,
  Vijayalakshmi Srinivasan, and Kailash Gopalakrishnan.
\newblock Bridging the accuracy gap for 2-bit quantized neural networks (qnn),
  2018{\natexlab{a}}.

\bibitem[Choi et~al.(2018{\natexlab{b}})Choi, Wang, Venkataramani, Chuang,
  Srinivasan, and Gopalakrishnan]{Choi2018PACTPC}
Jungwook Choi, Zhuo Wang, Swagath Venkataramani, Pierce I-Jen Chuang,
  Vijayalakshmi Srinivasan, and Kailash Gopalakrishnan.
\newblock Pact: Parameterized clipping activation for quantized neural
  networks.
\newblock \emph{ArXiv}, abs/1805.06085, 2018{\natexlab{b}}.

\bibitem[Deng et~al.(2009)Deng, Dong, Socher, Li, Li, and
  Fei-Fei]{deng2009imagenet}
Jia Deng, Wei Dong, Richard Socher, Li-Jia Li, Kai Li, and Li~Fei-Fei.
\newblock Imagenet: A large-scale hierarchical image database.
\newblock In \emph{2009 IEEE conference on computer vision and pattern
  recognition}, pp.\  248--255. Ieee, 2009.

\bibitem[Fang et~al.(2020)Fang, Shafiee, Abdel-Aziz, Thorsley, Georgiadis, and
  Hassoun]{Fang2020PostTrainingPL}
Jun Fang, Ali Shafiee, Hamzah Abdel-Aziz, David Thorsley, Georgios Georgiadis,
  and Joseph Hassoun.
\newblock Post-training piecewise linear quantization for deep neural networks.
\newblock \emph{arXiv: Computer Vision and Pattern Recognition}, 2020.

\bibitem[Frankle \& Carbin(2018)Frankle and Carbin]{Frankle2018TheLT}
Jonathan Frankle and Michael Carbin.
\newblock The lottery ticket hypothesis: Finding sparse, trainable neural
  networks.
\newblock In \emph{ICLR}, 2018.

\bibitem[He et~al.(2016)He, Zhang, Ren, and Sun]{resnet}
Kaiming He, Xiangyu Zhang, Shaoqing Ren, and Jian Sun.
\newblock Deep residual learning for image recognition.
\newblock \emph{2016 IEEE Conference on Computer Vision and Pattern Recognition
  (CVPR)}, pp.\  770--778, 2016.

\bibitem[Henry et~al.(2019)Henry, Tang, and Heinecke]{henry2019leveraging}
Greg Henry, Ping Tak~Peter Tang, and Alexander Heinecke.
\newblock Leveraging the bfloat16 artificial intelligence datatype for
  higher-precision computations.
\newblock In \emph{2019 IEEE 26th Symposium on Computer Arithmetic (ARITH)},
  pp.\  69--76. IEEE, 2019.

\bibitem[Hoare(1961)]{quickselect}
C.~A.~R. Hoare.
\newblock Algorithm 65: Find.
\newblock \emph{Commun. ACM}, 4\penalty0 (7):\penalty0 321–322, July 1961.
\newblock ISSN 0001-0782.
\newblock \doi{10.1145/366622.366647}.
\newblock URL \url{https://doi.org/10.1145/366622.366647}.

\bibitem[Huang et~al.(2017)Huang, Liu, and Weinberger]{densenet}
Gao Huang, Zhuang Liu, and Kilian~Q. Weinberger.
\newblock Densely connected convolutional networks.
\newblock \emph{2017 IEEE Conference on Computer Vision and Pattern Recognition
  (CVPR)}, pp.\  2261--2269, 2017.

\bibitem[Jain et~al.(2019)Jain, Gural, Wu, and Dick]{jain2019quantization}
Sambhav~R. Jain, Albert Gural, Michael Wu, and Chris~H. Dick.
\newblock Trained quantization thresholds for accurate and efficient
  fixed-point inference of deep neural networks, 2019.

\bibitem[Kalamkar et~al.(2019)Kalamkar, Mudigere, Mellempudi, Das, Banerjee,
  Avancha, Vooturi, Jammalamadaka, Huang, Yuen, et~al.]{kalamkar2019study}
Dhiraj Kalamkar, Dheevatsa Mudigere, Naveen Mellempudi, Dipankar Das, Kunal
  Banerjee, Sasikanth Avancha, Dharma~Teja Vooturi, Nataraj Jammalamadaka,
  Jianyu Huang, Hector Yuen, et~al.
\newblock A study of bfloat16 for deep learning training.
\newblock \emph{arXiv preprint arXiv:1905.12322}, 2019.

\bibitem[Krishnamoorthi(2018)]{krishnamoorthi2018quantizing}
Raghuraman Krishnamoorthi.
\newblock Quantizing deep convolutional networks for efficient inference: A
  whitepaper, 2018.

\bibitem[Louizos et~al.(2017)Louizos, Welling, and
  Kingma]{Louizos2017LearningSN}
Christos Louizos, Max Welling, and Diederik~P. Kingma.
\newblock Learning sparse neural networks through l0 regularization.
\newblock \emph{ArXiv}, abs/1712.01312, 2017.

\bibitem[Micikevicius et~al.(2017)Micikevicius, Narang, Alben, Diamos, Elsen,
  Garc{\'i}a, Ginsburg, Houston, Kuchaiev, Venkatesh, and
  Wu]{Micikevicius2017MixedPT}
Paulius Micikevicius, Sharan Narang, Jonah Alben, Gregory~Frederick Diamos,
  Erich Elsen, David Garc{\'i}a, Boris Ginsburg, Michael Houston, Oleksii
  Kuchaiev, Ganesh Venkatesh, and Hao Wu.
\newblock Mixed precision training.
\newblock \emph{ArXiv}, abs/1710.03740, 2017.

\bibitem[Nahshan et~al.(2019)Nahshan, Chmiel, Baskin, Zheltonozhskii, Banner,
  Bronstein, and Mendelson]{Nahshan2019LossAP}
Yury Nahshan, Brian Chmiel, Chaim Baskin, Evgenii Zheltonozhskii, Ron Banner,
  Alex~M. Bronstein, and Avi Mendelson.
\newblock Loss aware post-training quantization.
\newblock \emph{ArXiv}, abs/1911.07190, 2019.

\bibitem[Simonyan \& Zisserman(2015)Simonyan and Zisserman]{vgg}
Karen Simonyan and Andrew Zisserman.
\newblock Very deep convolutional networks for large-scale image recognition.
\newblock \emph{CoRR}, abs/1409.1556, 2015.

\bibitem[Smirnov(1948)]{KS_Test}
N.~Smirnov.
\newblock Table for estimating the goodness of fit of empirical distributions.
\newblock \emph{Ann. Math. Statist. 19 (1948), no. 2, 279--281.}, 1948.
\newblock URL \url{https://projecteuclid.org/euclid.aoms/1177730256}.

\bibitem[Sun et~al.(2019)Sun, Choi, Chen, Wang, Venkataramani, Srinivasan, Cui,
  Zhang, and Gopalakrishnan]{Sun2019Hybrid8F}
Xiao Sun, Jungwook Choi, Chia-Yu Chen, Naigang Wang, Swagath Venkataramani,
  Vijayalakshmi Srinivasan, Xiaodong Cui, Wei Zhang, and Kailash
  Gopalakrishnan.
\newblock Hybrid 8-bit floating point (hfp8) training and inference for deep
  neural networks.
\newblock In \emph{NeurIPS}, 2019.

\bibitem[{Sun} et~al.(2017){Sun}, {Ren}, {Ma}, and {Wang}]{meProp2017}
Xu~{Sun}, Xuancheng {Ren}, Shuming {Ma}, and Houfeng {Wang}.
\newblock meprop: Sparsified back propagation for accelerated deep learning
  with reduced overfitting.
\newblock \emph{arXiv:1706.06197}, 2017.

\bibitem[Vaswani et~al.(2017)Vaswani, Shazeer, Parmar, Uszkoreit, Jones, Gomez,
  Kaiser, and Polosukhin]{Vaswani2017AttentionIA}
Ashish Vaswani, Noam Shazeer, Niki Parmar, Jakob Uszkoreit, Llion Jones,
  Aidan~N. Gomez, Lukasz Kaiser, and Illia Polosukhin.
\newblock Attention is all you need.
\newblock \emph{ArXiv}, abs/1706.03762, 2017.

\bibitem[Wang et~al.(2018)Wang, Choi, Brand, Chen, and
  Gopalakrishnan]{wang2018quantization}
Naigang Wang, Jungwook Choi, Daniel Brand, Chia-Yu Chen, and Kailash
  Gopalakrishnan.
\newblock Training deep neural networks with 8-bit floating point numbers,
  2018.

\bibitem[Widrow \& Koll{\'a}r(2008)Widrow and
  Koll{\'a}r]{widrow2008quantization}
Bernard Widrow and Istv{\'a} Koll{\'a}r.
\newblock Quantization noise: roundoff error in digital computation, signal
  processing, control, and communications, 2008, 2008.

\bibitem[Wiedemann et~al.(2020)Wiedemann, Mehari, Kepp, and
  Samek]{wiedemann2020ditheredBackprop}
Simon Wiedemann, Temesgen Mehari, Kevin Kepp, and Wojciech Samek.
\newblock Dithered backprop: A sparse and quantized backpropagation algorithm
  for more efficient deep neural network training, 2020.

\bibitem[Wu et~al.(2018)Wu, Li, Chen, and Shi]{wu2018quantization}
Shuang Wu, Guoqi Li, Feng Chen, and Luping Shi.
\newblock Training and inference with integers in deep neural networks, 2018.

\bibitem[{Ye} et~al.(2019){Ye}, {Yang}, {Dai}, {Chen}, and
  {Zhao}]{acceleratedSpars2019}
Xucheng {Ye}, Jianlei {Yang}, Pengcheng {Dai}, Yiran {Chen}, and Weisheng
  {Zhao}.
\newblock Accelerating cnn training by sparsifying activation gradients.
\newblock \emph{arXiv:1908.00173}, 2019.

\end{thebibliography}
\bibliographystyle{iclr2021_conference}


\newpage

\appendix{}

\renewcommand\thefigure{\thesection.\arabic{figure}} 
\renewcommand\thetable{\thesection.\arabic{table}} 
\renewcommand\theequation{\thesection.\arabic{equation}}  
\setcounter{figure}{0}  
\setcounter{table}{0}
\setcounter{equation}{0}

\crefalias{section}{appsec}
\crefalias{subsection}{appsec}
\crefalias{subsubsection}{appsec}



\section{Supplementary Material}

\subsection{Neural gradients distribution --- discussion}
\label{secAp: dist discussion}

In order to analyze how well neural gradients fit to a lognormal distribution, we measure in \cref{tab:App_KS-test} the Kolmogorov-Smirnov test \citep{KS_Test} with different distributions in different models and datasets, similar to \cref{tab:KS-test}. Notice that lognormal and loglaplace distributions achieve much higher fit than the other distribution, with a small statistical advantage of the lognormal distribution.

\begin{figure}[h]
  \centering
  \includegraphics[width=0.5\linewidth]{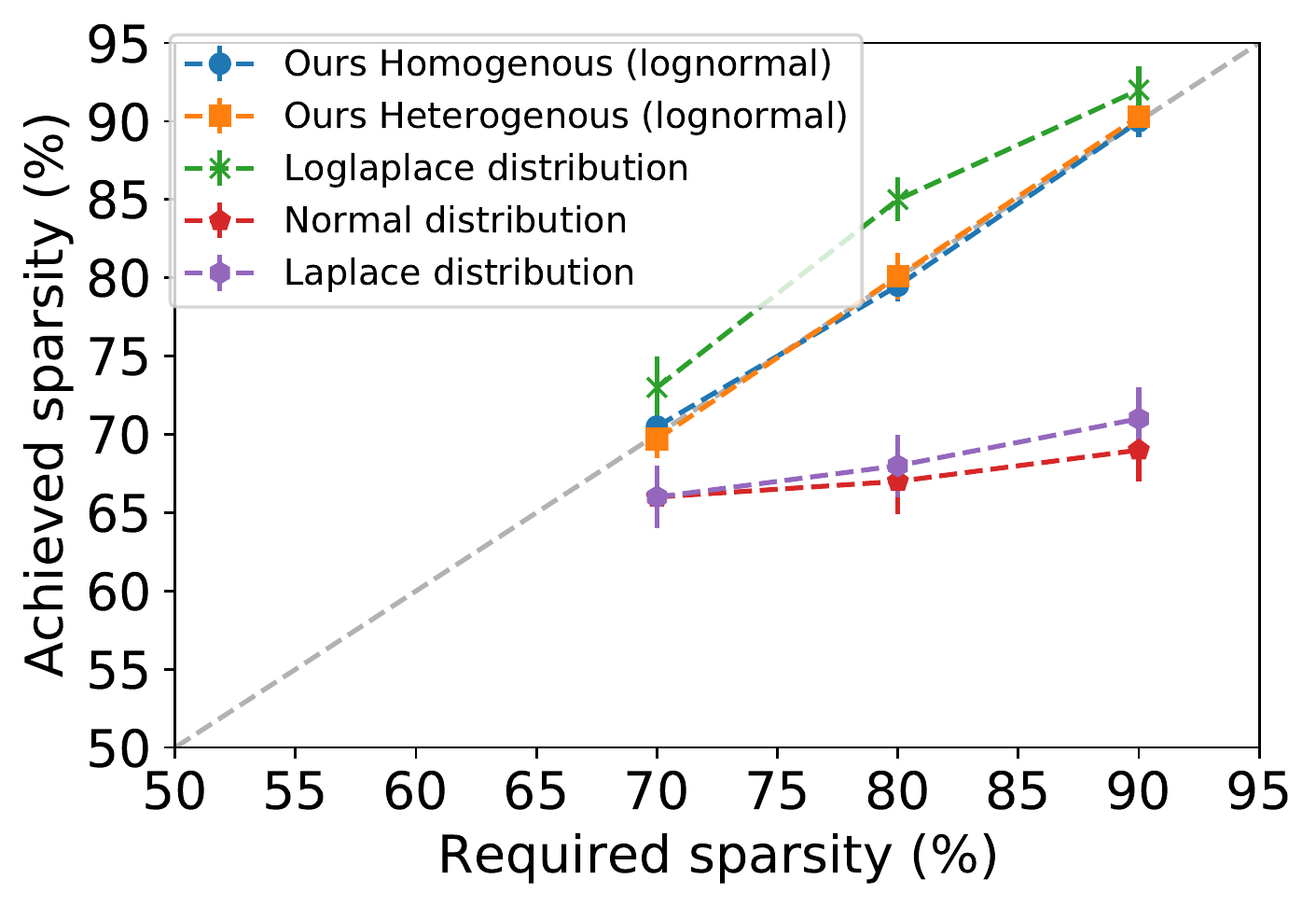}  
  \caption{}
\caption{Comparison of the required and achieved sparsity of stochastic pruning with different distribution assumptions.  Notice that lognormal distribution achieves the minimum deviation from the requires sparsity. This emphasizes the goodness of fit of the neural gradients to a lognormal distribution.}
\label{fig:requiredVsachievedDistr}
\end{figure}

In order to emphasize of the choice of the lognormal distribution as the best fit, we show in \cref{fig:requiredVsachievedDistr} the same experiment described in \cref{fig:requiredVsAchieved} with additional distributions, i.e solve equation \cref{cdf} with different distribution assumptions and check the achieved sparsity in that case. We can empirically notice that the lognormal distribution is a better fit for the neural gradients than the other distributions.
\begin{table}[h]
\vspace{-3mm}
\centering
\caption{Mean ($\pm$ std) over all layers over time of Kolmogorov-Smirnov test on different models and datasets. Notice the lognormal distribution get the higher fit across all models }
\resizebox{\columnwidth}{!}{%
\begin{tabular}{c||ccccccc}
\toprule
Model-Dataset & Laplace & Normal & Uniform & Cauchy &  Loglaplace & Lognormal\\ 
\hline
 BERT-CoLA & 0.46$\pm$0.02 & 0.46$\pm$0.02 & 0.96$\pm$0.02 & 0.46$\pm$0.01 & 0.07$\pm$0.012 & \textbf{0.05$\pm$0.002} \\
  BERT- MRPC & 0.41$\pm$0.03 & 0.39$\pm$0.04 & 0.87$\pm$0.04 &  0.41$\pm$0.03 & 0.07$\pm$0.015 & \textbf{0.04$\pm$0.002} \\
  ResNet18-ImageNet & 0.32$\pm$0.1 & 0.38$\pm$0.1 & 0.77$\pm$0.1 &  0.37$\pm$0.15 & 0.05$\pm$0.001 & \textbf{0.02$\pm$0.002} \\
MobileNetV2-ImageNet & 0.22$\pm$0.09 & 0.22$\pm$0.09 & 0.69$\pm$0.12 &  0.26$\pm$0.06 & 0.09$\pm$0.02 & \textbf{0.07$\pm$0.003} \\
VGG16-ImageNet & 0.37$\pm$0.09 & 0.35$\pm$0.08 & 0.8$\pm$0.1 &  0.39$\pm$0.08 & 0.08$\pm$0.032 & \textbf{0.06$\pm$0.002} \\
DenseNet121-ImageNet & 0.32$\pm$0.1 & 0.33$\pm$0.1 & 0.76$\pm$0.06 &  0.38$\pm$0.09 & 0.075$\pm$0.01 & \textbf{0.05$\pm$0.001} \\
\bottomrule
\end{tabular}
}
\label{tab:App_KS-test}
\end{table}

\begin{figure}[h]
\begin{subfigure}{.24\textwidth}
  \centering
  \includegraphics[width=\linewidth]{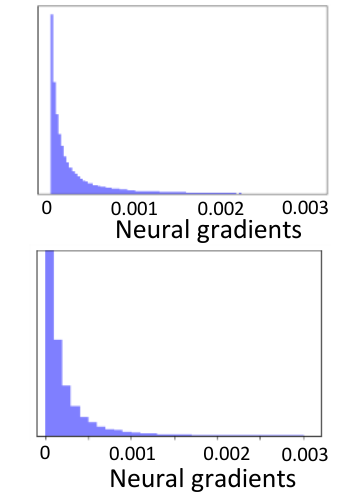}  
  \caption{}
 \end{subfigure}
\begin{subfigure}{.24\textwidth}
  \centering
  \includegraphics[width=\linewidth]{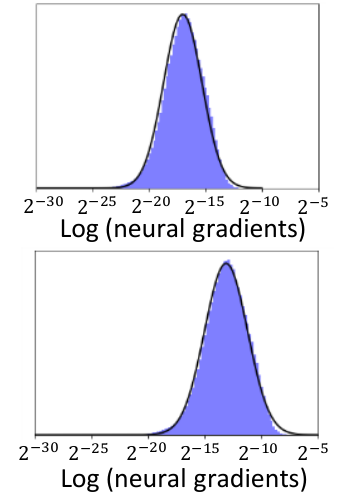}  
  \caption{}
\end{subfigure}
\begin{subfigure}{.24\textwidth}
  \centering
  \includegraphics[width=\linewidth]{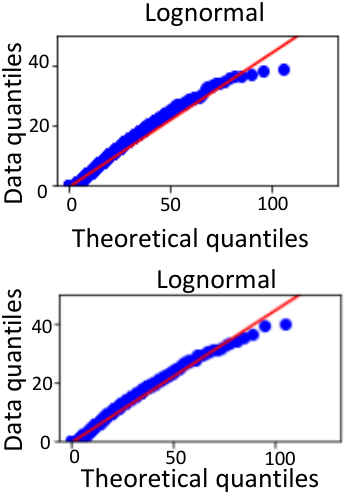}  
  \caption{}
\end{subfigure}
\begin{subfigure}{.24\textwidth}
  \centering
  \includegraphics[width=\linewidth]{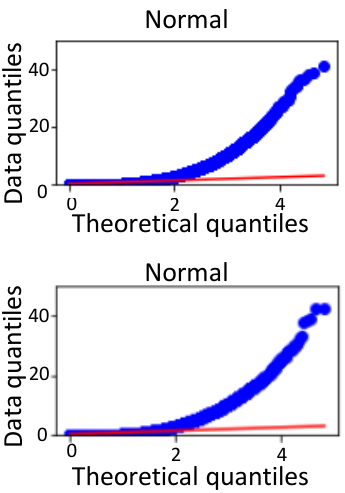}  
  \caption{}
\end{subfigure}
\caption{Identifying the distribution of neural gradients (normal vs. lognormal) for additional layers in ResNet18 - ImageNet dataset, similar to \cref{fig:normalVsLognormal}.}
\label{fig:normalVsLognormalApp}

\end{figure}

\subsection{Neural gradients before and after batch-norm}
\label{sec:mixtureAppendix}

A neural network is a composition of different blocks that include linear and non-linear functions. In CNNs the most common block contains the trio Conv-BN-ReLU. Consider a CNN with L layers, the output, $\mathcal{A}_{i;3}$, of layer $ i \in \{1, ..., L\}$ is:

\begin{align}
\begin{split}
\textbf{[Conv]}\, \mathcal{A}_{i;1} = W_i \times \mathcal{A}_{i-1;3}
& \quad \textbf{[BN]}\, \mathcal{A}_{i;2} = \mathrm{BN}_{\beta,\gamma}(\mathcal{A}_{i;1})  
\quad \textbf{[ReLU]}\, \mathcal{A}_{i;3} = \max(0,\mathcal{A}_{i;2})\,,
\end{split}
\end{align}

And the corresponding gradients:

\begin{equation}
    \label{eq:gradRelu}
    \textbf{[ReLU]} \quad \nabla \mathcal{A}_{i;2} = \nabla \mathcal{A}_{i;3}\cdot \delta({A}_{i;2} > 0)
\end{equation}
\begin{equation}
\begin{split}
    \label{eq:gradBN}
  \textbf{[BN]} \quad \nabla \mathcal{A}_{i;1} &= \frac{\gamma_i}{\sqrt{\sigma^2 + \epsilon}}\left[\nabla \mathcal{A}_{i;2} -  \frac{1}{N}\left(\nabla \gamma \left( \frac{\mathcal{A}_{i;1} - \mu_i}{\sqrt{\sigma_i^2 + \epsilon}} \right) + \nabla \beta \right) \right] \\
  \nabla \beta &= \sum(\nabla \mathcal{A}_{i;2}) \quad \nabla \gamma = \sum\left(\nabla \mathcal{A}_{i;2} \times \frac{\mathcal{A}_{i;1} - \mu_i}{\sqrt{\sigma_i^2 + \epsilon}}\right)
  \end{split}
\end{equation}
\begin{equation}
   \textbf{[Conv]} \quad \nabla \mathcal{A}_{i-1;3} = \nabla \mathcal{A}_{i;1} \times W_i^T \quad \nabla W_i = \nabla \mathcal{A}_{i;1} \times \mathcal{A}_{i-1;3} 
\end{equation}

where $\times$ defines a convolution operation, $W_i$ is the weights matrix, and $\mathrm{BN}$ is a batch-norm operation. The corresponding gradients are $\nabla\mathcal{A}_{i;1-3}, \nabla W_i, \nabla \beta$ and $\nabla \gamma$

The distribution of $\nabla \mathcal{A}_{i-1;3}$ fit to lognormal distribution as shown in the main paper.
The distribution of $\nabla \mathcal{A}_{i;1}$ can be approximated as a bi-modal lognormal distribution. The two modes stem from two different components of $\nabla \mathcal{A}_{i;2}$: the left mode originates from the zero-valued elements (that are abundant because of the ReLU) and the right mode from the rest of $\nabla \mathcal{A}_{i;2}$. We split the elements of $\nabla \mathcal{A}_{i;1}$ according to the value of the same element in $\nabla \mathcal{A}_{i;2}$ (zero or non-zero), then apply in each of them Kolmogorov-Smirnov test. 

\begin{figure}[h]
  
  \centering
  \includegraphics[width=14cm,height=9cm]{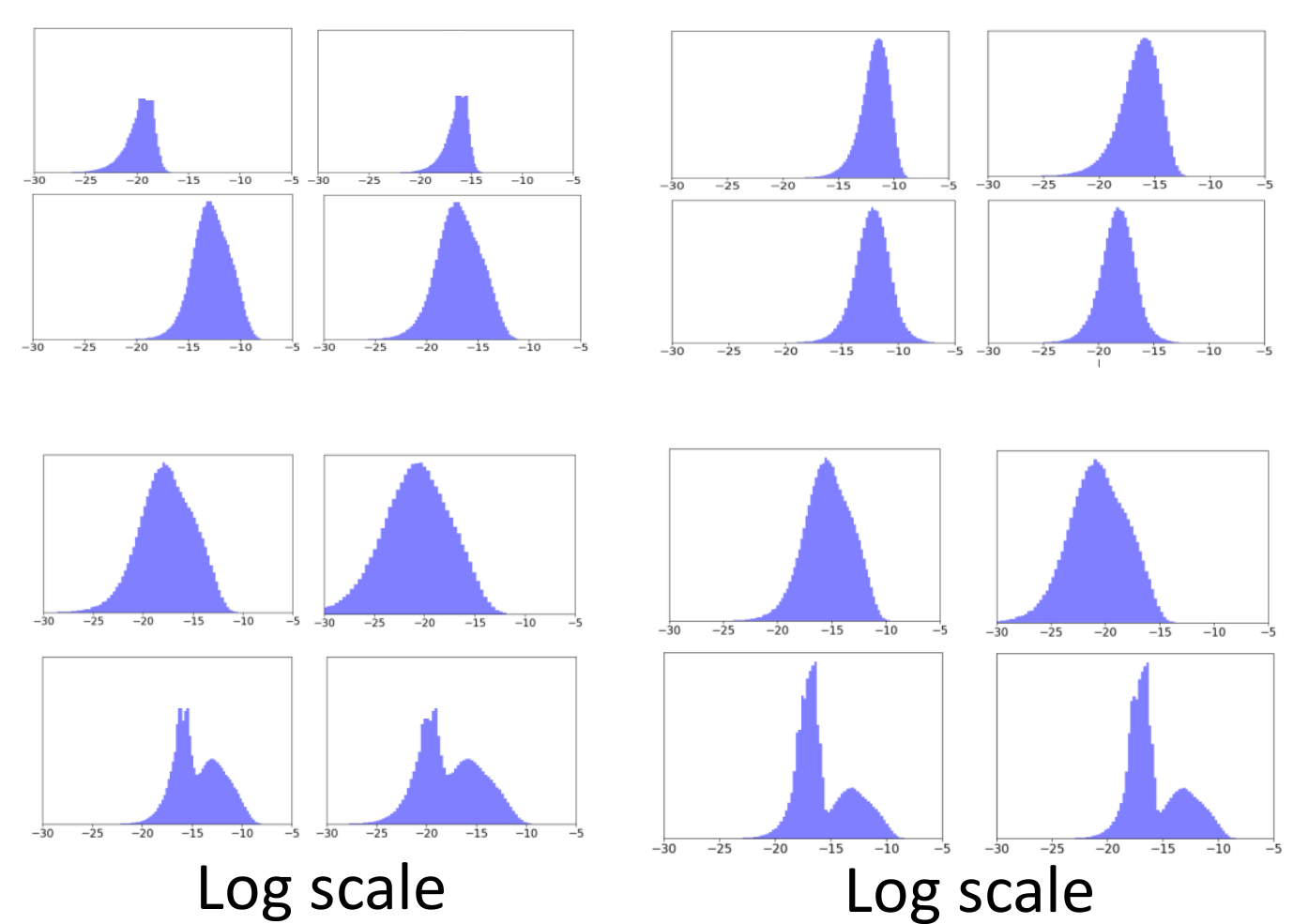}  
\caption{\textbf{(Top left)} $\nabla \mathcal{A}_{i;1}$ distribution divided to the two components of $\nabla \mathcal{A}_{i;2}$ in two different layers. The left mode (top) originates from the zero element of $\nabla \mathcal{A}_{i;2}$ and the right mode (bottom) from the non zero elements, ResNet18 on ImageNet. \textbf{(Top right)} Histograms of neural gradients in different layers of Transformer \citep{Vaswani2017AttentionIA} for WMT'16 En-De dataset in the decoder attention layer (top) and encoder embedding (bottom).\textbf{(Bottom left)} Histograms of neural gradients in different layers of DensNet121, ImageNet dataset. \textbf{(Bottom right)} Histograms of neural gradients in different layers of Vgg16, ImageNet dataset. } 
\label{fig:logNormalLeftRight}
\end{figure}

The two modes can be separated and analyzed by splitting the entries of $\nabla \mathcal{A}_{i;1}$ according to the value of the same entries in $\nabla \mathcal{A}_{i;2}$ (zero or non-zero); see the results of the Kolmogorov-Smirnov test to these distributions in \cref{tab:KS-testBiModal} - notice each of them can best be described as a lognormal distribution. Examples can be seen in \cref{fig:logNormalLeftRight}.

To estimate $\nabla \mathcal{A}_{i;1}$, we divide the data of $\nabla \mathcal{A}_{i;2}$ to zero and non-zero elements. Then, we apply Kolmogorov-Smirnov test \citep{KS_Test} to each of them, and show in \cref{tab:KS-testBiModal}  they fit to lognormal distribution  - obtaining a total of bi modal lognormal distribution in  $\nabla \mathcal{A}_{i;1}$. 

\begin{table}[h]
\centering
\caption{Mean ($\pm$ std) over all layers of Kolmogorov-Smirnov test on $\nabla \mathcal{A}_{i;1}$ over the zero and non zero elements of $\nabla \mathcal{A}_{i;2}$ ResNet18 on ImageNet}
\resizebox{\columnwidth}{!}{%
\begin{tabular}{c||ccccccc}
\toprule
Distribution & Laplace & Normal & Uniform & Cauchy & Logistic & Loglaplace & Lognormal\\ 
\hline
 zeros  & 0.43$\pm$0.13 & 0.41$\pm$0.09 & 0.93$\pm$0.19 & 0.44$\pm$0.12 & 0.53$\pm$0.13 & 0.1$\pm$0.03 & \textbf{0.08$\pm$0.03} \\
non-zeros & 0.48$\pm$0.11 & 0.49$\pm$0.13 & 0.86$\pm$0.09 & 0.43$\pm$0.11 & 0.51$\pm$0.14 & 0.04$\pm$0.012 & \textbf{0.02$\pm$0.01} \\
\bottomrule
\end{tabular}
}
\label{tab:KS-testBiModal}
\end{table}

\paragraph{Histograms.}
\label{sec:appendixHist}


In \cref{fig:logNormalLeftRight} we show  histograms of $\nabla \mathcal{A}_{i;1}$ divided to the zero and non-zero elements from $\nabla \mathcal{A}_{i;2}$ in ResNet18, ImageNet which as shown in \cref{sec:mixtureAppendix} each of them fit to lognormal distribution. Additionally we show histograms of different layers of Transformer \citep{Vaswani2017AttentionIA}, DenseNet121 and Vgg16 which at log-scale following to normal distribution.

\subsection{From sparsity to the threshold --- the full solution }
\label{sec:spToThFull}
\begin{align}
\begin{split}
S &= \mathbb{E}_\varepsilon \left[ \frac{1}{2} + \frac{1}{2}\mathrm{erf}\left(\frac{\ln{(\alpha\cdot\varepsilon) - \mu}}{\sqrt{2}\sigma}\right)\right] \underset{\varepsilon' = \frac{\varepsilon}{e^\mu}}{=} \int_{0}^{\frac{\alpha}{e^\mu}} \left[ \frac{1}{2} + \frac{1}{2}\mathrm{erf}\left(\frac{\ln{(\tau)}}{\sqrt{2}\sigma}\right)\right] \frac{e^\mu}{\alpha}  \,d\tau \\
&= \frac{1}{2} + \frac{e^\mu}{2\alpha}\left[e^{\frac{\sigma^2}{2}}\mathrm{erf}\left(\frac{\sigma^2 - \ln(\tau)}{\sqrt{2}\sigma}\right) + \tau \mathrm{erf}\left(\frac{\ln(\tau)}{\sqrt{2}\sigma}\right) \right] \Big|_{0}^{\frac{\alpha}{e^\mu}} \\
&= \frac{1}{2} + \frac{e^\mu}{2\alpha}\left[ e^{\frac{\sigma^2}{2}}\mathrm{erf}\left(\frac{\sigma}{\sqrt{2}} - \frac{\ln{(\frac{\alpha}{e^\mu})}}{\sqrt{2}\sigma}\right) + \frac{\alpha}{e^\mu} \cdot \mathrm{erf}\left(\frac{\ln{(\frac{\alpha}{e^\mu})}}{\sqrt{2}\sigma}\right) - e^\frac{\sigma^2}{2} \right]
\end{split}
\end{align}

\subsection{Floating point - relative error (Lognormal distribution)}
\label{secApp:fp8RelativeFull}

We assume that $x \sim \mathrm{Lognormal}(\mu,\sigma^2)$. Note that $E=\lfloor\ln x\rfloor\approx \ln x \sim \mathcal{N}(\mu,\sigma^2)$. We split the range into three parts according to $E$: (i) $-E_{\max}\leq E\leq E_{\max}$; (ii) $E\geq E_{\max}$; (iii) $E\leq -E_{\max}$, and calculate the expected contribution for each term of the relative error in \cref{eq:relativeError}:

 \subsubsection{The case of $-E_{\max}\leq E\leq E_{\max}$}
 \label{sec:lognormalMedCase}
In this case $E$ can be expressed and no distortion appears due to clipping i.e., $E=E_q$. Therefore, the only contribution related to $\eta(n_1,n_2)$ is due to the distortion in the mantissa. This distortion is approximated through an additive noise $m_q=m+\varepsilon$, where the noise has a uniform distribution   $\varepsilon \sim \mathbb{U}[-\Delta/2,\Delta/2]$. We can then calculate the expected relative error as follows
\begin{equation*}
\begin{split}
    E\left[\left| \frac{x_q-x}{x}\right|\Bigg | -E_{\max}\leq E\leq E_{\max} \right] = &  E\left[\left|\frac{m\cdot 2^{E}- m_q\cdot 2^{E_q}}{m\cdot 2^E}\right| \Bigg | -E_{\max}\leq E\leq E_{\max}\right]
    \\= & E\left[\left|\frac{m\cdot 2^{E}- (m+\varepsilon)\cdot 2^{E}}{m\cdot 2^E}\right| \Bigg | -E_{\max}\leq E\leq E_{\max}\right]
    \\= & E\left|\frac{\varepsilon}{m}\right|=  E \left( \frac{|\varepsilon|}{|m|}\right)=  E |\varepsilon| \cdot E\left| \frac{1}{ m}\right|
\end{split}
\end{equation*}

The last factorization is permissible since $\varepsilon$ is independent of $m$.  The expectation of $|\varepsilon|$ in the range $[-\Delta/2,\Delta/2]$ is calculated as follows: 
\begin{equation}
    E|\varepsilon| = \int _{-\frac{\Delta}{2}}^{\frac{\Delta}{2}} \dfrac{\left|\varepsilon\right|}{\Delta}\cdot d\varepsilon= \dfrac{\varepsilon\cdot\left|\varepsilon\right|}{2\Delta}\Bigg|_{-\frac{\Delta}{2}}^{\frac{\Delta}{2}}= \dfrac{\left|\Delta\right|}{4} = \frac{1}{4\cdot (2^{n_1}-1)}
\end{equation}

We turn to establish $ E\left| \frac{1}{ m}\right|$. Given a real value $x\in \mathbb{R^{+}}$, by definition $m$ is positive and thus can be expressed as follows:
\begin{equation}
    \left| \frac{1}{ m}\right| = \frac{1}{m} = \frac{1}{2^{\ln x-\lfloor\ln x\rfloor }} = \frac{1}{2^{\varepsilon_m}}
\end{equation}

where $\varepsilon_m$ can be treated as a uniform variable in the range [0,1]. Therefore, we have that
\begin{equation}
     E\left| \frac{1}{ m}\right| = \int_0^1 \frac{1}{2^{\varepsilon_m}}\cdot d \varepsilon = -\dfrac{1}{\ln\left(2\right){\cdot}2^x}\bigg|_0^1 = -\dfrac{1}{\ln\left(2\right){\cdot}2} \approx 0.721
\end{equation}

We can finally state the expected relative error for the case where $x$ is supported by the dynamic range of $x_q$ 

\begin{equation}
    E\left[\left| \frac{x_q-x}{x}\right|\Bigg | -E_{\max}\leq E\leq E_{\max} \right] = \frac{1}{8\cdot \ln\left(2\right)\cdot (2^{n_1}-1)}
\end{equation}

We turn to find the probability  $ P\left(-E_{\max}\leq E\leq E_{\max} \right)$. Assuming the lognormal distribution of x we get:
 \begin{equation}
 \begin{split}
   P\left(-E_{\max}\leq E\leq E_{\max} \right) & 
    =\Phi\left(\frac{E_{\max}}{\sigma}\right) -  \Phi\left(\frac{-E_{\max}}{\sigma}\right) 
   \\ &= 2\Phi\left(\frac{E_{\max}}{\sigma}\right)  -1
   \end{split}
 \end{equation}

 where $\Phi(x)$ is the cumulative distribution function of the standard normal distribution.

\subsubsection{The case of  $E\geq E_{\max}$}
 \label{sec:lognormalbigCase}

 \begin{equation}
 \begin{split}
    E\left[\left| \frac{x_q-x}{x}\right|\Bigg | E\geq E_{\max} \right]\cdot & P\left( E\geq E_{\max} \right) =
    \\& = \int_{E_{\max}}^{\infty} \frac{2^{E}-2^{E_{\max}}}{2^{E}} \cdot \frac{1}{\sqrt{2 \pi} \sigma} \mathrm{e}^{-\frac{E^{2}}{2 \sigma^{2}}} \mathrm{d}E
   \\ & = \frac{1}{2}\operatorname{\mathrm{erf}}\left(\frac{E}{\sqrt{2}\sigma}\right) - 2^{E_{\max}-1}\mathrm{e}^\frac{\sigma^2\ln^2\left(4\right)}{8}\operatorname{\mathrm{erf}}\left(\frac{E}{\sqrt 2\sigma}+\frac{\sigma\ln\left(4\right)}{\sqrt 8}\right) \Big |_{E_{\max}}^\infty 
   \\& =  2^{E_{\max}-1}\mathrm{e}^\frac{\sigma^2\ln^2\left(2\right)}{2}\left(\operatorname{\mathrm{erf}}\left(\frac{\sigma \ln 2 }{\sqrt{2}}+\frac{E_{\max}}{\sqrt{ 2}\sigma }\right)-1 \right) +
   \\& -\frac{1}{2}\operatorname{\mathrm{erf}}\left(\dfrac{E_{\max}}{\sqrt 2\sigma}\right)+\frac{1}{2}
   \end{split}
 \end{equation}

\subsubsection{The case of  $E\leq -E_{\max}$}
This is the underflow case. Here we have by definition that $x_q=0$ and therefore we get  
\begin{equation}
 E\left[\left| \frac{x_q-x}{x}\right|\Bigg | E\leq -E_{\max} \right] = E\left[\left| \frac{0-x}{x}\right|\Bigg | E\leq -E_{\max} \right] = 1 
\end{equation}

This case has a probability of $ P\left( E\leq -E_{\max} \right)=\Phi\left(-\frac{E_{\max}}{\sigma}\right) = 1- \Phi\left(\frac{E_{\max}}{\sigma}\right)$

\subsubsection{Final Expected Relative Error  }
\label{SecAppen:finalRelated}
Combining the three terms we have that 

\begin{equation}
\begin{split}
 E\left[\left| \frac{x_q-x}{x}\right|\right] =& \frac{2\Phi\left(\frac{E_{\max}}{\sigma}\right)  -1}{8\cdot \ln\left(2\right)\cdot (2^{n_1}-1)} + 
 2^{E_{\max}-1}\mathrm{e}^\frac{\sigma^2\ln^2\left(2\right)}{2}\left(\operatorname{\mathrm{erf}}\left(\frac{\sigma \ln 2 }{\sqrt 2}+\frac{E_{\max}}{\sqrt 2\sigma }\right)-1\right)+
   \\ &-\frac{1}{2}\operatorname{\mathrm{erf}}\left(\dfrac{E_{\max}}{\sqrt 2\sigma}\right)+\frac{3}{2}- \Phi\left(\frac{E_{\max}}{\sigma}\right)
 \end{split}
 \label{final}
\end{equation}

Given any $N$-bit FP format, seek a mantissa-exponent partition that minimizes the expected relative error such that $n_1+n_2=N-1$. Minimizing \cref{final} yields this optimal partition. To do so we set $n_1= N-n_2-1$, equate the derivative to zero and solve. Empirically we found the computational cost of such solution is negligible and can be done online without affecting the NN running time.

Moreover, we can notice that for small values of $\sigma$ a lognormal distributed tensor is similar to a normal distributed tensor. In such cases, the minimization of the relative error in \cref{final} induced to allocate 0 bits to the exponent, i.e fixed point quantization. This fit to previous results \citep{banner2018post,baskin2018nice} which shows impressive results by quantization the fwd pass with fixed point quantization.

We can notice that although we use a different analytical measurement for stochastic pruning (cosine similarity) and for quantization (relative error), both of them are "analytical related" and define normalize values for the distance between original and noised tensor. The use of relative error in floating point format was already used in previous works\footnote{Accuracy and Stability of Numerical Algorithms: Second Edition; Nicholas J. Higham}.

\begin{figure}[h]
 \centering
 \includegraphics[width=0.7\linewidth]{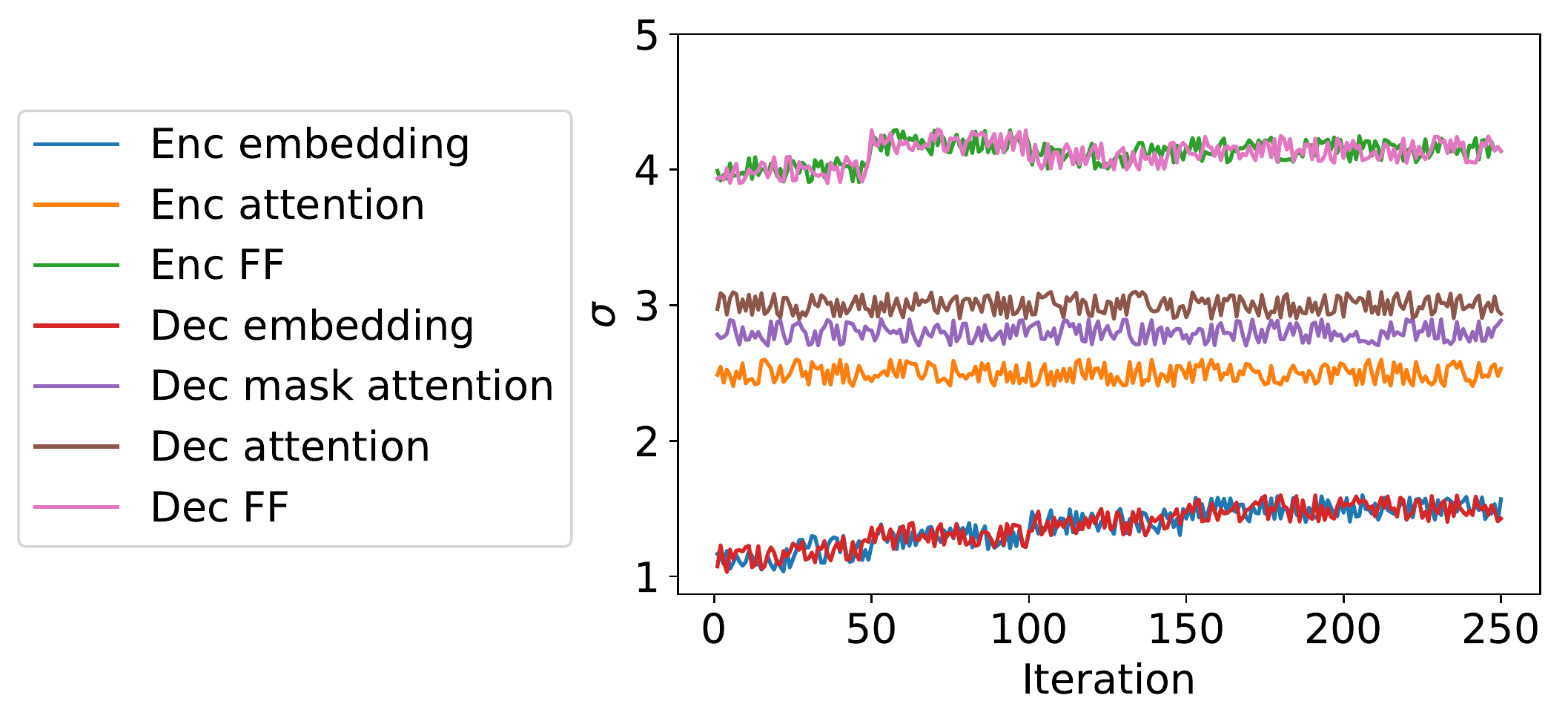}  
\caption{(a) Std of layers gradient in different layers in the transformer \citep{Vaswani2017AttentionIA} in WMT'16 En-De dataset assuming lognormal distribution. Notice the variability of std, which make the fp quantization challenging \citep{Micikevicius2017MixedPT}.} 
\label{fig:transformerStd}
\end{figure}

\begin{figure}[h]
 \centering
 \includegraphics[width=0.9\linewidth]{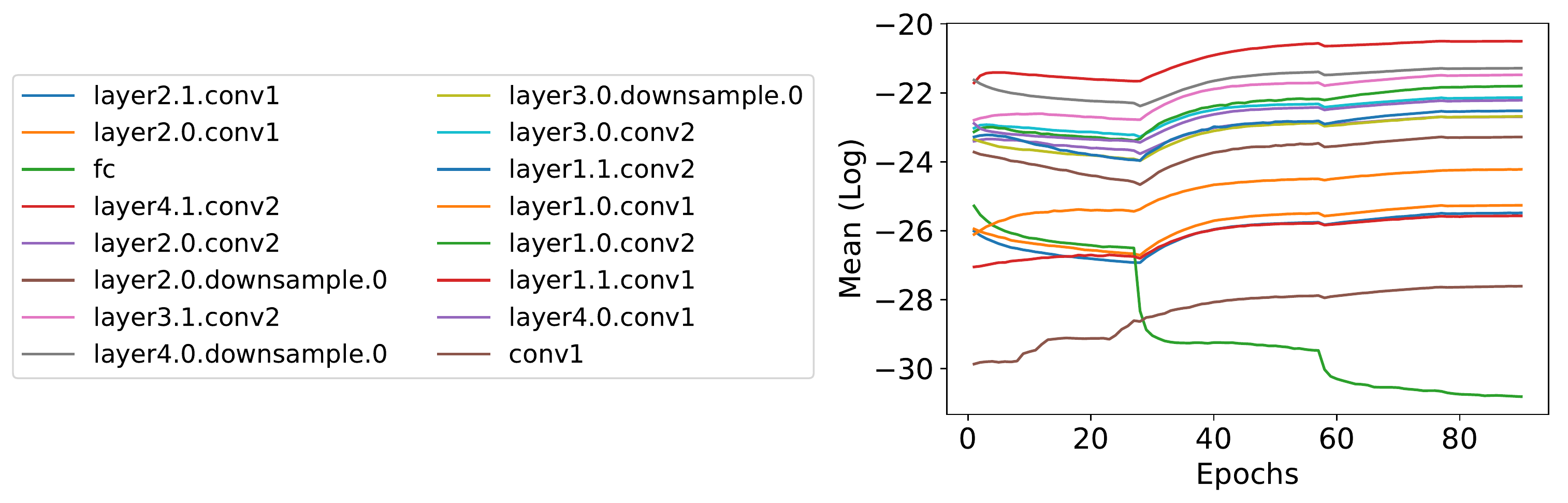}  
\caption{Layer gradients means (log scale) in all layers of ResNet18, ImageNet dataset. Notice that the first ("Conv1") and last ("fc") suffer from a different statistic which requires a layerwise scaling factor.} 
\label{fig:resnet_first_last}
\end{figure}

\subsection{Floating point - relative error (Normal distribution)}
\label{secApp:fp8RelativeFull-Normal}

We assume that $x \sim N(\mu,\sigma)$, and similar to \cref{secApp:fp8RelativeFull} we split the range into three parts according to $E$: (i) $-E_{\max}\leq E\leq E_{\max}$; (ii) $E\geq E_{\max}$; (iii) $E\leq -E_{\max}$. It is similar to the lognormal case, with maximum exponent of $2^{E_{max}}$.

 \subsubsection{The case of $-E_{\max}\leq E\leq E_{\max}$}

\begin{equation}
    \frac{1}{8\cdot \ln\left(2\right)\cdot (2^{n_1}-1)} 2\Phi\left(\frac{2^{E_{\max}}}{\sigma}\right)  -1
\end{equation}

\subsubsection{The case of  $E\geq E_{\max}$}

 \begin{equation}
 \begin{split}
    2^{E_{\max}-1}\mathrm{e}^\frac{\sigma^2\ln^2\left(2\right)}{2}\left(\operatorname{\mathrm{erf}}\left(\frac{\sigma \ln 2 }{\sqrt{2}}+\frac{2^{E_{\max}}}{\sqrt{ 2}\sigma }\right)-1 \right) -\frac{1}{2}\operatorname{\mathrm{erf}}\left(\dfrac{2^{E_{\max}}}{\sqrt 2\sigma}\right)+\frac{1}{2}
   \end{split}
 \end{equation}

\subsubsection{The case of  $E\leq -E_{\max}$}

\begin{equation}
1- \Phi\left(\frac{2^{E_{\max}}}{\sigma}\right)
\end{equation}

\begin{figure}[h]
 \centering
 \includegraphics[width=0.5\linewidth]{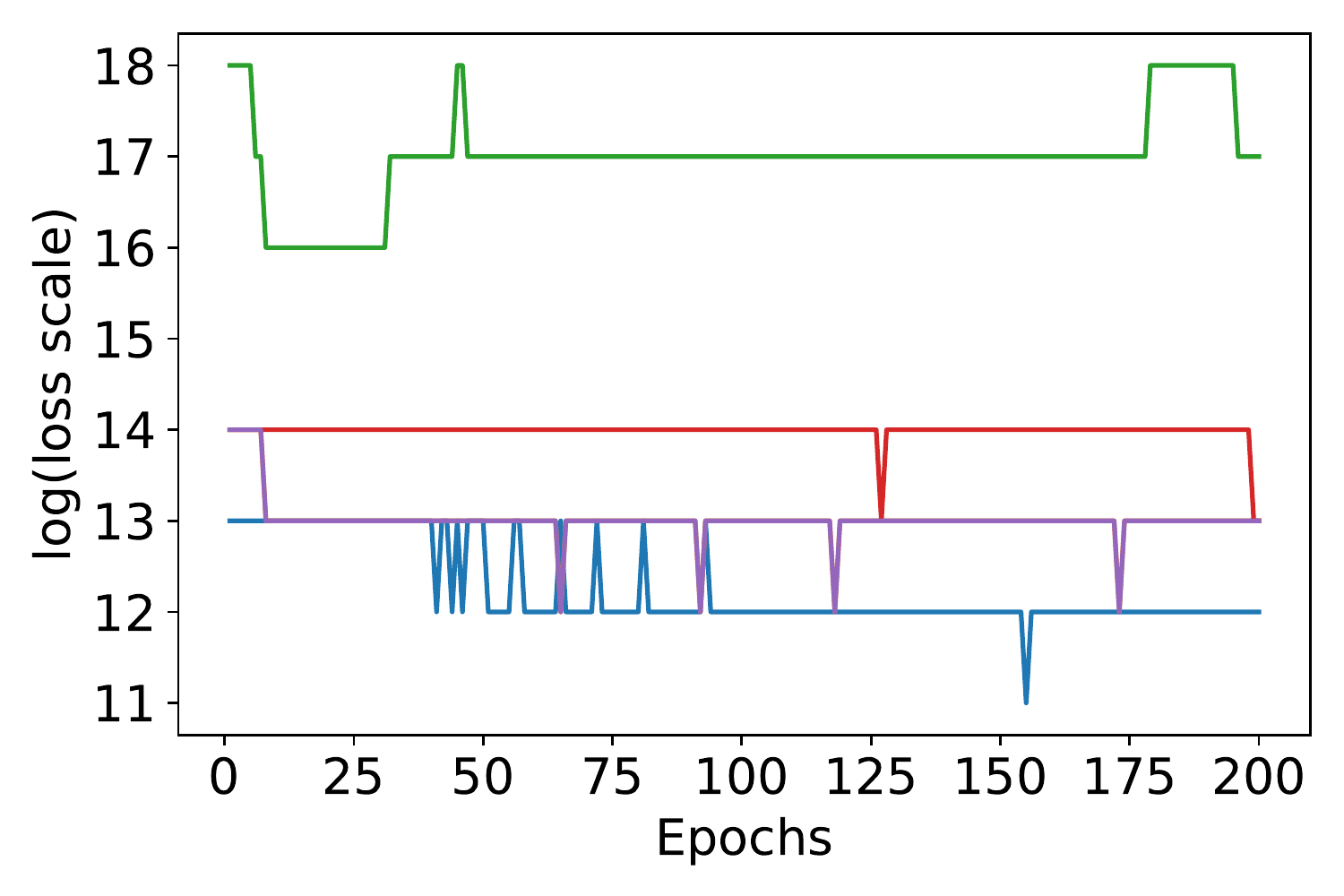}  
\caption{The proposed gradient scale factor in different layer of ResNet18 in cifar10 dataset for FP5 format (1-4-0). Notice the variability across the layers which emphasize the importance of a different scale factor per layer.} 
\label{fig:lossScaleAp}
\end{figure}

\subsection{Gradient scaling}
\label{sec:apGradientScaling}

The suggested neural gradient scale algorithm presented in \cref{alg:scale}  sacrifices the smallest elements to keep the largest gradients magnitude representable. 

As shown in \cref{fig:imagenetMax} the maximum of the gradients’ distribution at each layer is quite stable throughout the training. Therefore, we can infrequently sample the values of these maxima. This, in combination with the scaling in log scale, allows us to create an efficient hardware implementation of the suggested gradient scaling.

\begin{algorithm}[h!]
	\caption{Gradient scaling}
	\label{alg:scale}
	\begin{algorithmic}[1]
		\State\underline{\textbf{Input: }} Neural gradient X, number of exponent bits $n_2$
		\State\underline{\textbf{Output: }} Quantized gradient $X_Q$ using FP quantization $ Q_{FP}$
		\State $E_{\max} = 2^{n_2-1}$ 
		\State $\mu_l = \floor{\log_2{{\max (|X|)}}} / \log_2{(E_{\max})} $
		\State $X_Q = 2^{\mu_l} \cdot Q_{FP}(X / 2^{\mu_l})$ 
	
	\end{algorithmic}
\end{algorithm}



\subsection{Stochastic pruning complexity}
\label{sec:complexity}
The suggested stochastic pruning method solution requires the mean and standard deviation of the input --- each operation has a complexity of $\mathcal{O}(n)$ for an input size $n$. In practice, we perform this procedure of measuring the mean and standard deviation once every epoch. Additionally, it requires the solution of \cref{spToTh} which is independent in the input size. We empirically found it converges in a few iterations. In total, we found that the process of obtaining a threshold ,$\alpha$, for a required sparsity adds less than 5\% to the iteration time (for the single iteration, once in an epoch, when we calculate the threshold). For the rest of the iterations, there is no overhead caused by the stochastic pruning while achieving a significant improvement because of the induced sparsity.

We can compare our method with top-k, which requires every iteration $\mathcal{O}(n\log n)$ for a naive implementation (sorting) and can be reduced using quick-select \citep{quickselect} as was proposed by \citep{SWAT2020} and vary between $\mathcal{O}(n)$ and $\mathcal{O}(n^2)$.

\subsection{Algorithm for bi-Modal lognormal pruning}
\label{sec:biModalSparsAppendix}

Pruning of the bi-modal gradients ($\nabla \mathcal{A}_{i;1}$) is crucial for reducing the computational and bandwidth overhead. This is especially because it is involved in the calculation of the next layer gradient($\nabla \mathcal{A}_{i-1;3}$) and weights gradients of the convolution ($\nabla W_i$). Unlike  $\nabla \mathcal{A}_{i;2}$, which naturally has a high sparsity level induced by the ReLU, $\nabla \mathcal{A}_{i;1}$ is naturally not sparse at all.  In the following, we explain how to find the threshold $\alpha$ which induces the required sparsity ratio $S$ in the bi-modal lognormal gradient.

To achieve even higher sparsity levels we can apply stochastic pruning method to the mixture of lognormally distributed gradients, $\nabla \mathcal{A}_{i;1}$. These gradients are comprised of two lognormal distributions with different means, variances, and the number of elements. Dealing with this distribution in the same theoretical framework as was done for the lognormal distribution is challenging. However, a simple solution can allow us to apply the same results to this distribution. We will rely on the fact that most of the values in the left mode are several orders of magnitude smaller than the ones in the right mode. Hence for a large enough sparsity value, we can say the following - if we will ignore the left mode and calculate the threshold assuming only the right mode exists then for almost all of the values in the left mode $|x| < \alpha\cdot\varepsilon$ and thus will be pruned. Therefore, to achieve an overall sparsity level of $S$ we can find the threshold using the parameters of the right mode, adjusting the desired sparsity level down to account for the elements of the left mode that would be pruned as well.

\begin{figure}[h]
\centering
\begin{subfigure}{.48\textwidth}
  \centering
  \includegraphics[width=\linewidth]{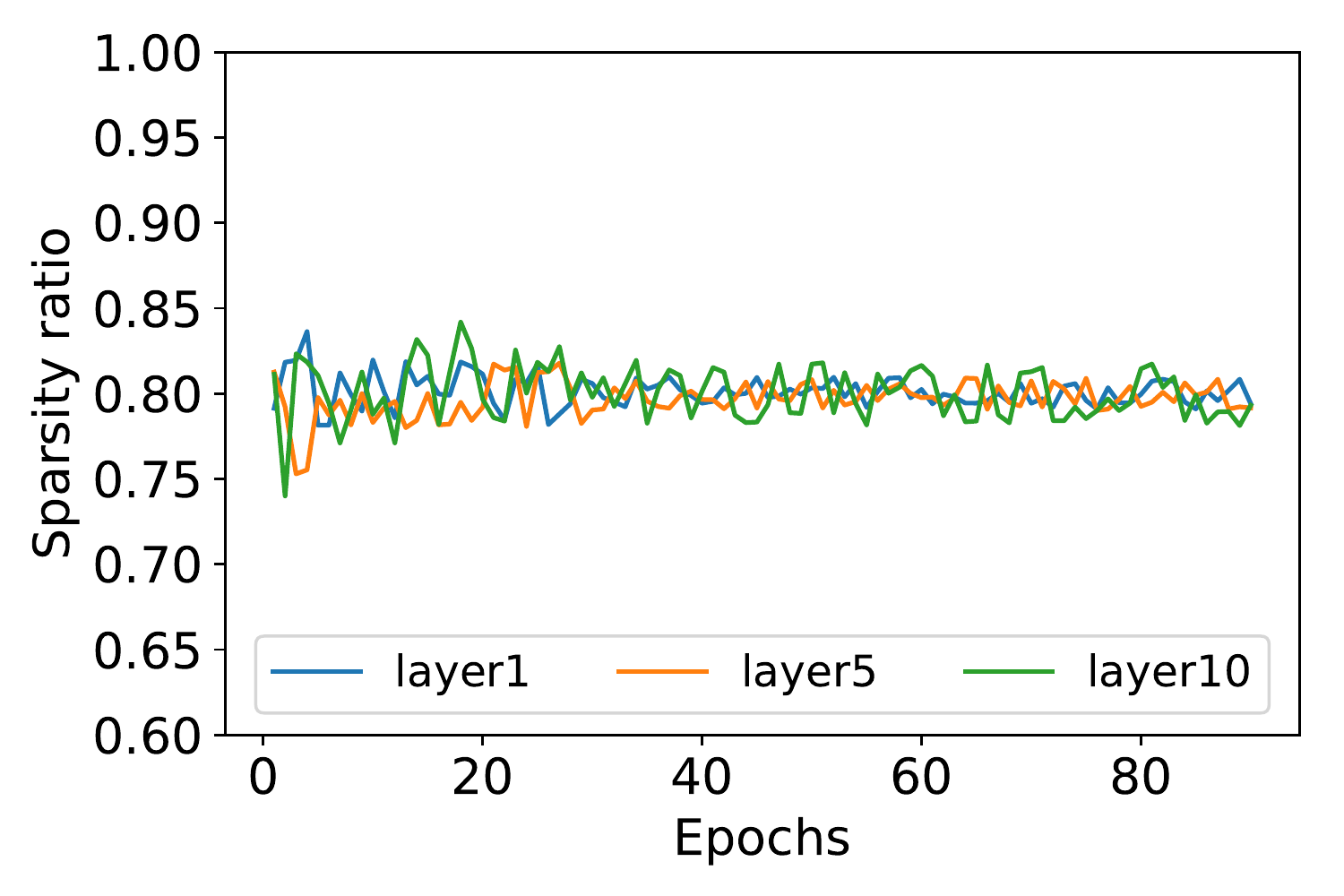}  
  \caption{}
  \label{}
\end{subfigure}
\begin{subfigure}{.48\textwidth}
  \centering
  \includegraphics[width=\linewidth]{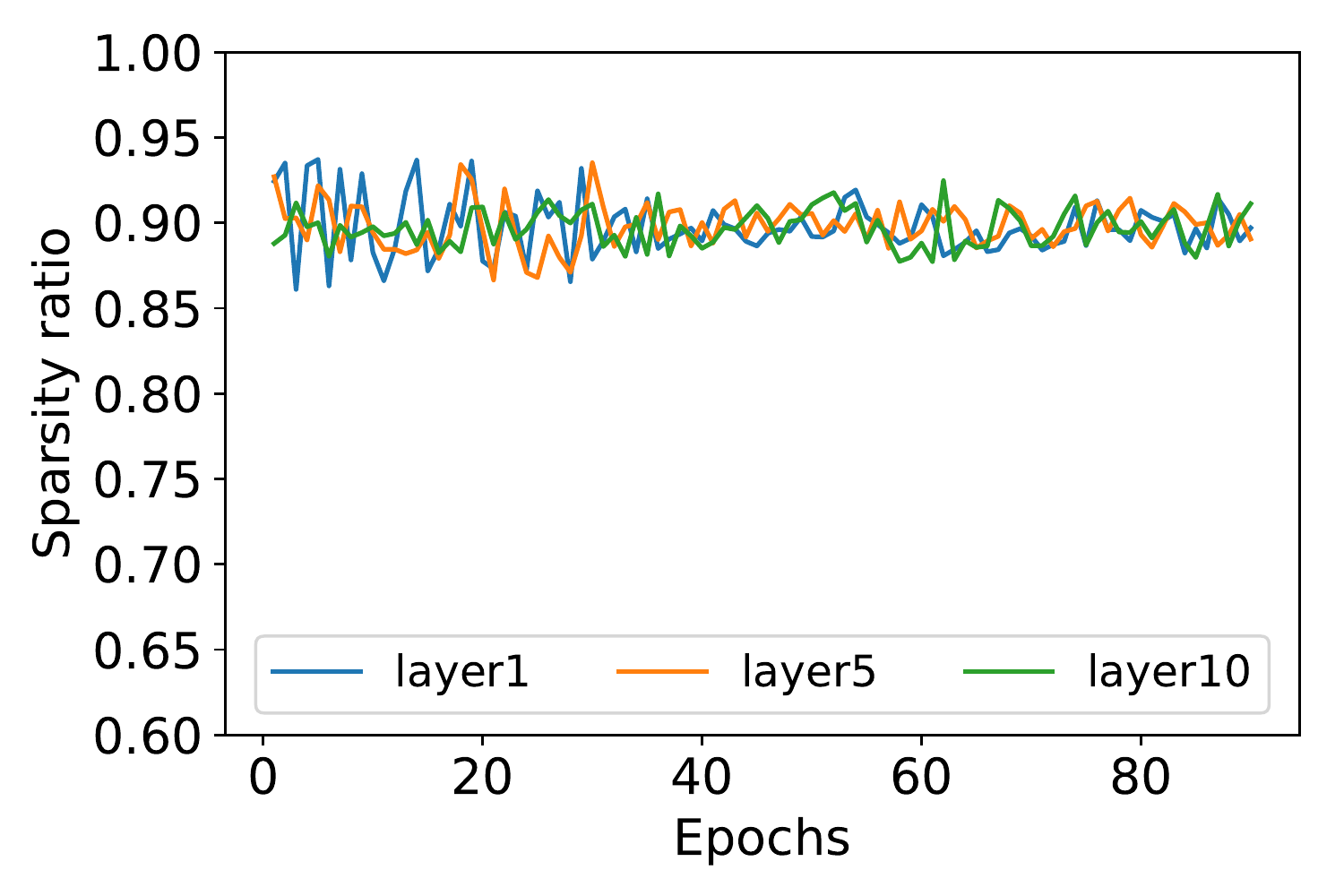}  
  \caption{}
  \label{}
\end{subfigure}
\caption{The obtained sparsity in different layers of ResNet-18 with ImageNet dataset for required sparsity of 80\% (a) and 90\% (b) }
\label{fig:actualRatio}
\end{figure}

In order to achieve a sparsity level $S$ we use the following procedure --- (1) calculate the ratio of the gradients that are in the left mode, denoted $l$, this is done by dividing the amount of values that were zero in the gradient before ($\nabla \mathcal{A}_{i;2}$) by the total number of elements in the gradient. (2) Here we assume that $S$ is large enough, i.e. $l<S$. thus in order to achieve a total sparsity of $S$ out of the entire tensor we need to sparsity $S' = \frac{S-l}{1-l}$ of the right mode. (3) Calculate the mean $\mu$ and variance $\sigma$ of the right mode, by taking only the values in the tensor that their corresponding values in the pre-BN gradients are non-zero. (4) Solve the equation to find the threshold using $\sigma$ and $S'$  and apply stochastic pruning to the entire tensor using it.\\
We found that for high values of sparsity, above 0.7, the method is very accurate and achieves the desired sparsity thorough-out the training. For lower sparsity levels the actual sparsity induced tends to be a bit lower because the threshold is lower and the assumption that all of the values in the left mode are zeroed deteriorates progressively as the sparsity level decreases.  In \cref{fig:actualRatio} we show that the proposed method indeed achieves the desired sparsity.

\subsection{Cosine similarity and heterogeneous stochastic pruning} 
\label{sec:allocationAppendix}

\subsubsection{Cosine similarity of stochastic pruning - general}
Given a vector of gradients 
$X$, we would like to measure the cosine similarity (cosine of the angle) between the vector before and after stochastic pruning, i.e. $X$ and $T_{\alpha,\varepsilon}\left (x\right)$: 
\begin{equation}\label{eq:1}
\cos(\theta)=\dfrac{X\cdot T_{\alpha,\varepsilon}\left (x\right)}{||X||_2\cdot ||T_{\alpha,\varepsilon}\left (x\right)||_2}
\end{equation}

At high dimensions, the relative error made by substituting $||X||$ with $E||X||$ becomes asymptotically negligible \citep{biau2015high}. This distance concentration phenomenon simplifies Eq. \ref{eq:1}:
\begin{equation}\label{eq:2}
\cos(\theta)\approx\dfrac{\mathbb{E}[X\cdot T_{\alpha,\varepsilon}(X)]}{\mathbb{E}||X||_2\cdot \mathbb{E}||T_{\alpha,\varepsilon}\left (x\right)||_2}
\end{equation}

Solving the different components of eq. \ref{eq:2} for $X \sim \text{lognormal}(0,\sigma^2)$ with a maximum value of $k\sigma$ (regarding the maximum value see \cref{Gumbel}) with n elements, we get (see \cref{sec:appendStochasticPrunDetailed}):

\begin{align}
\begin{split}
    \mathbb{E}[X\cdot T_{\alpha,\varepsilon}\left (X\right)]
      &=\frac{1}{2}n\cdot e^{2\sigma^2} \left[ 1 + \mathrm{erf}\left(\frac{k\sigma-2\sigma^2}{\sigma\sqrt{2}}\right) \right]
\end{split}
\label{eq:X_TX-stochastic}
\end{align}

\begin{align}
\begin{split}
     \mathbb{E}||T_{\alpha,\varepsilon}\left (X\right)||_2^2 
    &=  \frac{1}{2}n\cdot \alpha \cdot e^{\frac{\sigma^2}{2}}\left[1 - \mathrm{erf}\left(\frac{\sigma^2 - \ln(\alpha)}{\sigma\sqrt{2}}\right) \right] \\ &+ \frac{1}{2}n\cdot e^{2\sigma^2} \left[\mathrm{erf}\left(\frac{k\sigma-2\sigma^2}{\sigma\sqrt{2}}\right) -\mathrm{erf}\left(\frac{\ln(\alpha)-2\sigma^2}{\sigma\sqrt{2}}\right)\right] 
\end{split}
\label{eq:TXstochastic}
\end{align}

\begin{align}
\begin{split}
     \mathbb{E}||X||_2^2 & =  \frac{1}{2}n\cdot e^{2\sigma^2} \left[ 1 + \mathrm{erf}\left(\frac{k\sigma-2\sigma^2}{\sigma\sqrt{2}}\right) \right]
\end{split}
\label{eq:X-stochastic}
\end{align}

\begin{figure}[h]
\begin{subfigure}{.5\textwidth}
  \centering
  \includegraphics[width=\linewidth]{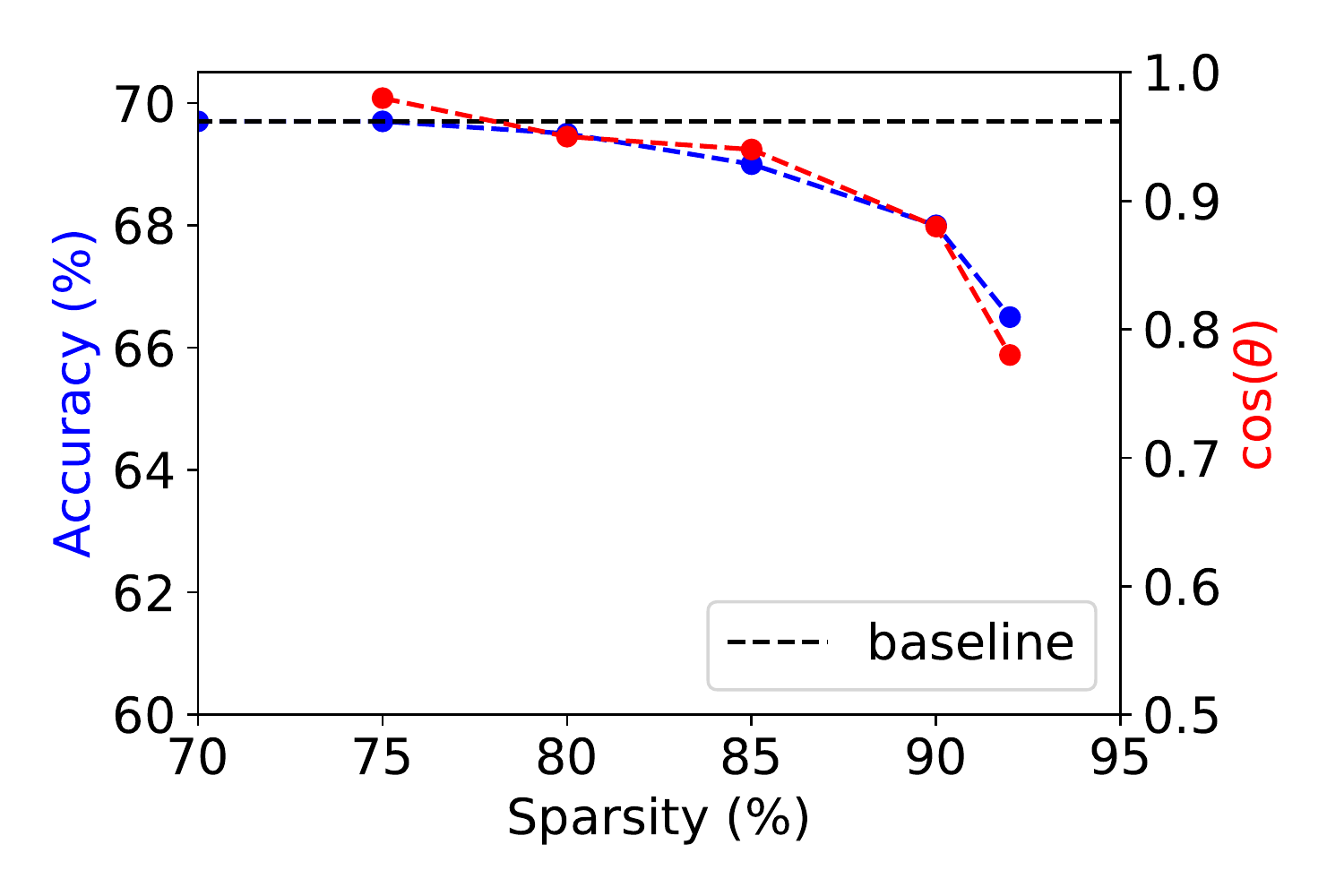}  
  \caption{}
  \label{fig:AccVsCosine}
\end{subfigure}
\begin{subfigure}{.5\textwidth}
  \centering
  \includegraphics[width=\linewidth]{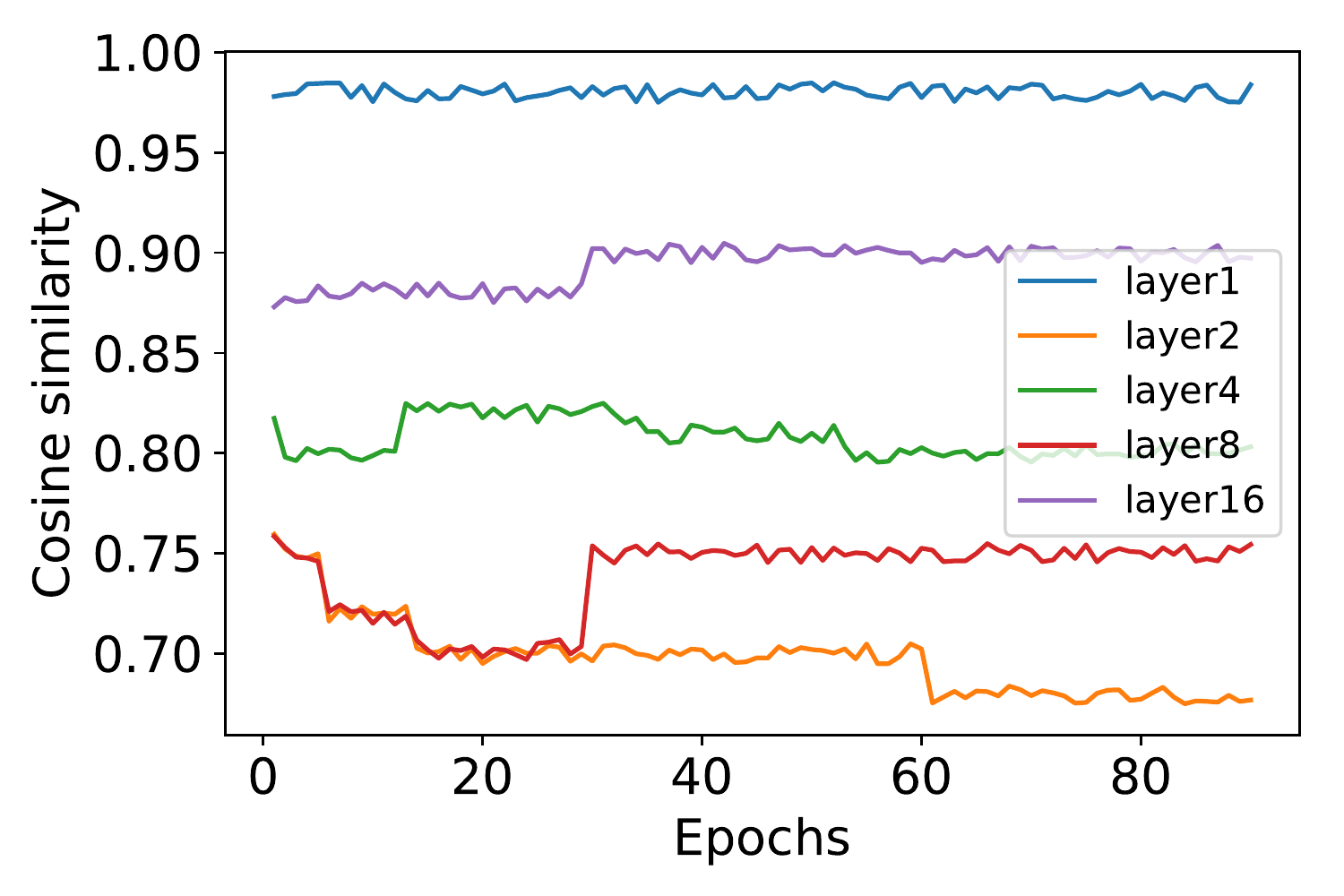}  
  \caption{}
  \label{fig:cosineLayers}
\end{subfigure}
\caption{(a) ResNet18 validation accuracy on the ImageNet data-set for different sparsity levels and the mean of the cosine similarity between the pruned and original gradient, notice the high correlation between the two. (b) The cosine similarity in different layers of ResNet18 on Imagenet dataset for 92 \% pruning. Notice the variability between the different layers.}
\label{fig:cosine}
\end{figure}

\begin{figure}[h]
\begin{subfigure}{.5\textwidth}
  \centering
  \includegraphics[width=\linewidth]{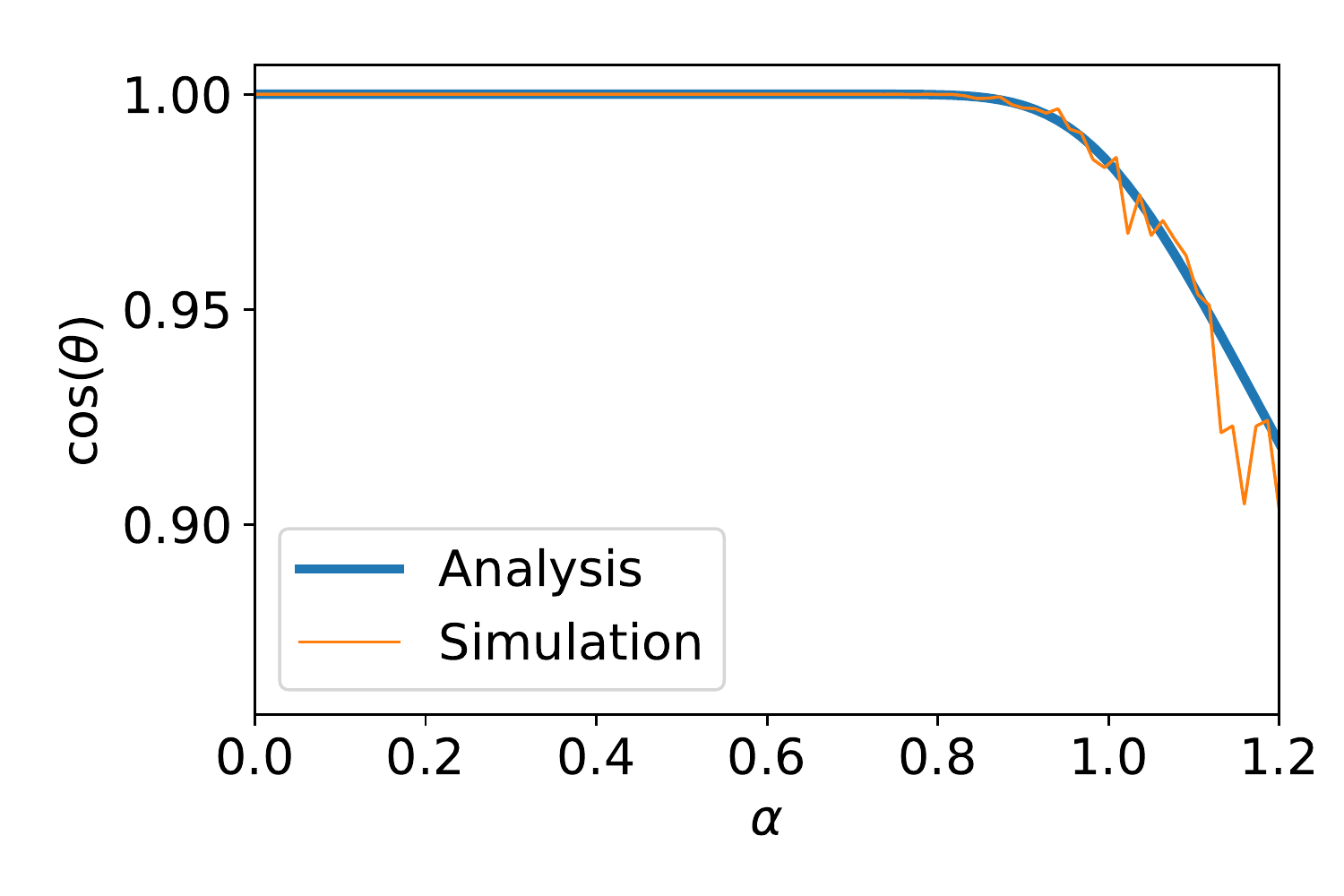}  
  \caption{}
  \label{}
\end{subfigure}
\begin{subfigure}{.5\textwidth}
  \centering
  \includegraphics[width=\linewidth]{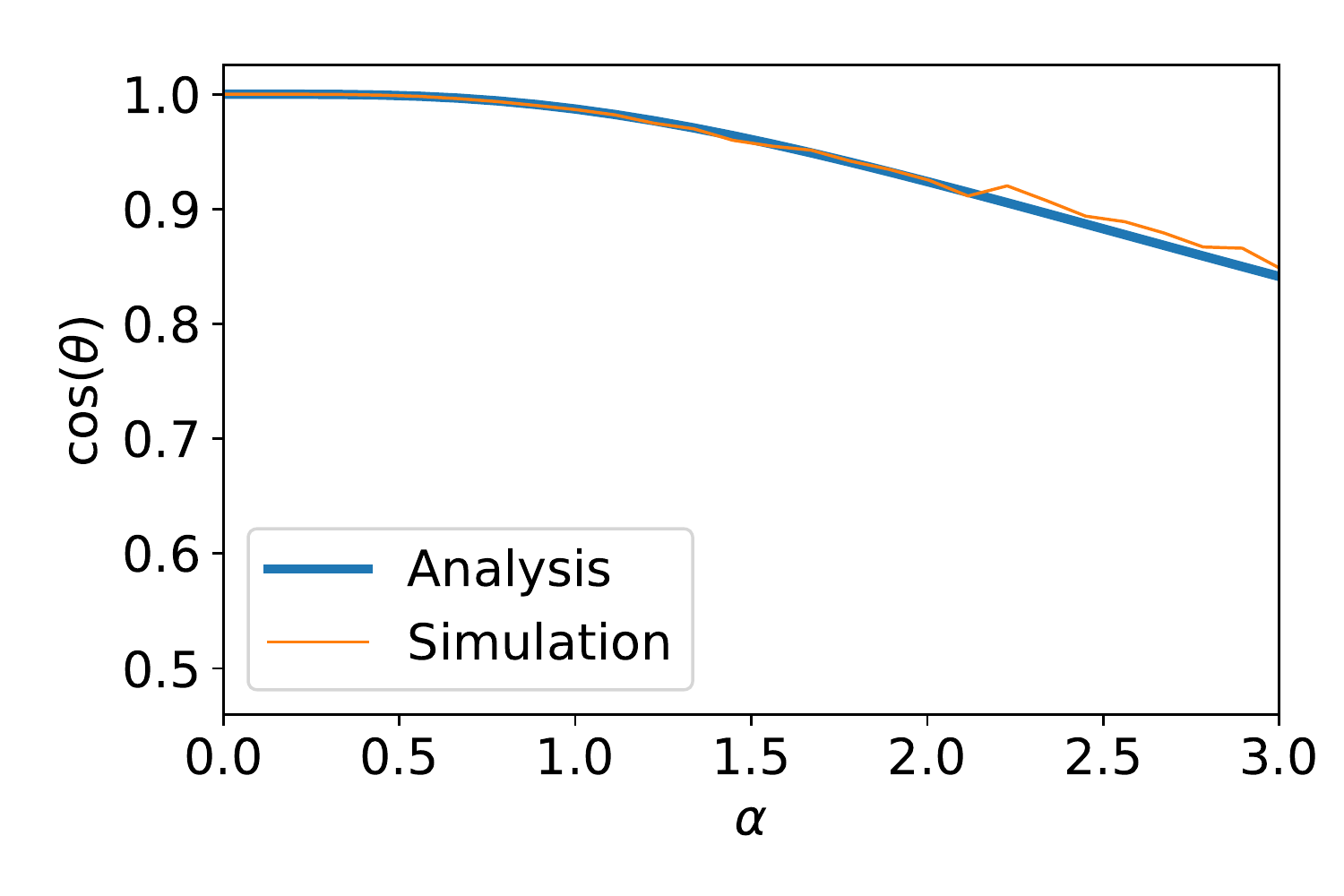}  
  \caption{}
  \label{fig:sub-second}
\end{subfigure}
\caption{Comparison between simulation and analysis ( \cref{eq:X_TX-stochastic}, \cref{eq:TXstochastic} and \cref{eq:X-stochastic} ) of the cosine similarity for $\sigma = 3$ (a) and $\sigma = 5$(b).}
\label{fig:simulationAnalysisCosine}
\end{figure}

\cref{fig:simulationAnalysisCosine} shows that these analytical results 
 are in agreement with our simulations. \\
We notice in \cref{fig:AccVsCosine} that the cosine similarity between original and pruned neural gradients can be used as a proxy for validation accuracy degradation under stochastic pruning.
Interestingly, in \cref{fig:cosineLayers} we show that using the cosine similarity, we observed that stochastic pruning takes a different toll from the different layers i.e. pruning all layers to the same sparsity level damages some of them more than others.
We propose an algorithm that using the analytical measured cosine similarity (\cref{eq:X_TX-stochastic},\cref{eq:TXstochastic},\cref{eq:X-stochastic}) preserves the cosine similarity of some of the layers, by decreasing their sparsity level, while increasing the sparsity level of other layers — maintaining the overall sparsity budget (mean sparsity of all the layers). The motivation for that and the algorithm are described in \cref{Heterogeneous_general} and \cref{Heterogeneous_algorithm}.

\subsubsection{Heterogeneous stochastic pruning}
\label{Heterogeneous_general}
Using stochastic pruning, we noticed that different layers, with different distributional parameters, seem to be more sensitive to high levels of sparsity. This sensitivity can affect the overall accuracy achieved by the network. We notice two important parameters that contribute to the sensitivity of the layer: (1) its depth in the network, and (2) the error it suffers at a given sparsity level. The depth of the layer in the network is important because if a layer is deeper (closer to the output) then corrupting its gradients will cause the gradients reaching lower levels (closer to the input) to be corrupted as well. So, corruption of the gradients of the first layer, for example, affects only the first layer itself, while corrupting the gradients of the second layer affect both its gradients and the first layer's gradients, etc. \\
To measure the distortion of the tensor due to the pruning, we used the cosine similarity (see definition in \cref{eq:1}) between the tensor before and after the stochastic pruning process. This measure reflects the preservation of the angle of the high-dimensional gradients tensor, under the pruning process. We noticed that in our homogeneous pruning scheme through all layers had the same desired sparsity level and a very similar achieved sparsity, the cosine similarity varied significantly between the layers. This is seen clearly in \cref{fig:cosineLayers} where each layer has a fairly stable cosine similarity that might be very different than the other layers, this is similar to the stability of each layer's distributional parameters that also vary between layers. In fact, as observed in \cref{fig:cos_simVsSparsityTheoretical}, the variance in cosine similarity between the layers can be explained exactly by the variance in the distributional parameters, for example the std $\sigma$.\\

\subsubsection{Heterogeneous algorithm}
\label{Heterogeneous_algorithm}
While the effect of each layer's pruning on the overall outcome (accuracy) stems from its depth in the network and its distortion (measured through cosine similarity) its contribution to the overall sparsity is controlled by the number of its elements and its sparsity level. Since we aim to maximize the overall sparsity while preserving the baseline accuracy, we propose using different sparsity levels for the different layers. This would allow us to preserve the cosine similarity of the deeper layers, that affect the gradients of more layers while raising the sparsity of lower levels to maintain the overall sparsity. This can be done effectively because for most architectures the number of elements in the neural gradients of deeper layers is lower than these of shallow layers (usually because of max-pooling). This allows us to lower the sparsity of deeper layers significantly while raising the sparsity of shallower ones only slightly and still preserve the overall sparsity.\\
We propose an algorithm in which the user determines the overall desired sparsity and the minimum cosine similarities for the $L'$ deepest layers, which can change from layer to layer. When determining each layer's threshold (at the beginning of each epoch, as in the regular homogeneous algorithm) if the layer is in the $L'$ deepest layers the algorithm checks if its expected cosine similarity is greater than the minimum cosine similarity defined for that layer. If it is not, then the threshold is decreased so to achieve the minimum cosine similarity defined for that layer. After going through the $L'$ deepest layers, that are first in the backward-pass, we know what was the sparsity for each layer so far. because we preserved the cosine similarity, the sparsity of these layers is expected to be lower than the required overall sparsity. In order to compensate for that, we can set a higher sparsity level for the rest of the layers, for which we don't define a minimum cosine similarity. This higher sparsity level can be calculated based on the sparsity of the first $L'$ layers, which we already know, and the total number of elements in them compared to the rest of the layers. \\
In order to refrain from over-pruning the shallow layers we set a maximum sparsity level for the layers for which we don't set a minimum cosine similarity level. From our experiments we find that setting a maximum of 0.96-0.98 will prevent over-pruning of these layers, which might deteriorate the final accuracy. Of course, using this maximum sparsity while aiming for both very high overall sparsity levels and high minimum cosine similarity for many of the layers might prevent us from reaching the desired overall sparsity. However, for most use-cases this doesn't pose significant restrictions on the parameters and at worse causes fairly small changes in overall sparsity.\\
Typical values we use to achieve high sparsity levels while preserving baseline accuracy are 0.95-0.98 minimum cosine similarity for the higher layers, where usually the deeper layers will receive higher minimum cosine similarities. For example, using ResNet18 on ImageNet, we have 4 basic blocks, we might preserve the last two with a minimum cosine similarity of 0.98 and the second one with 0.95, leaving the first block and the single layer before it, to compensate by having a higher sparsity level. In \cref{fig:sparsityVsEpochHetrogeneous} we show an example of the different sparsity levels between the different layers when using heterogeneous pruning on ResNet18 trained on ImageNet.   

\subsubsection{Neural gradients lognormal distributions are truncated}
\label{Gumbel}
When evaluating the cosine similarity of the different layers we noticed that the cosine similarity measure is very sensitive to the large valued components of the tensor, that is, pruning the same tensor after removing a few of the largest magnitude components will cause a far lower cosine similarity at the same sparsity level. This is because the cosine similarity represents the change in a high-dimensional angle, that is governed by the largest components.\\
Having first developed the equations in \cref{sec:allocationAppendix} for a general lognormal distribution, we found a significant discrepancy between the expected and actual cosine similarities. That leads us to the understanding that the neural gradients' distributions actually differ slightly from the lognormal distribution by the lack of high-value components in the abundance that we might expect from tensors of their size where each value is drawn from a lognormal distribution. We verified and quantified this observation by examining the Gumbel distribution, that models the distribution of the maximum obtained from drawing a certain number of samples from a base distribution, in this case the lognormal distribution. We found that in fact the maximum values of the gradients were significantly lower than the ones expected from the Gumbel distribution of the corresponding lognormal distribution and size of the tensor. In \cref{fig:Gumbel} we show examples of the actual maximum values of the tensors compared to the expected values using the Gumbel distribution.\\
Finally, we re-modeled the distribution of the gradients as a truncated lognormal distribution, that does not receive values above a certain value, denoted as $k$ times the std $\sigma$. The value $k$ is obtained from the distribution itself, like $\sigma$ and $\mu$. We found that taking the estimated $k$ as the maximum of the obtained tensor makes a very noisy estimator, and does not match the actual cosine similarity closely, so in practice we take a high quantile (for example 0.997) of the distribution. It is important to note that this slight difference from the lognormal distribution is highly important when calculating the cosine similarity, but is negligible when dealing with sparsity, for example, because it is not sensitive to the lack of a few high-valued components.  

\subsubsection{Cosine Similarity of Stochastic pruning - full derivation}
\label{sec:appendStochasticPrunDetailed}
In this section we extend on the full derivation of the cosine similarity for stochastic pruning.

To calculate the expected similarity between the $n$-dimensional tensors $X$ and $T_{\alpha,\varepsilon}(X)$, we need to calculate $\mathbb{E}[X\cdot T(X)]$, $\mathbb{E}||X||_2$, and $\mathbb{E}||T_{\alpha,\varepsilon}\left (X\right)||_2 $, as follows:

\begin{equation}
    \mathbb{E}[X\cdot T_{\alpha,\varepsilon}\left (X\right)] = \mathbb{E} \left [ \sum_i x_i\cdot T_{\alpha,\varepsilon}\left (x_i\right) \right] =  \sum_i \mathbb{E}\left[x_i\cdot T_{\alpha,\varepsilon}\left (x_i\right) \right]= n\cdot \mathbb{E}\left[x\cdot T_{\alpha,\varepsilon}\left (x\right) \right] 
\label{2}
\end{equation}

And similary:
\begin{equation}
    \mathbb{E}||X||_2 = n\cdot \mathbb{E}\left[x \right]^2 
\label{2}
\end{equation}
\begin{equation}
    \mathbb{E}||T_{\alpha,\varepsilon}\left (X\right)||_2 = n\cdot \mathbb{E}\left[T_{\alpha,\varepsilon}\left (x\right) \right]^2 
\label{2}
\end{equation}

Assuming  $X \sim \text{LogNormal}(0,\sigma^2)$ with maximum value $k\sigma$, the cosine similarity between $X$ and $T_{\alpha,\varepsilon}\left (X\right)$ is : 
\begin{equation}
\begin{split}
     \mathbb{E}||X||_2^2 & =  \frac{1}{2}n\cdot e^{2\sigma^2} \left[ 1 + \mathrm{erf}\left(\frac{\ln(k\sigma)-2\sigma^2}{\sigma\sqrt{2}}\right) \right]
\end{split}
\end{equation}
\begin{equation}
\begin{split}
     \mathbb{E}||T(X)||_2^2 & = n \int_0^1\int_{\varepsilon\alpha}^{\alpha} \alpha^2\cdot  f(x) \,dx \,d\varepsilon + n \int_{\alpha}^{k\sigma} x^2 \cdot f(x) \,dx \\
      &= n\int_0^1\int_{\varepsilon\alpha}^{\alpha} \frac{\alpha^2}{x\sigma\sqrt{2\pi}}e^{-\frac{\ln(x)^2}{2\sigma^2}} \,dx \,d\varepsilon+ n\int_{\alpha}^{k\sigma} \frac{x}{\sigma\sqrt{2\pi}}e^{-\frac{\ln(x)^2}{2\sigma^2}} \,dx  \\ 
    &= \frac{1}{2}n\cdot\alpha^2 \int_0^1 \left[\mathrm{erf}\left(\frac{\ln(\alpha)}{\sigma\sqrt{2}}\right) - \mathrm{erf}\left(\frac{\ln(\varepsilon\alpha)}{\sigma\sqrt{2}}\right)\right] \,d\varepsilon \\ &+ \frac{1}{2}n\cdot e^{2\sigma^2} \left[\mathrm{erf}\left(\frac{\ln(k\sigma)-2\sigma^2}{\sigma\sqrt{2}}\right) -\mathrm{erf}\left(\frac{\ln(\alpha)-2\sigma^2}{\sigma\sqrt{2}}\right)\right] \\ &= \frac{1}{2}n\cdot\alpha^2\cdot\mathrm{erf}\left(\frac{\ln(\alpha)}{\sigma\sqrt{2}}\right) \\ &-\frac{1}{2}n\cdot \left[\alpha \cdot e^{\frac{\sigma^2}{2}}\mathrm{erf}\left(\frac{\sigma^2 - \ln(\varepsilon\alpha)}{\sigma\sqrt{2}}\right) + \alpha^2\varepsilon\cdot \mathrm{erf}\left(\frac{\ln(\varepsilon\alpha)}{\sigma\sqrt{2}}\right)\right] \Big|_{0}^{1} \\ &+ \frac{1}{2}n\cdot e^{2\sigma^2} \left[\mathrm{erf}\left(\frac{\ln(k\sigma)-2\sigma^2}{\sigma\sqrt{2}}\right) -\mathrm{erf}\left(\frac{\ln(\alpha)-2\sigma^2}{\sigma\sqrt{2}}\right)\right] \\
    &= \frac{1}{2}n\cdot\alpha^2\cdot \mathrm{erf}\left(\frac{\ln(\alpha)}{\sigma\sqrt{2}}\right) \\ &-\frac{1}{2}n\cdot \left[\alpha \cdot e^{\frac{\sigma^2}{2}}\mathrm{erf}\left(\frac{\sigma^2 - \ln(\alpha)}{\sigma\sqrt{2}}\right) + \alpha^2\cdot \mathrm{erf}\left(\frac{\ln(\alpha)}{\sigma\sqrt{2}}\right) - \alpha\cdot e^{\frac{\sigma^2}{2}} \right]  \\ &+ \frac{1}{2}n\cdot e^{2\sigma^2} \left[\mathrm{erf}\left(\frac{\ln(k\sigma)-2\sigma^2}{\sigma\sqrt{2}}\right) -\mathrm{erf}\left(\frac{\ln(\alpha)-2\sigma^2}{\sigma\sqrt{2}}\right)\right] \\ 
    &= \frac{1}{2}n\cdot \alpha \cdot e^{\frac{\sigma^2}{2}}\left[1 - \mathrm{erf}\left(\frac{\sigma^2 - \ln(\alpha)}{\sigma\sqrt{2}}\right) \right] \\ &+ \frac{1}{2}n\cdot e^{2\sigma^2} \left[\mathrm{erf}\left(\frac{\ln(k\sigma)-2\sigma^2}{\sigma\sqrt{2}}\right) -\mathrm{erf}\left(\frac{\ln(\alpha)-2\sigma^2}{\sigma\sqrt{2}}\right)\right] 
\end{split}
\label{eq:TX}
\end{equation}

\begin{equation}
\begin{split}
    \mathbb{E}[X\cdot T(X)] &=   n\int_0^1\int_{\varepsilon\alpha}^{\alpha} \alpha \cdot x\cdot  f(x) \,dx \,d\varepsilon \cdot  + n \int_{\alpha}^{k\sigma} x^2 \cdot f(x)\,dx\\
       &= \int_0^1\int_{\varepsilon\alpha}^{\alpha} \frac{\alpha}{\sigma\sqrt{2\pi}}e^{-\frac{\ln(x)^2}{2\sigma^2}} \,dx \,d\varepsilon  + n\int_{\alpha}^{k\sigma} \frac{x}{\sigma\sqrt{2\pi}}e^{-\frac{\ln(x)^2}{2\sigma^2}} \,dx  \\ 
       &\underset{y = \ln(x)}{=} n\int_0^1\int_{\ln(\varepsilon\alpha)}^{\ln(\alpha)} \frac{\alpha}{\sigma\sqrt{2\pi}} e^{-\frac{y^2}{2\sigma^2}}e^y \,dy \,d\varepsilon + n\int_{\ln(\alpha)}^{\ln(k\sigma)} \frac{1}{\sigma\sqrt{2\pi}} e^{-\frac{y^2}{2\sigma^2}}e^{2y} \,dy \\
      &= n\int_0^1\int_{\ln(\varepsilon\alpha)}^{\ln(\alpha)} \frac{\alpha}{\sigma\sqrt{2\pi}} e^{-\frac{(y-\sigma^2)^2}{2\sigma^2}}e^{\frac{\sigma^2}{2}} \,dy \,d\varepsilon+ n\int_{\ln(\alpha)}^{\ln(k\sigma)} \frac{1}{\sigma\sqrt{2\pi}} e^{-\frac{(y-2\sigma^2)^2}{2\sigma^2}}e^{2\sigma^2} \,dy \\
      &= n\cdot\alpha\cdot e^{\frac{\sigma^2}{2}}\cdot\frac{1}{2}\int_0^1\left(1 + \mathrm{erf}\left(\frac{y-\sigma^2}{\sigma\sqrt{2}}\right)\right) \Big|_{\ln(\varepsilon\alpha)}^{\ln(\alpha)} \,d\varepsilon \\ &+ n\cdot e^{2\sigma^2}\cdot\frac{1}{2}\left(1 +  \mathrm{erf}\left(\frac{y-2\sigma^2}{\sigma\sqrt{2}}\right)\right) \Big|_{\ln(\alpha)}^{\ln(k\sigma)} \\
      &=\frac{1}{2}n\cdot e^{\frac{\sigma^2}{2}}\cdot\alpha\int_0^1\left[ \mathrm{erf}\left(\frac{\ln(\alpha)-\sigma^2}{\sigma\sqrt{2}}\right) -  \mathrm{erf}\left(\frac{\ln(\varepsilon\alpha)-\sigma^2}{\sigma\sqrt{2}}\right)\right] \,d\varepsilon \\
      &+ \frac{1}{2}n\cdot e^{2\sigma^2} \left[\mathrm{erf}\left(\frac{\ln(k\sigma)-2\sigma^2}{\sigma\sqrt{2}}\right) -\mathrm{erf}\left(\frac{\ln(\alpha)-2\sigma^2}{\sigma\sqrt{2}}\right)\right] \\
      &= \frac{1}{2}n\cdot e^{\frac{\sigma^2}{2}}\cdot\alpha \cdot \mathrm{erf}\left(\frac{\ln(\alpha)-\sigma^2}{\sigma\sqrt{2}}\right) \\ &-  \frac{1}{2}n\cdot e^{\frac{\sigma^2}{2}} \left[ e^{\frac{3\sigma^2}{2}}\mathrm{erf}(\frac{2\sigma^2-\ln(\alpha\varepsilon)}{\sqrt{2}\sigma}) - \alpha\varepsilon\cdot \mathrm{erf}(\frac{\sigma^2 - \ln(\varepsilon\alpha)}{\sqrt{2}\sigma}) \right] \Big|_{0}^{1}\\
        &+ \frac{1}{2}n\cdot e^{2\sigma^2} \left[\mathrm{erf}\left(\frac{\ln(k\sigma)-2\sigma^2}{\sigma\sqrt{2}}\right) -\mathrm{erf}\left(\frac{\ln(\alpha)-2\sigma^2}{\sigma\sqrt{2}}\right)\right] \\
      &= \frac{1}{2}n\cdot e^{\frac{\sigma^2}{2}}\cdot\alpha \cdot \mathrm{erf}\left(\frac{\ln(\alpha)-\sigma^2}{\sigma\sqrt{2}}\right) \\ &-  \frac{1}{2}n\cdot e^{\frac{\sigma^2}{2}} \left[ e^{\frac{3\sigma^2}{2}}\mathrm{erf}(\frac{2\sigma^2-\ln(\alpha)}{\sqrt{2}\sigma}) - \alpha\cdot \mathrm{erf}(\frac{\sigma^2 - \ln(\alpha)}{\sqrt{2}\sigma}) - e^{\frac{3\sigma^2}{2}} \right]\\
        &+ \frac{1}{2}n\cdot e^{2\sigma^2} \left[\mathrm{erf}\left(\frac{\ln(k\sigma)-2\sigma^2}{\sigma\sqrt{2}}\right) -\mathrm{erf}\left(\frac{\ln(\alpha)-2\sigma^2}{\sigma\sqrt{2}}\right)\right] \\
      &= \frac{1}{2}n\cdot e^{2\sigma^2}\left[ 1 - \mathrm{erf}(\frac{2\sigma^2-\ln(\alpha)}{\sqrt{2}\sigma})\right] \\ &+ \frac{1}{2}n\cdot e^{2\sigma^2} \left[1 -\mathrm{erf}\left(\frac{\ln(\alpha)-2\sigma^2}{\sigma\sqrt{2}}\right)\right] \\
      &=   \frac{1}{2}n\cdot e^{2\sigma^2} \left[ 1 + \mathrm{erf}\left(\frac{\ln(k\sigma)-2\sigma^2}{\sigma\sqrt{2}}\right) \right]
\end{split}
\label{eq:X_TX}
\end{equation}

\subsection{Encoding}
\label{sec:encodingAppendix}

\begin{figure}[h]
\centering
\begin{subfigure}{.48\textwidth}

  \centering
  \includegraphics[width=\linewidth]{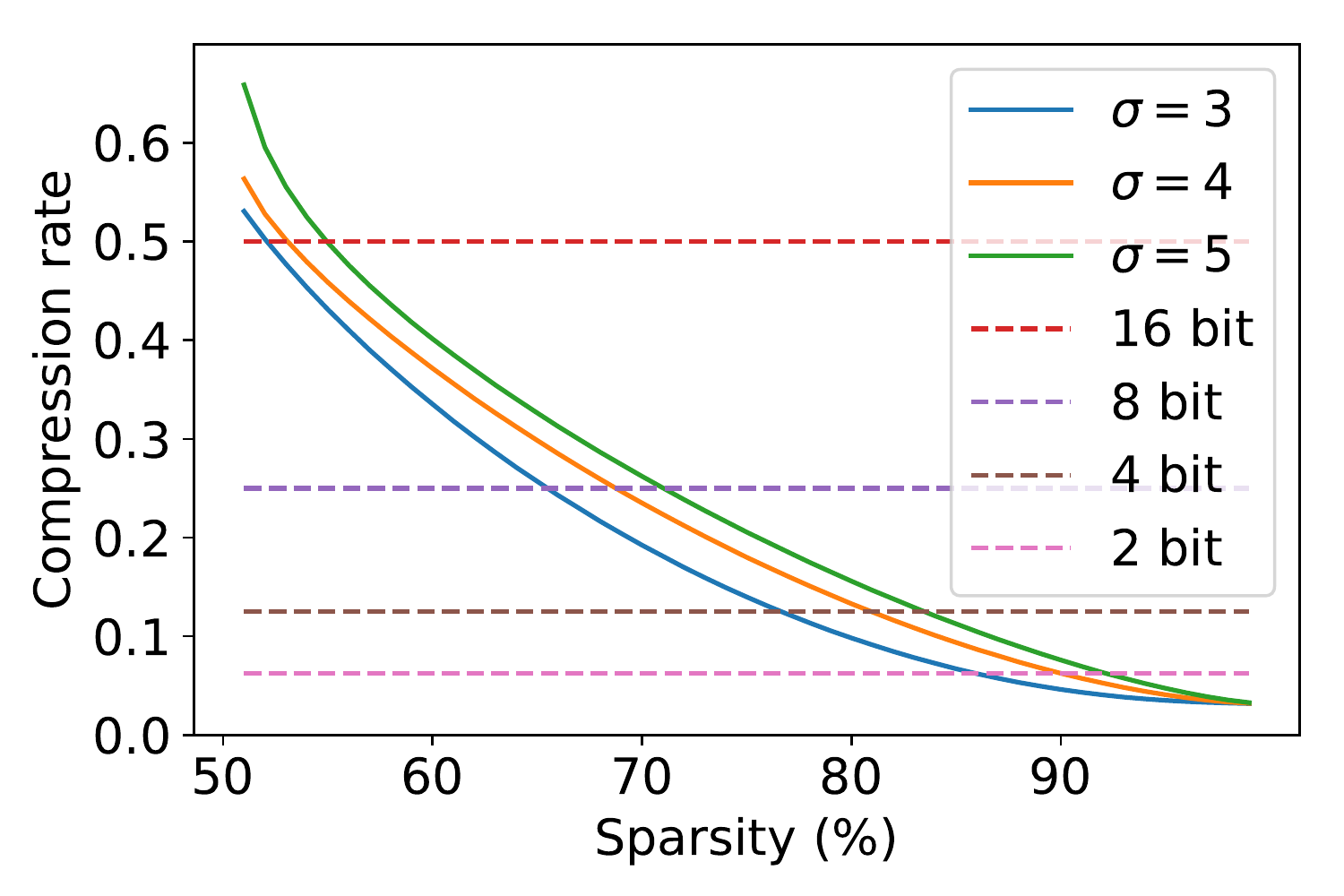}  
\caption{}
\label{fig:compressionSynthetic}
\end{subfigure}
\begin{subfigure}{.48\textwidth}

  \centering
  \includegraphics[width=\linewidth]{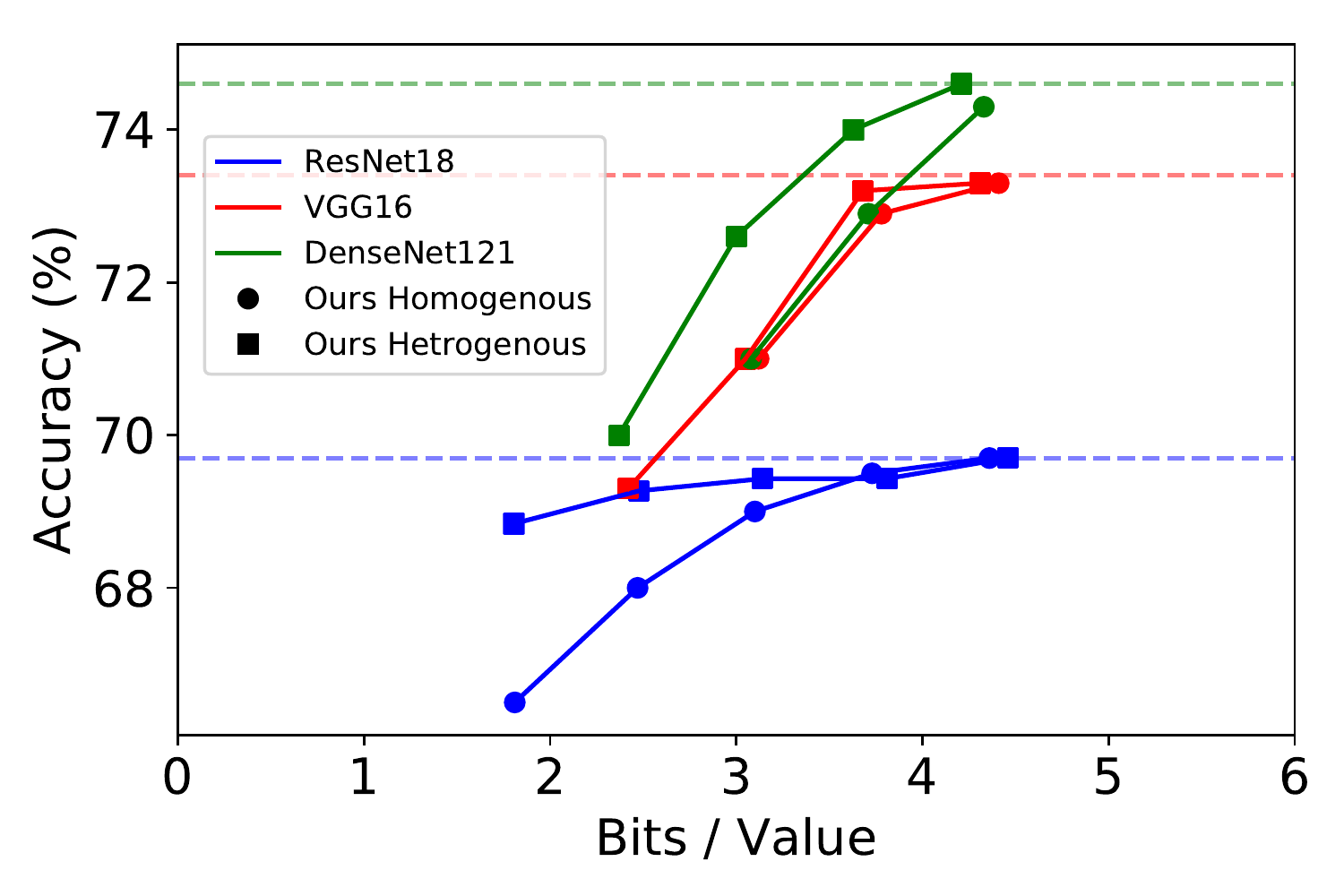}  
\caption{}
\label{fig:comparisoncompression}
\end{subfigure}
\caption{(a) Compression rate achieved for the proposed encoding for a synthetic lognormal distributed tensor for common gradients $\sigma$ and the comparison to quantization to different bitwidth. Notice for 90\% sparsity the compression rate is similar to quantization to 2 bits. (b) Bits for each value using the proposed encoding method for the results in \cref{fig:Comparison}. }
\end{figure}

We suggest a custom encoding method that fully utilizes the abundance of both zeros and $\pm \alpha$ generated by the stochastic pruning, at their relative frequency. As shown in \cref{fig:compressionSynthetic}  the compression ratio achieved by the proposed encoding, relative to original FP32, is equivalent to quantizing to 4 bits at 80\% sparsity, and to only 2 bits at 90\%. In \cref{fig:comparisoncompression} we show the actual bits per value for the results shown in \cref{fig:results} using the proposed encoding.

\begin{table}[h]
\centering
\caption{Simple proposed encoding for the values after stochastic pruning}
\begin{tabular}{c c c}
\toprule
Value & encoding \\ \toprule
0 & 0\\
$\alpha$ & 100\\
$-\alpha$ & 101\\
starting a FP value & 11\\
\bottomrule
\end{tabular}
\label{tab:bias}
\end{table}

\subsection{Layer gradients stability}

\begin{figure}[h]
\begin{subfigure}{.9\textwidth}
  \centering
  \includegraphics[width=\linewidth]{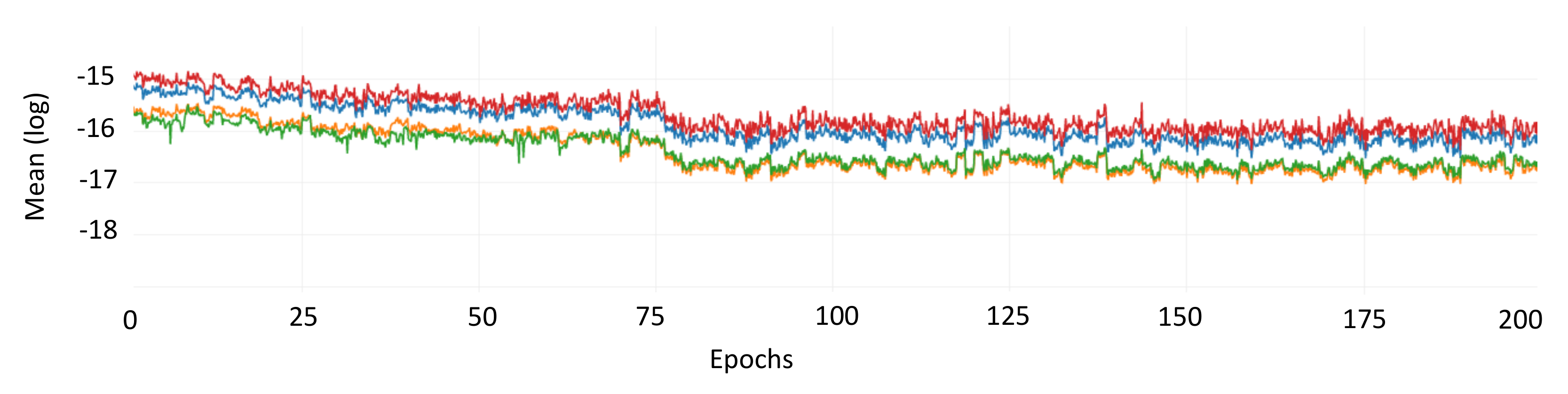}  
  \caption{}
  \label{}
\end{subfigure}

\begin{subfigure}{.9\textwidth}
  \centering
  \includegraphics[width=\linewidth]{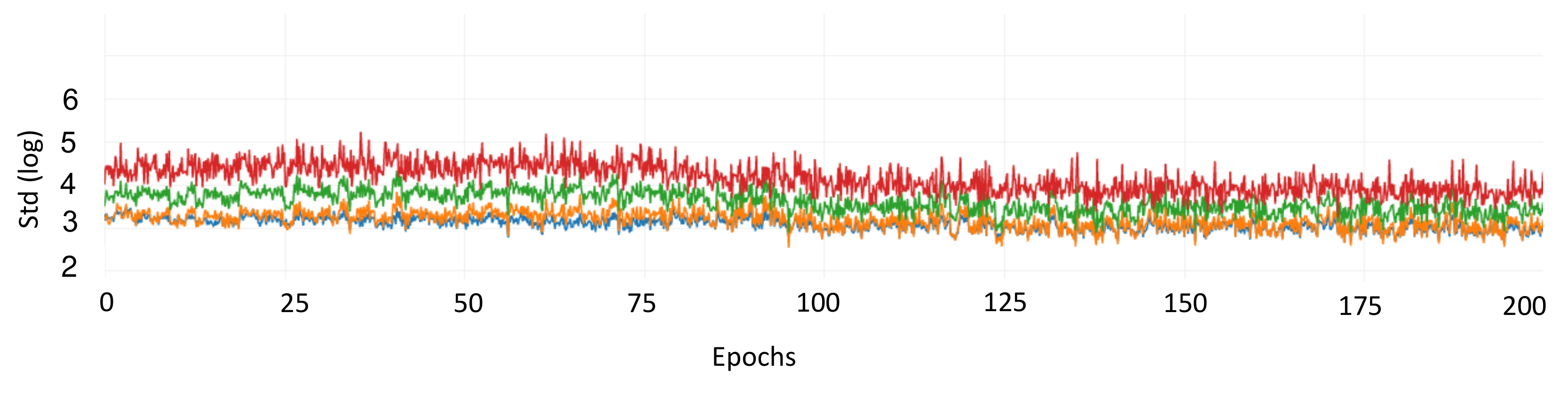}  
  \caption{}
  \label{fig:cifar100std}
\end{subfigure}
\caption{ The gradients mean (a) and std (b) of ResNet-18 , Cifar100 dataset in different layers. Notice they are stable across each epoch - so there is the posibility to calculate the threshold for the required sparsity once in an epoch }
\label{fig:actualMeanStd}
\end{figure}

\begin{figure}[h]
\begin{subfigure}{.5\textwidth}

  \centering
  \includegraphics[width=\linewidth]{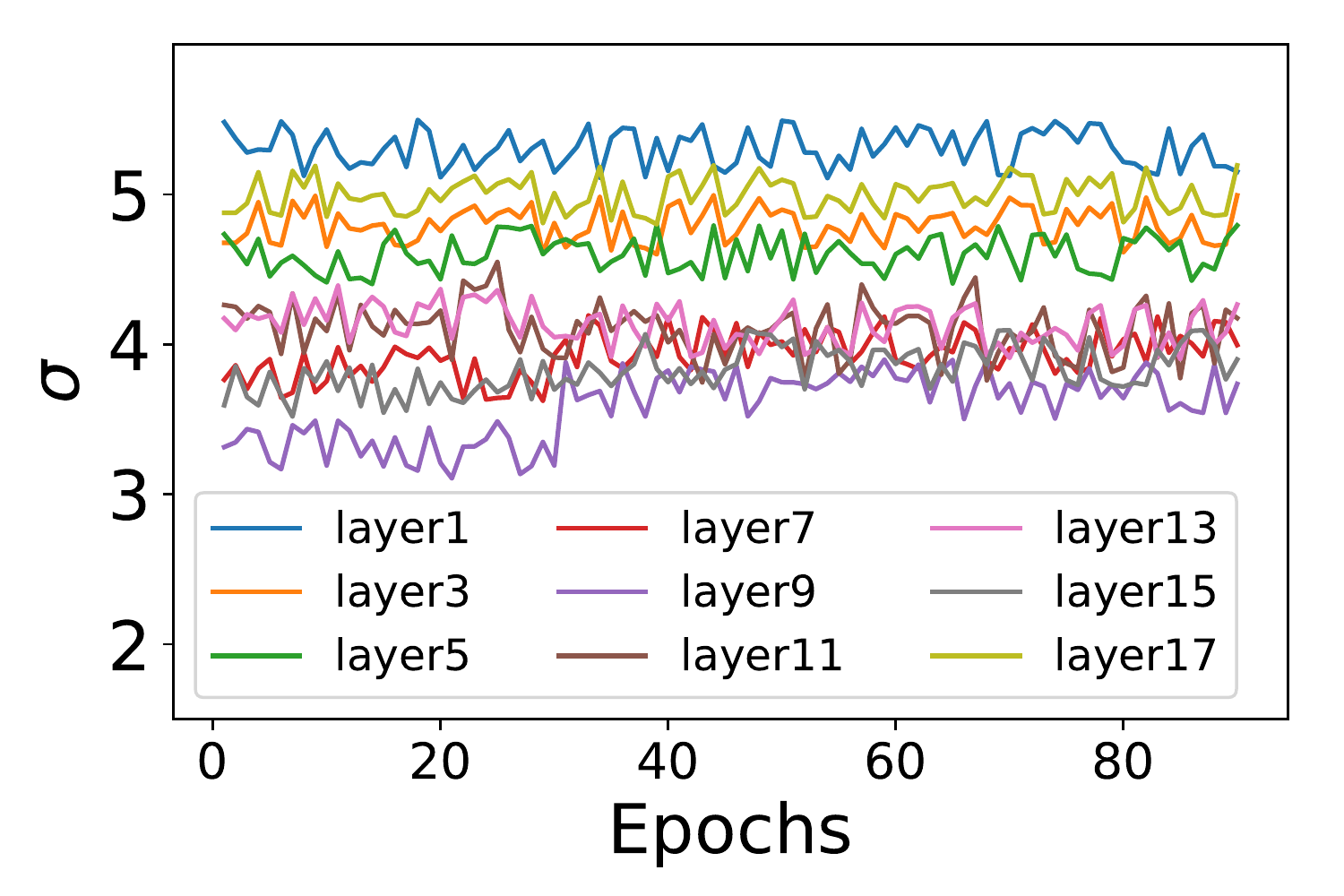}  
  \caption{}
  \label{fig:imagenetStd}
  \end{subfigure}
\begin{subfigure}{.5\textwidth}

  \centering
  \includegraphics[width=\linewidth]{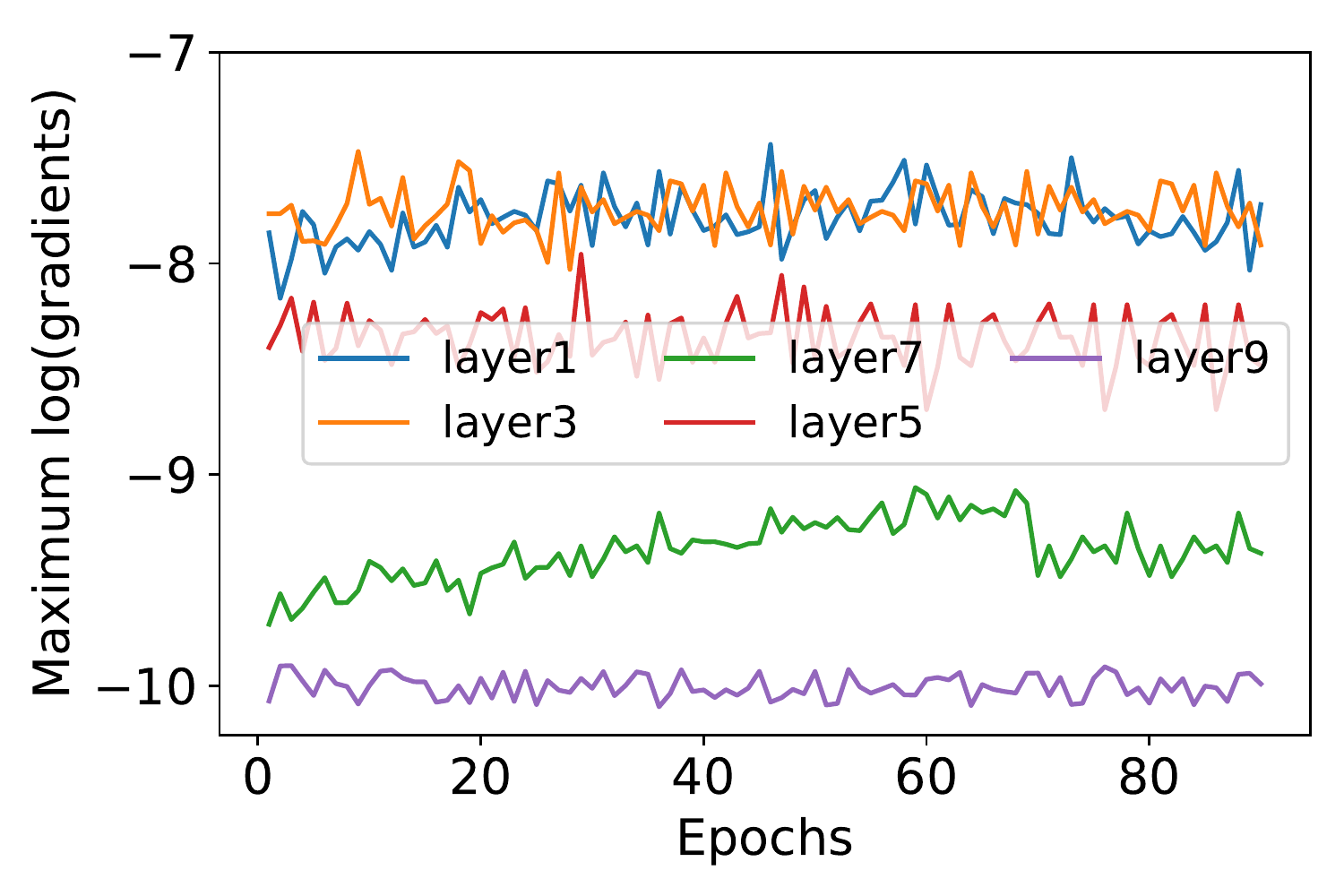}  
  \caption{}
  \label{fig:imagenetMax}
  \end{subfigure}
\caption{\textbf{(a):} $\sigma$ of different layers throughout training (ResNet18 on ImageNet). \textbf{(b):} Maximum of the neural gradient at log scale throughout training (ResNet18 on ImageNet).}
\end{figure}

The measure of mean and std in log scale of the layer gradients in ResNet18, cifar100 dataset is shown in \cref{fig:actualMeanStd}. Notice the stability inside each epoch, that allows us to sample them once in an epoch to calculate the corresponding threshold.

\subsection{Experiments}
\label{sec:ap_exp}
We implement all the experiments in PyTorch , models are trained on ImageNet dataset \citep{deng2009imagenet} and are evaluated on three different architectures ResNet18 \citep{resnet}, DenseNet121 \citep{densenet} and VGG16 \citep{vgg}. We train the models with the standard regime for 90 epochs, using SGD with momentum, reducing the learning rate by a factor of $\frac{1}{10}$ after 30 and 60 epochs. 
Note the difference between the ATP and our homogeneous method is only the way to find the threshold $\alpha$ for a required sparsity. Therefore, we believe the difference in the accuracy for the same sparsity (we had better results) is probably the result of implementation issues. Our heterogeneous method further improves the validation accuracy for a given sparsity level, as shown in \cref{fig:Comparison}.
Lastly, we note that ATP \citep{acceleratedSpars2019} for some reason only reported training accuracy, but not to be too strict (it might have been just a typo) we assumed that their validation accuracy was the same as the training accuracy --- as was in our experiments.

\begin{figure}[h]
  \centering
  \includegraphics[width=0.5\linewidth]{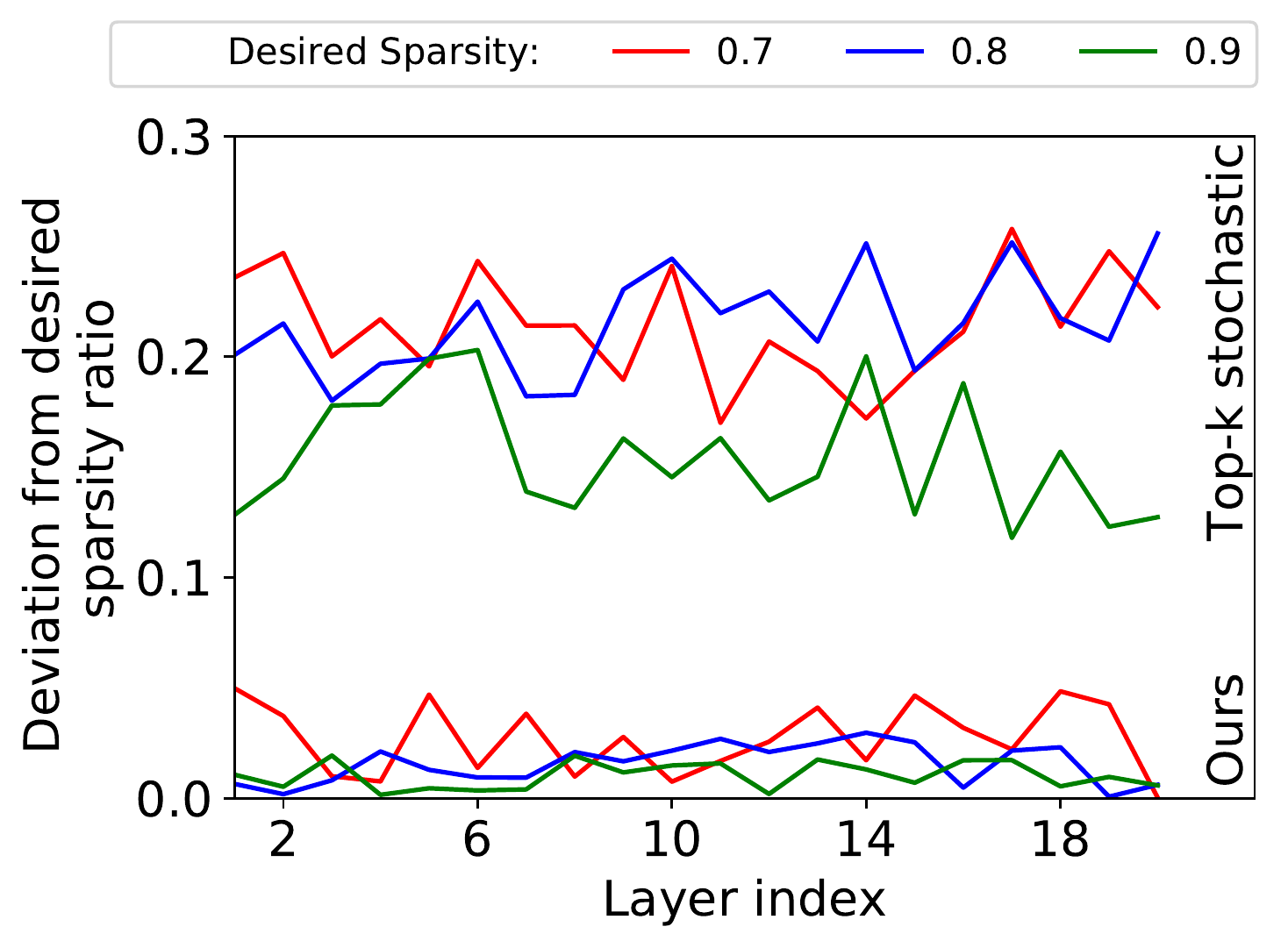}  
  \caption{}
\caption{Comparison of the deviation from the desired sparsity ratio of stochastic pruning Vs "top-k" + stochastic pruning, i.e finding the threshold $\alpha$ with "top-k" and then applying stochastic pruning. ( \cref{StochasticPruning}) in different layers. Notice the high deviation in  "top-k" + stochastic pruning is uniform across the layers}
\label{fig:deviation}
\end{figure}

\begin{figure}[h]
\begin{subfigure}{.5\textwidth}
  \centering
  \includegraphics[width=\linewidth]{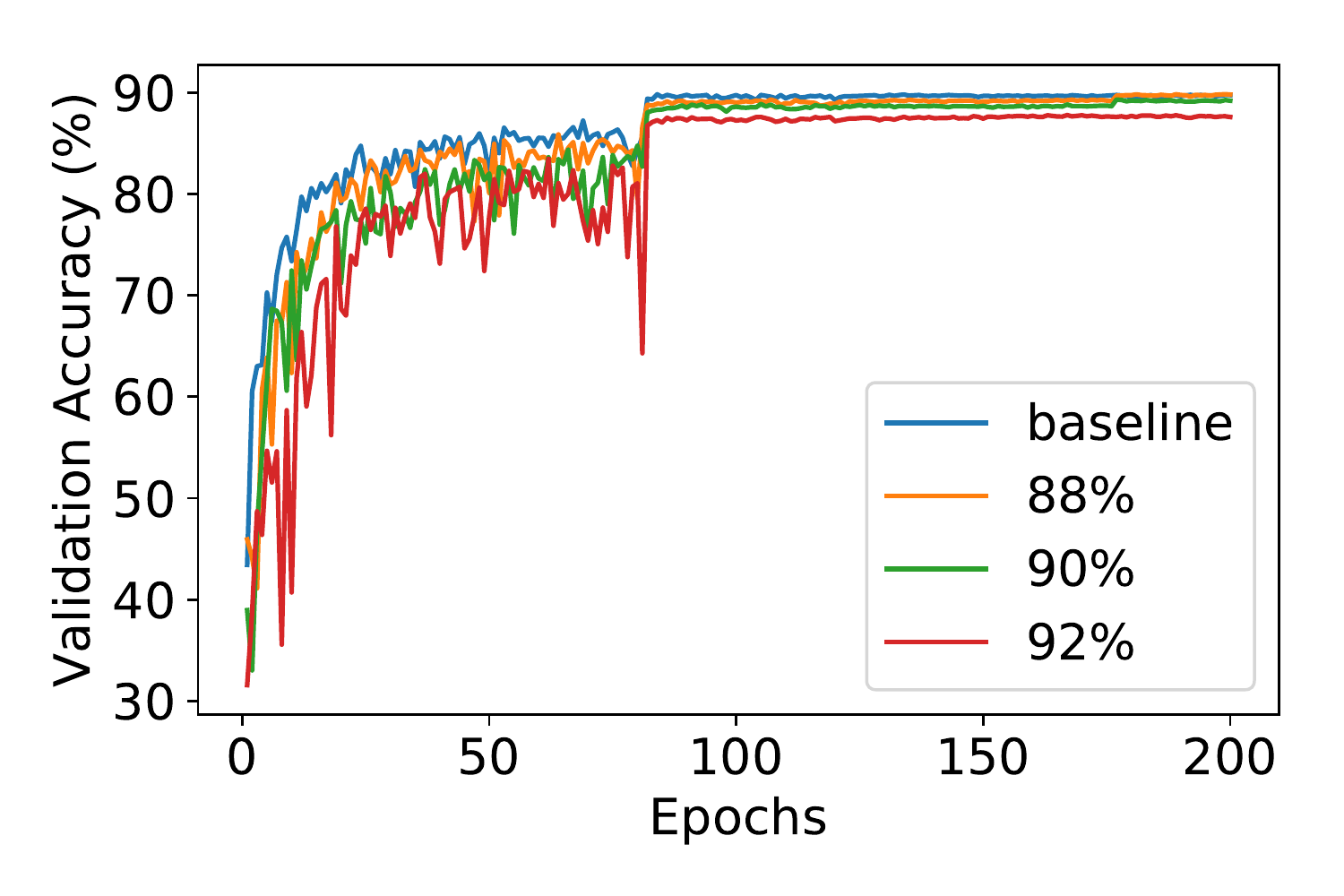}  
  \caption{}
  \label{}
\end{subfigure}
\begin{subfigure}{.5\textwidth}
  \centering
  \includegraphics[width=\linewidth]{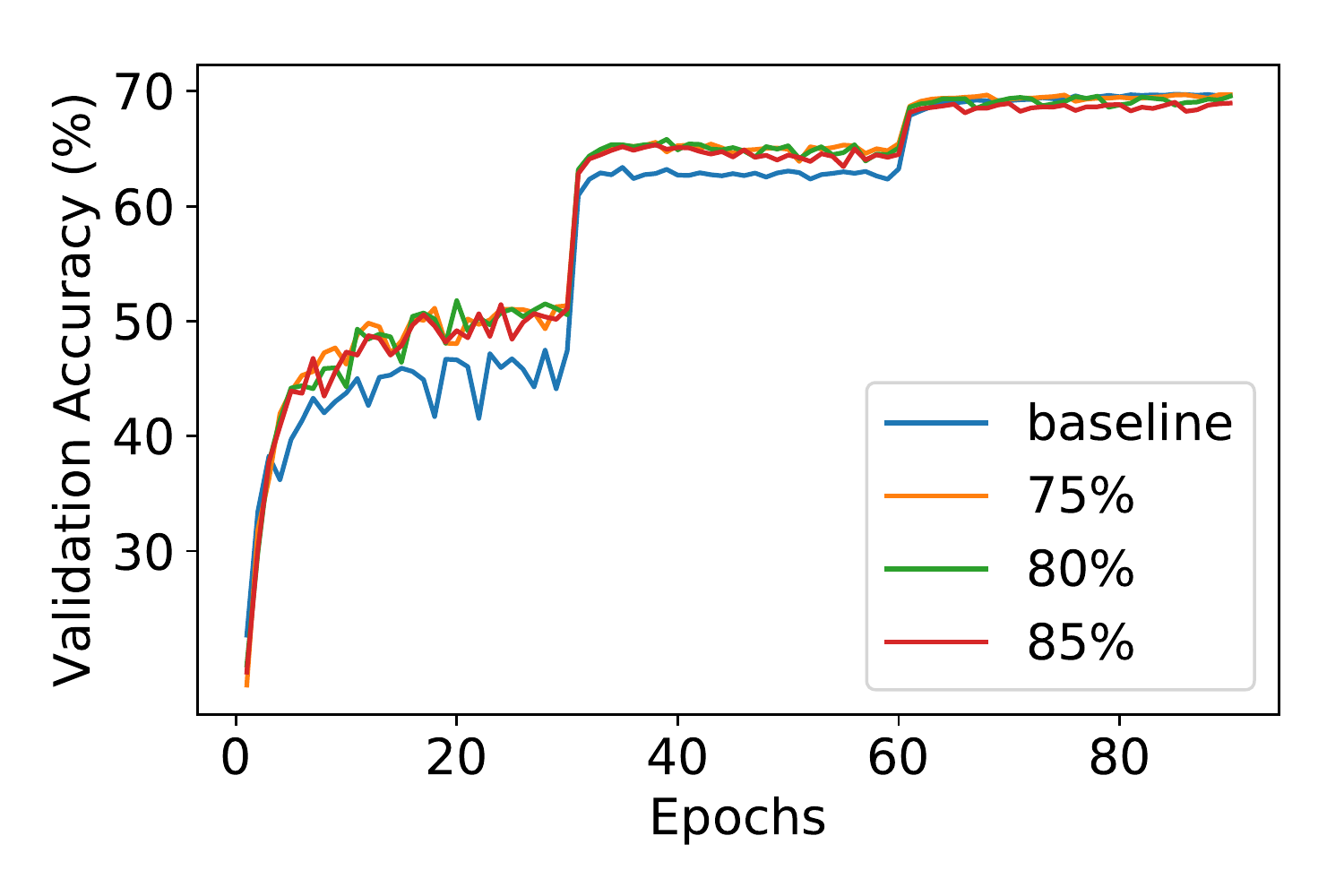}  
  \caption{}
  \label{fig:sub-second}
\end{subfigure}
\caption{ResNet18 training convergence with differente sparsity ratios for Cifar10(a) and ImagenNet(b) datasets. }
\label{fig:ImagenetTrain}
\end{figure}

\begin{figure}[h]
\begin{subfigure}{.5\textwidth}
  \centering
  \includegraphics[width=\linewidth]{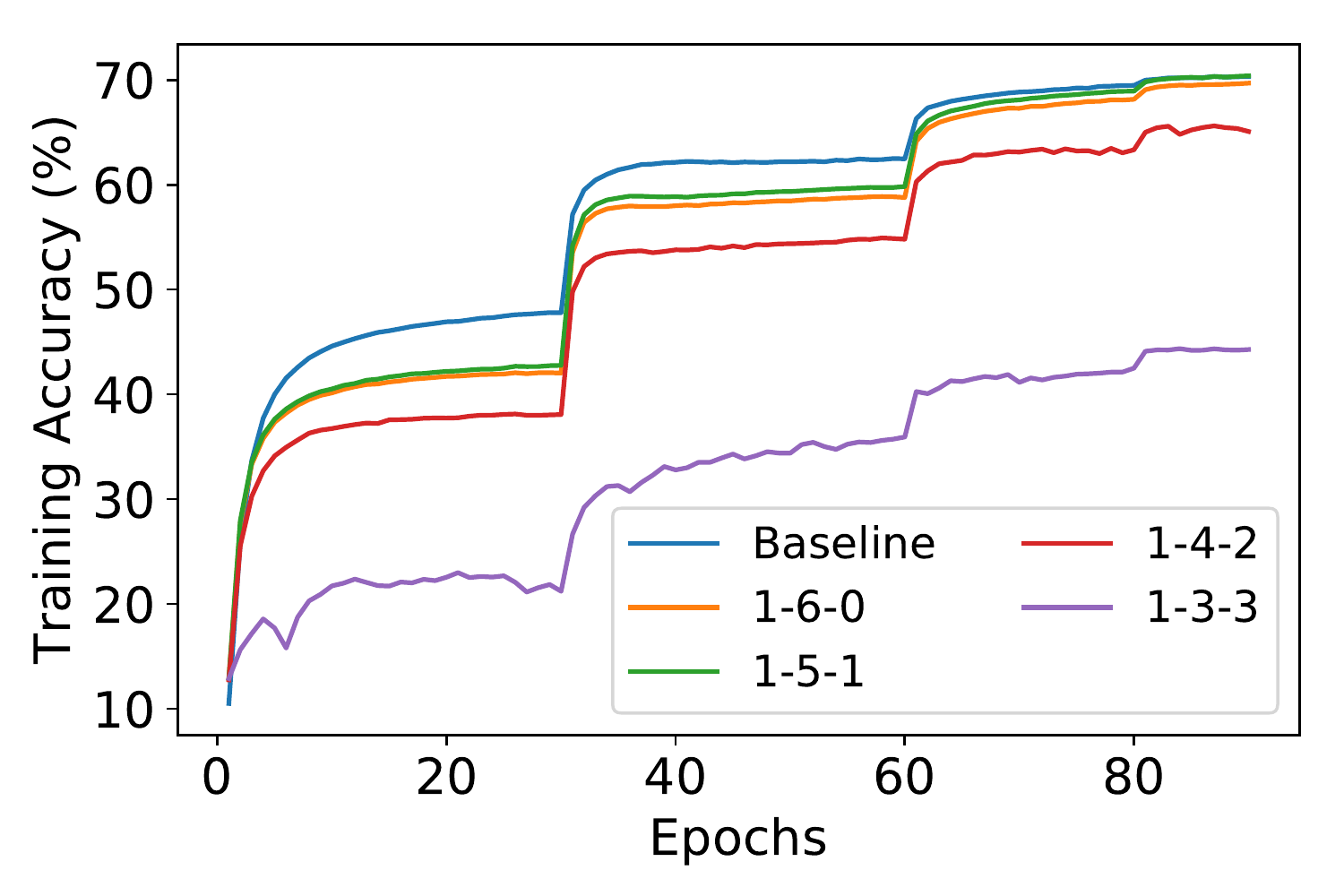}  
  \caption{}
  \label{fig:fp7ResultsTrain}
  \end{subfigure}
\begin{subfigure}{.5\textwidth}
  \centering
  \includegraphics[width=\linewidth]{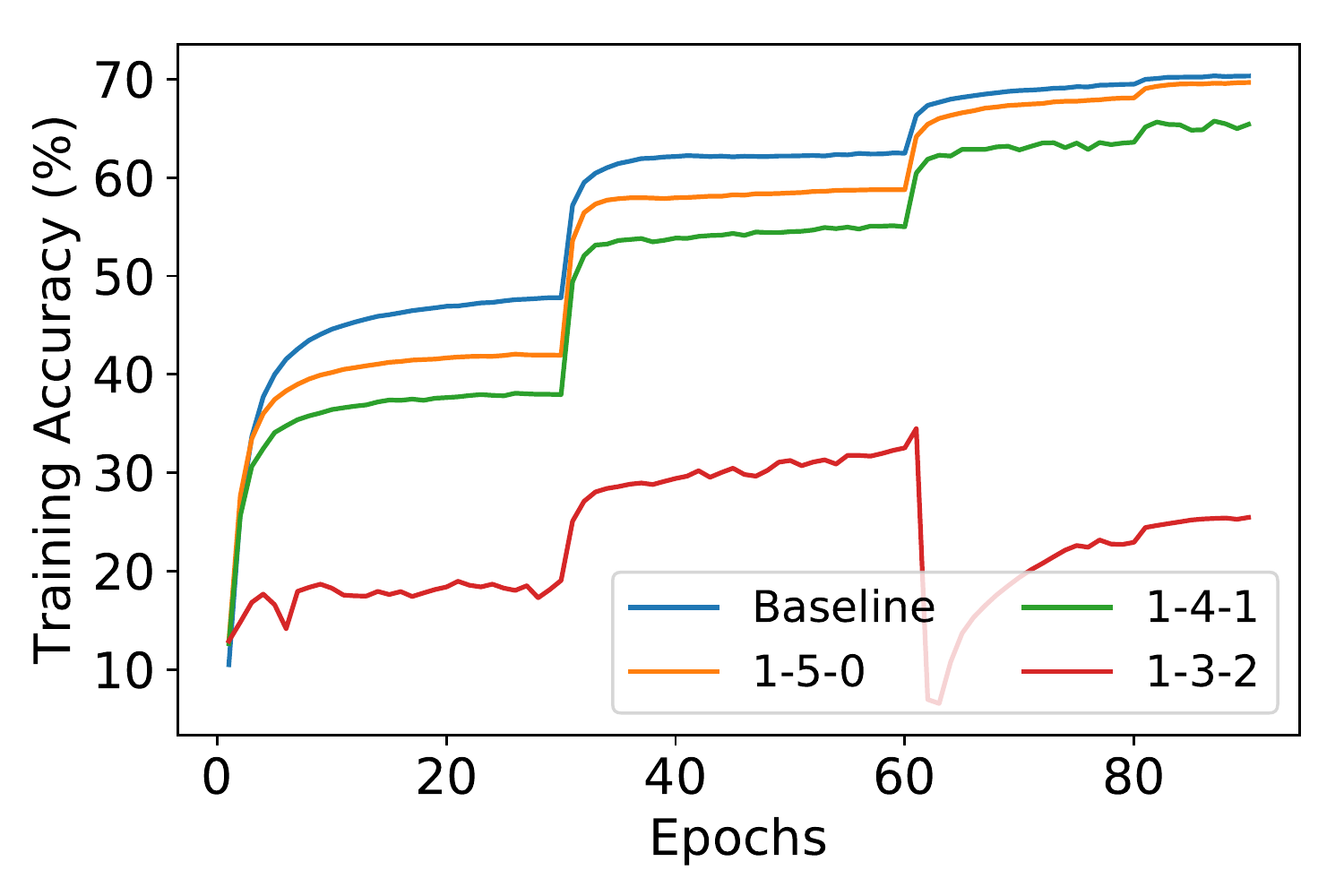}  
  \caption{}
  \label{fig:fp6ResultsTrain}
\end{subfigure}
\caption{FP7(a) and FP6(b) training convergence in different formats in ResNet18, ImageNet dataset. Notice the results fit the analysis in \cref{fig:expBits} }
\label{fig:fpTrain}
\end{figure}

\begin{figure}[h]
  
  \centering
  \includegraphics[width=0.6\linewidth]{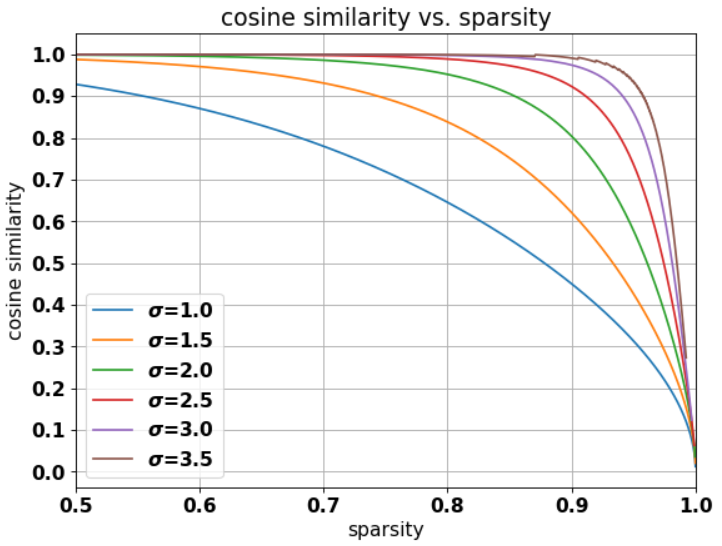}  
\caption{Cosine similarity vs. sparsity for different $\sigma$. For all distributions plotted here $k=2.5$. The values were calculated analytically using our analyses of the sparsity and cosine similarity using stochastic pruning. Notice that for the same sparsity values different distributions vary significantly in cosine similarity, this explains the difference in cosine similarity of the different layers when using homogeneous stochastic pruning (all layers having the same sparsity level).} 
\label{fig:cos_simVsSparsityTheoretical}
\vspace{-4mm}
\end{figure}

\begin{figure}[h]
  
  \centering
  \includegraphics[width=\linewidth]{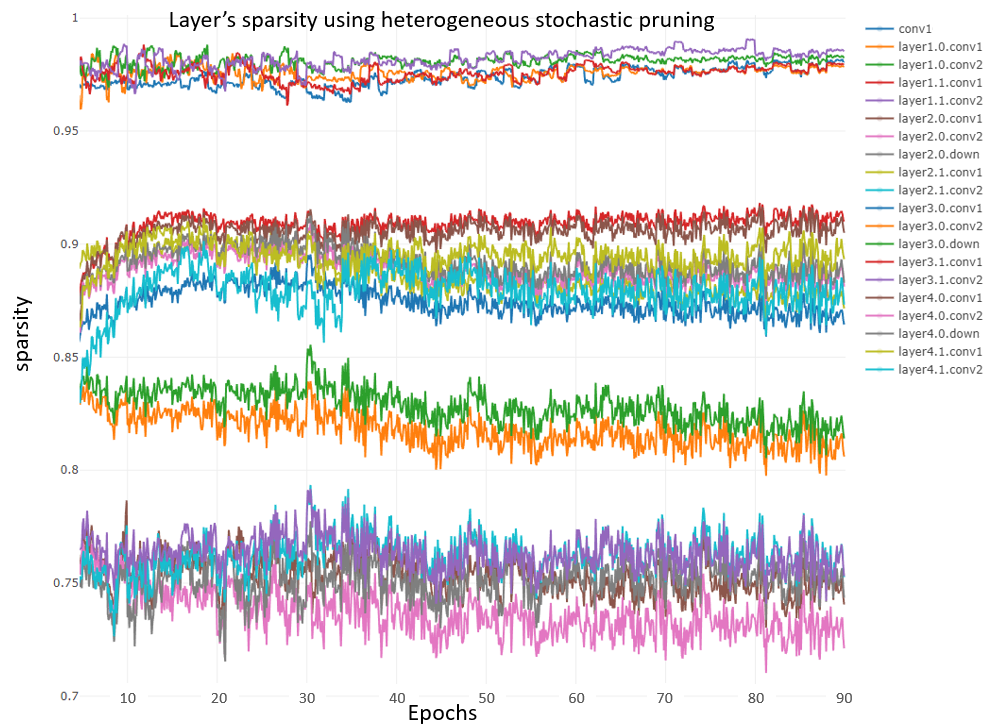}  
\caption{Actual sparsity of the different layers of ResNet18 trained on ImageNet using heterogeneous stochastic pruning. The desired sparsity was 92\% and the overall achieved one is 91.5\%, the minimum cosine similarity for blocks 3 \& 4 was 0.98 and for block 2 was 0.95, and the maximum allowed sparsity for the rest of the layers (block 1 and "conv1" before the first block) was 0.98. Notice that the high sparsity levels were induced on the first block and "conv1", and that some of the layers, especially from block 2, had much lower sparsity levels in order to preserve its cosine similarity. For this experiment the validation accuracy was less than 1\% degradation from baseline, compared to over 3\% degradation for the homogeneous algorithm at similar sparsity level.} 
\label{fig:sparsityVsEpochHetrogeneous}
\vspace{-4mm}
\end{figure}

\begin{figure}[h]
\centering
\begin{subfigure}{.48\textwidth}
  \centering
  \includegraphics[width=\linewidth]{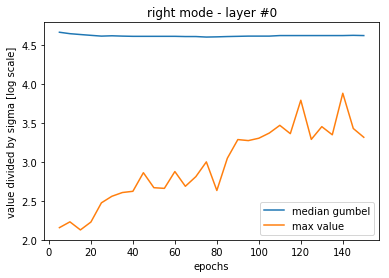}  
  \caption{}
  \label{}
\end{subfigure}
\begin{subfigure}{.48\textwidth}
  \centering
  \includegraphics[width=\linewidth]{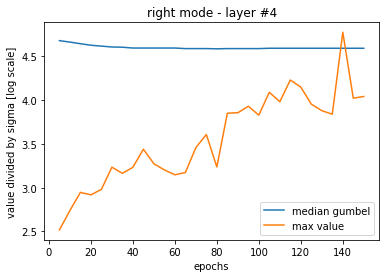}  
  \caption{}
  \label{}
\end{subfigure}
\begin{subfigure}{.48\textwidth}
  \centering
  \includegraphics[width=\linewidth]{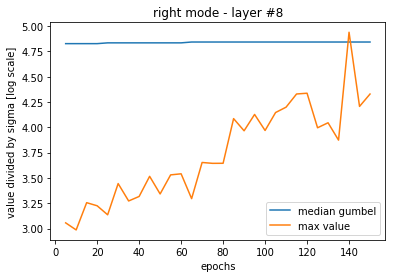}  
  \caption{}
  \label{}
\end{subfigure}
\begin{subfigure}{.48\textwidth}
  \centering
  \includegraphics[width=\linewidth]{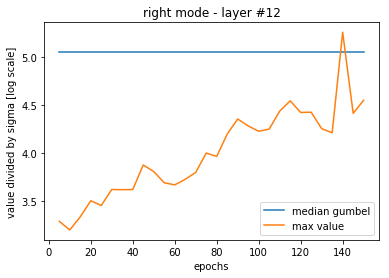}  
  \caption{}
  \label{}
\end{subfigure}
\caption{Comparison between the median value of the Gumbel distribution and the actual maximum value of the gradients for ResNet18 trained on Cifar10. The Gumbel distribution represents the distribution of the maximum value for a tensor of the size of the gradients tensor (each layer of different size) drawn from a lognormal distribution with the mean and std of the tensor. Here we use only the right mode of the bi-modal mixture of lognormal distributions. Notice that the actual maximum value is orders of magnitudes lower than the expected maximum value from the Gumbel distribution (y-axis is in log scale, and the value is divided by the std). The actual tensors are truncated at levels of approximately 3-5 $\sigma$. This observation lead us to model the distribution as a truncated lognormal distribution when we derived the cosine similarity analytically, because the cosine similarity is more sensitive to the presence of high-valued components in the tensor. The numbering of the layers here is from the deepest to the shallowest, according to the order of the gradients flow in the backward-pass. Notice that layer 12 gradients has twice as many elements as layer 8 that has twice as many as layers 4 and 0, thus the median of their Gumbel distributions is higher, measuring in sigmas, because if we draw more values we expect the maximum value to be larger. }
\label{fig:Gumbel}
\end{figure}

\begin{figure}[h]
\centering
\begin{subfigure}{.48\textwidth}
  \centering
  \includegraphics[width=\linewidth,trim={0 0 0 0cm}]{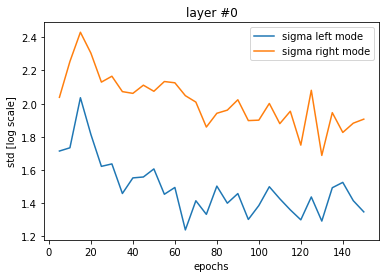}  
  \caption{}
  \label{}
\end{subfigure}
\begin{subfigure}{.48\textwidth}
  \centering
  \includegraphics[width=\linewidth,trim={0 0cm 0 0}]{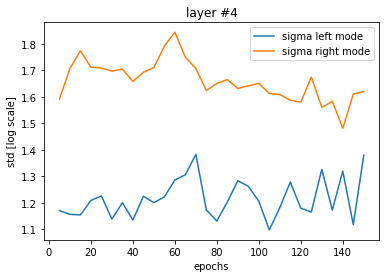}  
  \caption{}
  \label{}
\end{subfigure}
\begin{subfigure}{.48\textwidth}
  \centering
  \includegraphics[width=\linewidth,trim={0 0cm 0 0}]{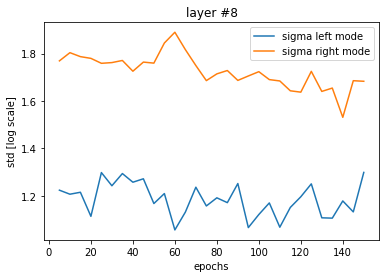}  
  \caption{}
  \label{}
\end{subfigure}
\begin{subfigure}{.48\textwidth}
  \centering
  \includegraphics[width=\linewidth,trim={0 0cm 0 0}]{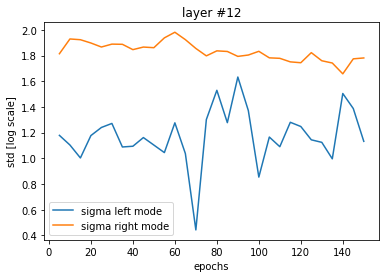}  
  \caption{}
  \label{}
\end{subfigure}
\caption{The std ($\sigma$) of each of the two modes of the gradients for ResNet18 trained on Cifar10, on various layers. Notice that the left mode is of lower variance. The numbering of the layers here is from the deepest to the shallowest, according to the order of the gradients flow in the backward-pass.}
\label{fig:Gumbel}
\end{figure}

\begin{figure}[h]
\centering
\begin{subfigure}{.48\textwidth}
  \centering
  \includegraphics[width=\linewidth,trim={0 0 0 0cm}]{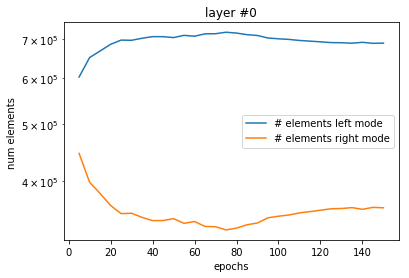}  
  \caption{}
  \label{}
\end{subfigure}
\begin{subfigure}{.48\textwidth}
  \centering
  \includegraphics[width=\linewidth,trim={0 0cm 0 0}]{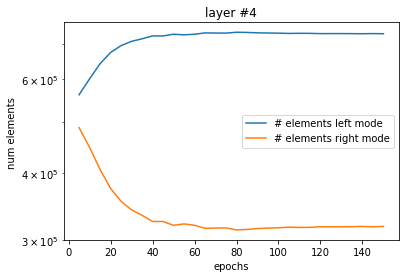}  
  \caption{}
  \label{}
\end{subfigure}
\begin{subfigure}{.48\textwidth}
  \centering
  \includegraphics[width=\linewidth,trim={0 0cm 0 0}]{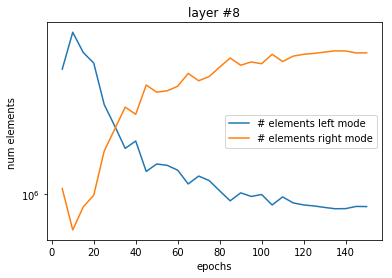}  
  \caption{}
  \label{}
\end{subfigure}
\begin{subfigure}{.48\textwidth}
  \centering
  \includegraphics[width=\linewidth,trim={0 0cm 0 0}]{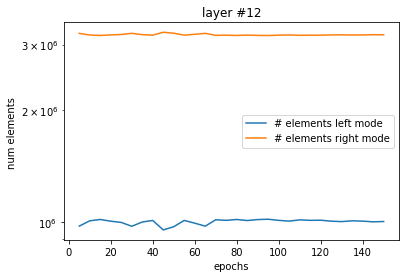}  
  \caption{}
  \label{}
\end{subfigure}
\caption{ The number of elements of each of the two modes of the gradients for ResNet18 trained on Cifar10, on various layers. Notice that the y-axis is in log scale. Different layers vary in behaviour, but the dominant effect is the increase in the proportion of the left mode, indicating that a large portion of the values are mapped to zero by the consequent ReLU layer, because the left modes' values originate from zeros values in the ReLU gradients. Up to 80\% of values might be in the left mode. This has major consequences when trying to prune the bi-modal distributions to values lower than that using our algorithm. However this is less of a problem when pruning to high sparsity levels. The numbering of the layers here is from the deepest to the shallowest, according to the order of the gradients flow in the backward-pass.}
\label{fig:Gumbel}
\end{figure}


\end{document}